\def\reportfontpath{fonts/optimistic}
\documentclass[10pt,a4paper,authoryear]{nus-techreport}
\input{paper-macros}
\definecolor{orbiaGradientStart}{HTML}{0421BE}
\definecolor{orbiaGradientEnd}{HTML}{48B5F4}
\reportsetup{
  accent=1C74E8,secondary=0421BE,resource-layout=stacked,
  title-size=27pt,title-leading=27pt,abstract-breakable=true,
  title-author-gap=18pt,author-affiliation-gap=14pt,
  affiliation-resource-gap=14pt,resource-role-gap=10pt
}
\colorlet{orbiareview}{reportAccent}

\tcbset{nrAbstract/.append style={%
  fontupper={\fontsize{11pt}{13.5pt}\selectfont}}}

\newsavebox{\orbiaFooterPageBox}
\makeatletter
\renewcommand\ps@nusreport{%
  \def\@oddfoot{%
    \sbox{\orbiaFooterPageBox}{{\sffamily\bfseries
      \fontsize{\nr@pagesize}{\nr@pageleading}\selectfont
      \color{reportAccent}\thepage}}%
    \ifnum\value{page}>1\relax
      {\color{reportAccent}%
        \rule[\dimexpr0.5\ht\orbiaFooterPageBox-1.25pt\relax]{12mm}{2.5pt}}%
    \fi
    \hfil\usebox{\orbiaFooterPageBox}}%
  \let\@evenfoot\@oddfoot
  \let\@oddhead\@empty
  \let\@evenhead\@empty}
\makeatother

\colorlet{orbiaTableHeader}{reportAccent!10!white}
\colorlet{orbiaTableStripe}{reportAccent!3!white}
\colorlet{orbiaTableGroupHeader}{orbiaGradientEnd!16!white}
\colorlet{orbiaTableGroupStripe}{orbiaGradientEnd!4!white}
\colorlet{orbiaTableHighlight}{reportAccent!12!white}
\colorlet{orbiaRankFirst}{reportAccent!32!white}
\colorlet{orbiaRankSecond}{reportAccent!19!white}
\colorlet{orbiaRankThird}{reportAccent!8!white}

\DeclareCaptionFont{orbiaCaption}{\fontsize{10.5pt}{13pt}\selectfont}
\makeatletter
\AtEndCaption{%
  \ifdefstring{\@captype}{table}{\vspace{-0.1cm}}{}%
}
\makeatother

\makeatletter
\renewcommand\reportaffiliation[2][]{%
  \ifdefempty{\nr@affiliations}{}{\gappto\nr@affiliations{,~~}}%
  \gappto\nr@affiliations{{\normalsize\ifstrempty{#1}{}{\textsuperscript{#1}}#2}}}
\renewcommand\nr@resourceformat[3]{{%
  \sffamily\fontsize{\nr@authorsize}{\nr@authorleading}\selectfont
  \makebox[1.2em][c]{\textcolor{reportAccent}{#1}}\hspace{0.4em}%
  \ifstrempty{#3}{#2}{\href{#3}{#2}}%
  \ifdefstring{\nr@resourcelayout}{stacked}{\par}{}%
}}
\renewcommand\nr@resourceseparator{\par\vspace{4pt}}
\makeatother

\newsavebox{\orbiaTitleBox}
\newsavebox{\orbiaShadedBox}
\DeclareRobustCommand{\orbiaGradientText}[1]{%
  \begingroup
  \sbox{\orbiaTitleBox}{#1}%
  \leavevmode
  \sbox{\orbiaShadedBox}{%
    \pdfliteral page{q 7 Tr}%
    \rlap{\usebox{\orbiaTitleBox}}%
    \pdfliteral page{0 Tr}%
    \tikz[baseline=0pt,inner sep=0pt,outer sep=0pt]{%
      \path[use as bounding box]
        (0,{-\dp\orbiaTitleBox}) rectangle
        ({\wd\orbiaTitleBox},{\ht\orbiaTitleBox});
      \shade[top color=orbiaGradientStart,bottom color=orbiaGradientEnd,
        shading angle=45]
        (0,{-\dp\orbiaTitleBox}) rectangle
        ({\wd\orbiaTitleBox},{\ht\orbiaTitleBox});
    }%
    \pdfliteral page{Q}%
  }%
  \pdfxform\orbiaShadedBox
  \pdfrefxform\pdflastxform
  \endgroup
}
\newsavebox{\orbiaWordMetrics}
\DeclareRobustCommand{\OrbiaWord}{%
  \texorpdfstring{%
    \begingroup
    \sbox{\orbiaWordMetrics}{{\fontsize{29pt}{29pt}\selectfont Orbia}}%
    \raisebox{0pt}[\ht\orbiaWordMetrics][\dp\orbiaWordMetrics]{%
      {\fontsize{36pt}{27pt}\selectfont\orbiaGradientText{Orbia}}}%
    \endgroup
  }{Orbia}}
\newsavebox{\orbiaFirstTitleLine}
\newsavebox{\orbiaSecondTitleLine}
\DeclareRobustCommand{\orbiaTitleLines}{%
  \sbox{\orbiaFirstTitleLine}{\OrbiaWord{}: Do Generated Videos}%
  \sbox{\orbiaSecondTitleLine}{Form a Consistent 3D World?}%
  \usebox{\orbiaFirstTitleLine}\\
  \resizebox{\wd\orbiaFirstTitleLine}{!}{\usebox{\orbiaSecondTitleLine}}%
}
\title{\texorpdfstring{\orbiaTitleLines}{Orbia: Do Generated Videos Form a Consistent 3D World?}}
\reportauthor[1*]{Dongyue~Lu}
\reportauthor[1*]{Guian~Fang}
\reportauthor[1]{Hur~Lim}
\reportauthor[1]{Yonggang~Wu}
\reportauthor[1]{Yanrong~Wang}
\reportauthor[1]{Hongshan~Chen}
\reportauthor[2]{Shengju~Qian}
\reportauthor[1]{Yue~Liao}
\reportauthor[1]{Wei~Chow}
\reportauthor[1]{Lingdong~Kong}
\reportauthor[3]{Tianxin~Huang}
\reportauthor[4]{Yu~Yang}
\reportauthor[5]{Hongji~Yang}
\reportauthor[6]{Junchao~Huang}
\reportauthor[1]{Ao~Liang}
\reportauthor[7]{Yihao~Wang}
\reportauthor[7]{Zihan~Wang}
\reportauthor[8]{Rong~Li}
\reportauthor[1]{Hanlin~Chen}
\reportauthor[2]{Xin~Wang}
\reportauthor[1,\dagger]{Wei~Tsang~Ooi}
\reportauthor[1,\dagger]{Mike~Zheng~Shou}
\reportauthor[1,\dagger]{Shuicheng~Yan}
\reportaffiliation[1]{National University of Singapore}
\reportaffiliation[2]{LIGHTSPEED}
\makeatletter
\let\orbiaFirstAffiliationRow\nr@affiliations
\let\nr@affiliations\@empty
\makeatother
\reportaffiliation[3]{HKU}
\reportaffiliation[4]{ZJU}
\reportaffiliation[5]{University of Macau}
\reportaffiliation[6]{CUHK(SZ)}
\reportaffiliation[7]{Aalto University}
\reportaffiliation[8]{HKUST(GZ)}
\makeatletter
\preto\nr@affiliations{\orbiaFirstAffiliationRow\par\vspace{7pt}}
\makeatother
\reportrole[*]{Equal contributions~}
\reportrole[\dagger]{Corresponding authors}
\reportresource[project]{Project Page:~~\href{https://orbia-bench.github.io/}{\texttt{https://orbia-bench.github.io}}}{}
\reportresource[code]{GitHub:~~\href{https://github.com/worldbench/Orbia}{\texttt{https://github.com/worldbench/Orbia}}}{}
\reportresource[data]{Benchmark:~~\href{https://huggingface.co/datasets/Orbia/Orbia}{\texttt{https://huggingface.co/datasets/Orbia/Orbia}}}{}
\date{October 2026}
\abstract{Video generators serve as interactive worlds for exploration and simulation, yet realistic and visually appealing frames do not necessarily depict a consistent 3D world. Existing benchmarks assess selected \textbf{geometric properties} or use vision-language models to judge \textbf{memory} and \textbf{consistency}, leaving systematic 3D evaluation underexplored. We introduce \orbia{}, the first benchmark dedicated to a systematic and comprehensive evaluation of \textbf{3D consistency} in interactive world models. Through controlled exploration and revisit, we organize the evaluation into three dimensions: input preservation, generated-content persistence, and 3D self-consistency. For input preservation, registered RGB-D references and annotated anchors support appearance, geometry, and spatial stability measurements. For generated-content persistence, we compare revisits to regions absent from the input image with their first generated observations to test whether models preserve what they create. For 3D self-consistency, cross-view depth and feature correspondence, regional scale agreement, and held-out Gaussian rendering test whether generated views describe the same 3D scene. Minute-long exploration and revisits further test long-term consistency. Our construction and annotation pipeline brings together public datasets, self-captured scenes, and simulation, yielding 720 instances across indoor and outdoor environments and six exploration-and-revisit protocols. Evaluation spans 27 text-, action-, and camera-conditioned models through native-interface adapters. Our \orbia{} benchmark provides a \textbf{systematic basis} for diagnosing 3D inconsistency and developing more persistent video world models.}
\AtBeginDocument{\hypersetup{%
  citecolor=reportAccent,linkcolor=reportAccent,urlcolor=reportAccent,
  pdftitle={Orbia: Do Generated Videos Form a Consistent 3D World?}}}

\newsavebox{\orbiaHeaderLightspeedBox}
\AddToHookNext{shipout/foreground}{%
  \sbox{\orbiaHeaderLightspeedBox}{\includegraphics[width=4cm]{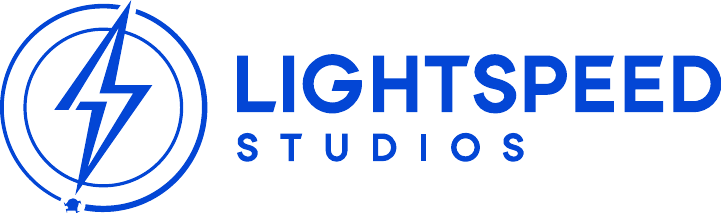}}%
  \begin{tikzpicture}[remember picture,overlay]
    \node[anchor=north east,inner sep=0pt] (orbiaHeaderLightspeed) at
      ([xshift=-2.5cm,yshift=-1.25cm]current page.north east)
      {\usebox{\orbiaHeaderLightspeedBox}};
    \node[anchor=south west,inner sep=0pt] at
      ([xshift=2.5cm]current page.north west |- orbiaHeaderLightspeed.south)
      {\includegraphics[height=\ht\orbiaHeaderLightspeedBox]{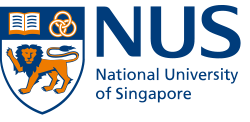}};
  \end{tikzpicture}%
}

\begin{document}
  \addtocontents{toc}{\protect\setcounter{tocdepth}{-1}}%
  \maketitle
  \vspace{0.3cm}
  \par\noindent
\begin{minipage}{\linewidth}
\centering
\includegraphics[width=\linewidth]{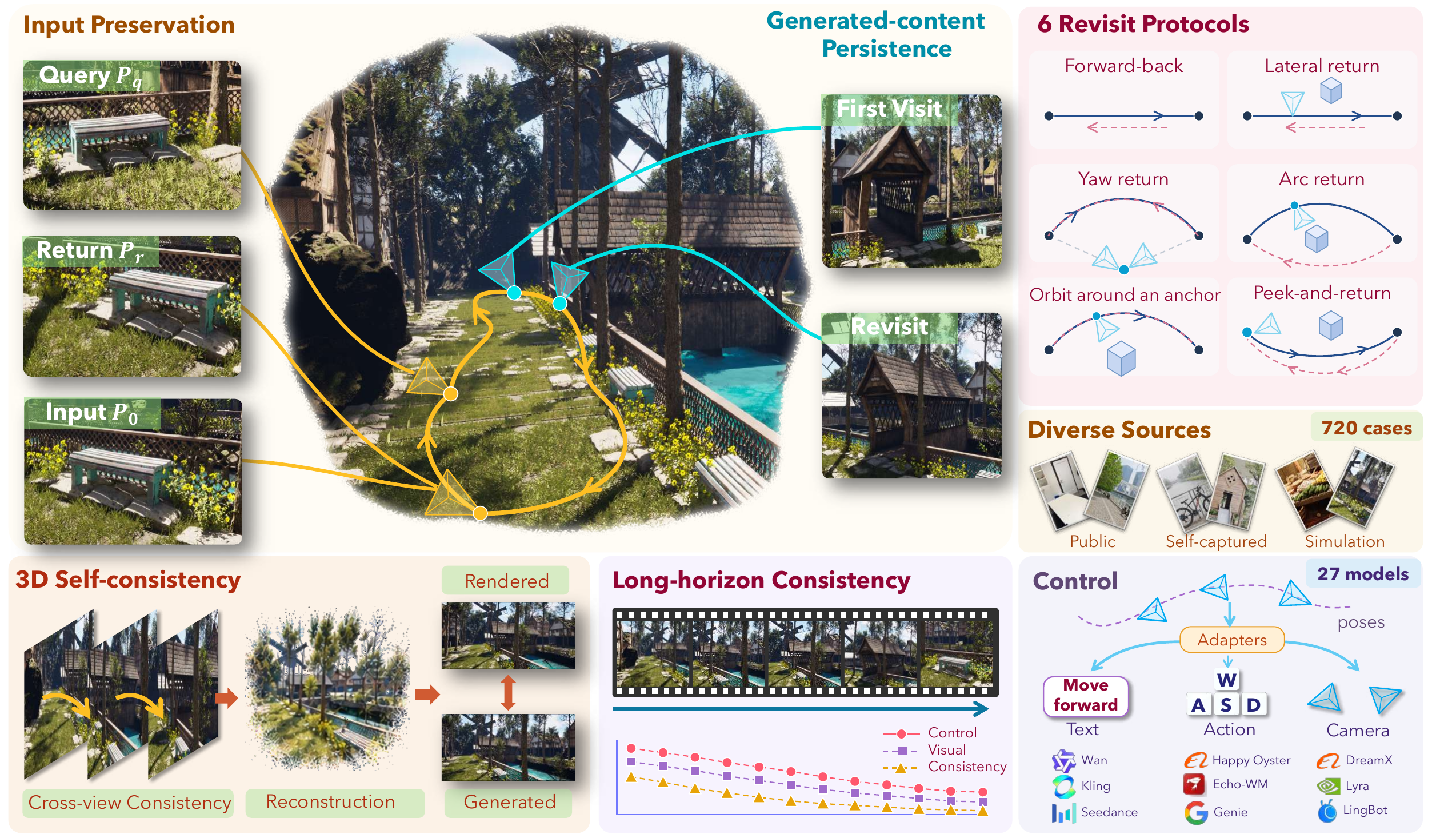}
\captionof{figure}{\textbf{Overview of \orbia{}.} This work introduces \orbia{}, a benchmark for evaluating 3D consistency along three dimensions: \textsuperscript{1}\textbf{\textit{input preservation}}, \textsuperscript{2}\textbf{\textit{generated-content persistence}}, and \textsuperscript{3}\textbf{\textit{3D self-consistency}}. \textbf{Six exploration-and-revisit templates} support short and minute-long evaluation across \textit{public datasets, self-captured scenes, and UE simulation}. We benchmark world models and video generators through \textit{text, action, and camera-pose interfaces}.}
\label{fig:teaser}
\end{minipage}
\par
  \clearpage
  \reportabstract
  \vspace{0.2cm}

\section{Introduction}
\label{sec:intro}
Video generation is expanding from visual synthesis toward world simulation~\citep{brooks2024sora,wan2025wan,deepmind2025veo31,gao2025seedance}, supporting interactive games~\citep{bruce2024genie,valevski2024gamengen}, open-domain exploration~\citep{parkerholder2025genie3,sun2025worldplay}, generative rendering~\citep{yu2024wonderworld,shen2026lyra2}, and embodied simulation~\citep{yang2023unisim,agarwal2025cosmos}. This shift is contextualized by \citep{kong2025worldmodelingsurvey}, which organizes world modeling around video, occupancy, and LiDAR representations. These applications require observations across time and viewpoints to remain compatible with a persistent 3D world. A player should return to the same environment, a renderer should reveal mutually consistent surfaces, and an embodied agent should reuse previously observed spatial relationships. Visually convincing frames alone cannot establish this structural reliability across viewpoints and repeated visits. As generated videos become environments for exploration and interaction, a fundamental question arises: \textbf{Do generated videos form a consistent 3D world?}

Existing benchmarks address different aspects of this question. These protocols assess perceptual quality~\citep{huang2024vbench,liu2024evalcrafter}, local or scene-level geometric self-coherence~\citep{duan2025worldscore,ying2026wbench,xu2026worldmark,zhao2026worldolympiad}, or memory and state retention upon revisit using same-view appearance comparisons, point-cloud comparisons, and vision-language judgments~\citep{ye2026mind,lu2026wrbench,xu2026worldroam}. WorldLens~\citep{liang2026worldlens} broadens driving-world evaluation to reconstruction, action following, downstream tasks, and human preference. Yet \textbf{systematic, quantitative evaluation of 3D consistency} remains underexplored. Geometric agreement does not guarantee content preservation, nor does a successful return establish consistency throughout exploration. A dedicated benchmark must jointly evaluate 3D structure and the preservation of input and generated content across viewpoints and over time during exploration and repeated revisits.

We introduce \orbia{}, the first benchmark dedicated to a systematic and comprehensive evaluation of 3D consistency in interactive world models. We organize evaluation into \textbf{three dimensions} (Figure~\ref{fig:teaser}). 

\dimensionlabel{evidence}{(1) Input preservation} assesses whether the model preserves the environment shown in the input as the viewpoint changes. Registered RGB-D references and annotated anchors measure the appearance, geometry, and spatial relationships of observed content. Exploration also reveals regions outside the input, whose appearance the model must generate. 

\dimensionlabel{persistence}{(2) Generated-content persistence} compares newly revealed content across visits, using its first generated appearance as the reference. The model is free to create unseen content, but should preserve what it has created. Beyond content retention, generated views must agree in their underlying 3D structure. 

\dimensionlabel{coherence}{(3) 3D self-consistency} tests whether generated views describe the same 3D scene through cross-view depth and feature correspondence, regional scale agreement, and held-out Gaussian rendering. These measurements examine both local surface compatibility and scene-level coherence. To assess resistance to temporal drift under longer exploration and more frequent revisits, we extend sequences to one minute and repeatedly return to previously visited regions. This tests whether models maintain scene geometry and content as the exploration history grows. We also examine how control accuracy and visual quality relate to the three consistency dimensions.

Joint evaluation requires cross-view references and controlled revisits. Our \textbf{construction pipeline} prepares registered RGB-D pairs, verifies anchor correspondence and surface support, and synthesizes trajectories connecting the reference poses. We select revisits to newly revealed regions and manually review anchors, reference views, and trajectories. To cover diverse scene distributions, we draw on \textsuperscript{1}\textit{real-world datasets}, \textsuperscript{2}\textit{self-captured scenes}, and \textsuperscript{3}\textit{simulated environments}, spanning indoor and outdoor settings in both real and virtual worlds. Processing over \textit{110k} source samples and investing over \textit{300} hours in annotation and curation yields \textbf{720 evaluation instances}, comprising $480$ short cases and $240$ paired minute-long extensions. Six exploration-and-revisit protocols cover trajectories with distinct spatial demands. Model-specific adapters translate the trajectories into text instructions, action sequences, or camera conditions for evaluation across control interfaces for the same exploration and revisit tasks.

To summarize, our contributions are fourfold:
\begin{itemize}[itemsep=2pt,parsep=0pt,topsep=3pt,partopsep=0pt]
\item We introduce \orbia{}, the first benchmark dedicated to systematic, quantitative evaluation of \textbf{3D consistency} in interactive world models across viewpoints and exploration horizons.
\item We decompose 3D consistency into \textcolor{evidence}{\textbf{\textit{input preservation}}}, \textcolor{persistence}{\textbf{\textit{generated-content persistence}}}, and \textcolor{coherence}{\textbf{\textit{3D self-consistency}}} to systematically assess content and geometry.
\item We build a \textbf{unified annotation pipeline} yielding 720 instances with registered evidence and controlled revisits across real and simulated environments.
\item We evaluate \textbf{27} text-, action-, and camera-conditioned models, examining how 3D consistency relates to control execution, visual quality, and exploration duration.
\end{itemize}

\begin{table}[t]
\centering
\begin{tablebody}{0.8pt}{1.3}

\caption[Comparison with representative benchmarks.]{\textbf{Comparison with representative benchmarks.} We compare input formats, data sources, evaluation references, and five consistency tests: \textsuperscript{1}\textit{local 3D self-consistency}, \textsuperscript{2}\textit{global 3D self-consistency}, \textsuperscript{3}\textit{input-object preservation}, \textsuperscript{4}\textit{revisit consistency}, and \textsuperscript{5}\textit{long-term drift}.
Inputs: \bt\ text, \ba\ action, \bc\ camera pose.
Data: \breal\ public real imagery, \bself\ self-captured, \bsim\ simulated, \bgen\ generated.
References: {\scriptsize\brgb}\ color images, {\scriptsize\bdepth}\ depth maps, {\scriptsize\bpose}\ camera poses, {\scriptsize\baction}\ reference actions.}
\label{tab:related}
\fontsize{7.5}{9}\selectfont
\begin{tabularx}{\linewidth}{@{}llll*{4}{>{\hsize=.9625\hsize\linewidth=\hsize\centering\arraybackslash}X}>{\hsize=1.15\hsize\linewidth=\hsize\centering\arraybackslash}X@{}}
\toprule
\rowcolor{orbiaTableHeader}
\multicolumn{4}{c}{\textbf{Benchmark Setup}} & \multicolumn{5}{c}{\textbf{Consistency Evaluation}}\\
\cmidrule(l){5-9}
\textbf{Benchmark} & \textbf{Input} & \textbf{Data} & \textbf{Reference} & \tablehead{Local\\Consis.} & \tablehead{Global\\Consis.} & \tablehead{Input\\Preserv.} & \tablehead{Revisit\\Consis.} & \tablehead{Long-term\\Drift}\\[-1pt]
\midrule
WorldScore~\citep{duan2025worldscore} & \bt\,\bc & \breal\,\bsim\,\bgen & \bpose & \byes & \bno & \bno & \bno & \bno\\
\rowcolor{orbiaTableStripe}
WorldOlympiad~\citep{zhao2026worldolympiad} & \bt\,\ba\,\bc & \breal\,\bsim & \bpose & \bno & \byes & \bno & \bno & \bno\\
WBench~\citep{ying2026wbench} & \bt\,\ba\,\bc & \breal\,\bself\,\bgen & \bpose & \byes & \bno & \byes & \byes & \bno\\
\rowcolor{orbiaTableStripe}
WorldMark~\citep{xu2026worldmark} & \bt\,\ba\,\bc & \breal\,\bsim\,\bgen & \bpose & \byes & \bno & \bno & \bno & \byes\\
iWorld-Bench~\citep{fang2026iworld} & \bt\,\ba\,\bc & \breal\,\bsim & \brgb\,\bpose & \bno & \bno & \bno & \byes & \bno\\
\rowcolor{orbiaTableStripe}
MIND~\citep{ye2026mind} & \ba & \bsim & \brgb\,\bpose & \bno & \bno & \bno & \byes & \bno\\
WRBench~\citep{lu2026wrbench} & \bt\,\bc & - & \bpose & \bno & \bno & \byes & \byes & \bno\\
\rowcolor{orbiaTableStripe}
WorldRoamBench~\citep{xu2026worldroam} & \bt\,\ba & \breal\,\bgen & \baction & \byes & \bno & \byes & \byes & \byes\\
LoopBench~\citep{wu2026worldtrace} & \ba & - & - & \bno & \bno & \bno & \byes & \byes\\
\rowcolor{orbiaTableStripe}
R2M-Bench~\citep{gu2026r2m} & \ba\,\bc & - & - & \byes & \bno & \bno & \byes & \bno\\
CycleBench~\citep{gu2026worldcycle} & \ba & \breal\,\bgen & \baction & \bno & \bno & \bno & \byes & \byes\\
\rowcolor{orbiaTableStripe}
MemoBench~\citep{chen2026memobench} & \bt\,\bc & \bself\,\bsim & \brgb\,\bdepth\,\bpose & \byes & \bno & \byes & \byes & \bno\\
\midrule
\rowcolor{orbiaTableHighlight}
\orbia{} (Ours) & \bt\,\ba\,\bc & \breal\,\bself\,\bsim & \brgb\,\bdepth\,\bpose & \byes & \byes & \byes & \byes & \byes\\
\bottomrule
\end{tabularx}
\end{tablebody}
\end{table}

\let\orbiaRelatedBottomTitleSpace\bottomtitlespace
\def\bottomtitlespace{6\baselineskip}
\section{Related Work}
\label{sec:related}
\let\bottomtitlespace\orbiaRelatedBottomTitleSpace

\subsection{Interactive Video World Models}
\label{sec:related_video_world_models}
Large-scale video generators~\citep{brooks2024sora,polyak2024moviegen,deepmind2025veo31,kong2024hunyuanvideo,wan2025wan} establish visual and temporal priors for generative simulation by synthesizing scene appearance and motion from text or visual inputs.
Interactive video world models build on these capabilities by conditioning future observations on user-controlled inputs.
Action-conditioned systems~\citep{bruce2024genie,valevski2024gamengen,li2025gamecraft,he2025matrixgame2,jiang2026abotworld} support movement and interaction, while language-directed methods~\citep{mao2025yume15,tang2025gamecraft2} provide semantic control over unfolding events.
Explicit camera-control approaches~\citep{wang2023motionctrl,he2024cameractrl} further expose viewpoint trajectories as a generation interface.
Together, these interfaces support exploring a generated environment through successive choices, introducing dependencies between current outputs, earlier observations, and the user's interaction history.

Streaming and open-domain systems~\citep{gao2026lingbot,gao2026lingbot2,team2026dreamx,zhu2025astra,zhu2026sanawm,duan2026echo,huang2026solarwm} pursue sustained, responsive exploration across diverse environments.
Beyond generating locally coherent transitions, maintaining a world through exploration requires relating new observations to previously visited regions.
History-conditioned generation and context memory~\citep{li2025gamecraft,sun2025worldplay} seek to preserve information from past observations when extending a rollout.
These developments also connect to the view of world models as interactive proxies providing feedback for agents~\citep{yang2026quovadis}.
They motivate evaluating the stability of a generated environment alongside visual quality and control adherence, particularly when users leave a region and later return.

Agentic world modeling connects predictive simulation to decision-making across environments~\citep{chu2026agenticworld}.
Visual navigation approaches couple language-conditioned future observation prediction with policy learning~\citep{dong2026lcvn}, or integrate visual foresight, planning, and memory within a unified model~\citep{dong2026uniwm}.
For autonomous driving, vision-language-action systems bring together perception, linguistic reasoning, and actionable outputs~\citep{hu2025vlasurvey}; OneVL uses future visual prediction and language explanations to supervise compact latent reasoning for planning~\citep{lu2026onevl}.
These applications make the persistence of spatial evidence relevant to both generation and downstream decisions.

A complementary line examines how generative capabilities are learned and aligned.
DanceOPD composes generative capability fields on student-induced states~\citep{zhou2026danceopd}, while on-policy diffusion self-distillation converts reward guidance into intermediate prediction targets~\citep{zhou2026diffusionopsd}.
These image-generation studies provide a training perspective on capability alignment that complements evaluation of consistency during interactive rollouts.

\subsection{3D Interactive World Models}
\label{sec:related_3d}
Explicit scene representations enable navigable world generation and editing~\citep{yu2024wonderworld,liang2024wonderland,szymanowicz2025bolt3d,worldlabs2025marble,worldlabs2026atlas}.
These systems associate generated content with a scene that can be queried from different viewpoints.
Complementary methods couple generative priors with reconstructed geometry or spatial conditioning~\citep{yu2024viewcrafter,ren2025gen3c,huang2025voyager,lu2026see4d,yang2026neoverse,inspatio2026worldfm,chen2026robust}.
Point clouds, Gaussian representations, and warped observations provide camera-aligned evidence for synthesizing unseen regions, while allowing learned priors to complete missing appearance.
Geometry may directly determine a rendered output or instead guide a generator that retains freedom to revise appearance and complete occluded regions.
This combination connects the flexibility of image and video generation with structured constraints on viewpoint changes.
Geometry-grounded weather video synthesis offers a related editing perspective, combining semantic appearance conditioning, simulated particle dynamics, and scene geometry to control weather effects~\citep{qian2026weathervid}.

The representation taxonomy of \citet{kong2025worldmodelingsurvey} places video, occupancy, and LiDAR generation within a common 3D/4D world-modeling landscape. Sensor-based world modeling provides complementary representations of dynamic scenes.
DynamicCity models semantic 4D occupancy~\citep{bian2025dynamiccity}, while LiDARCrafter generates and edits temporally coherent LiDAR sequences~\citep{liang2026lidarcrafter} and U4D introduces uncertainty-conditioned generation~\citep{xu2026u4d}.
Cross-sensor pretraining~\citep{kong2026largead} and multimodal scene-consistency learning~\citep{kong2025lasermixpp} address representation learning in these sensor-rich settings.
Together, these works motivate assessing scene structure and dynamics through evidence appropriate to each representation.

Spatial memory extends these mechanisms beyond short temporal contexts.
Surfel-indexed view retrieval~\citep{li2025vmem}, geometry-guided correspondences~\citep{shen2026lyra2}, and geometry-aligned or external memory~\citep{team2026alayaworld,yin2026evoke,qian2026matrixgame35} reconnect current viewpoints with previously observed content.
These advances address spatial forgetting and drift, and existing studies include reconstruction, revisit, or geometric-consistency diagnostics.
Nevertheless, using 3D information does not by itself establish whether original input content is faithfully retained, newly generated regions persist after leaving view, and repeated observations agree as one 3D world.
\orbia{} evaluates these complementary requirements jointly under controlled exploration, enabling comparisons across explicit 3D, geometry-guided, and memory-augmented approaches.

\subsection{World Model Evaluation}
\label{sec:related_evaluation}
Video benchmarks characterize visual quality, temporal consistency, and prompt alignment~\citep{huang2024vbench,liu2024evalcrafter}, while faithfulness-oriented evaluations probe physical plausibility and commonsense~\citep{zheng2025vbench2,bansal2024videophy,meng2024phygenbench,li2025worldmodelbench}.
Embodied benchmarks connect generated dynamics to action-level behavior and functional utility~\citep{qin2024worldsimbench,hu2025ewmbench,shang2026worldarena}.
WorldLens~\citep{liang2026worldlens} evaluates driving world models across generation, reconstruction, action following, downstream tasks, and human preference. Complementing this benchmark, \emph{3D and 4D World Modeling: A Survey}~\citep{kong2025worldmodelingsurvey} organizes evaluation perspectives spanning generation, forecasting, planning, reconstruction, and downstream utility.
Together, perceptual and functional criteria answer related but distinct questions about whether generated videos are visually convincing and useful for interaction.

Evaluation reliability also depends on evidence grounding.
DriveBench shows that plausible VLM answers can reflect linguistic priors rather than visual input~\citep{xie2025drivebench}, while Open-o3-Video makes video reasoning traceable through timestamps, objects, and bounding boxes~\citep{meng2026openo3video}.
These findings support pairing language-based assessments with explicit visual and geometric evidence.

Interactive suites examine control responsiveness, trajectory adherence, and geometric consistency~\citep{duan2025worldscore,xu2026worldmark,fang2026iworld,zhao2026worldolympiad}, while multi-turn protocols assess interaction adherence and resulting state changes~\citep{ying2026wbench,tang2025gamecraft2,wu2026omniworldbench}.
Complementary diagnostics isolate deformation and occlusion inconsistency~\citep{gu2025geco}, static-scene reconstructability~\citep{fan2026geot2v}, and reconstruction-grounded manipulation~\citep{wang2026robophys3d}.
Geometric evaluation therefore already addresses several failure modes, with different protocols emphasizing different aspects of spatial coherence.

Memory-oriented tests examine returns to previously observed scenes and recovery of persistent or dynamically updated states~\citep{ye2026mind,lu2026wrbench,chen2026memobench,chen2026reworld}, revisit geometry~\citep{xu2026worldroam}, and calibrated revisit similarity~\citep{gu2026r2m}.
Long-horizon tests track drift and state-returning behavior~\citep{wu2026worldtrace,gu2026worldcycle}, while agent-based evaluators provide transition and persistence judgments~\citep{chen2026harnesseval}.
Building on these complementary perspectives, \orbia{} jointly measures preservation of the input scene, persistence of newly generated regions, and 3D self-consistency across repeated exploration.
This organization separates fidelity to observed evidence from retention of generated content and compatibility of generated views.
Table~\ref{tab:related} summarizes the coverage of representative benchmarks.

\section{The \orbia{} Benchmark}
\label{sec:benchmark}
\orbia{} systematically evaluates the 3D consistency of interactive video world models. In this section, we introduce the evaluation principles and protocols (Section~\ref{sec:benchmark-protocols}), the data construction and annotation pipeline (Section~\ref{sec:benchmark-construction}), and the dataset scale and coverage across scene sources and exploration horizons (Section~\ref{sec:benchmark-coverage}).

\subsection{Evaluation Principles and Protocols}
\label{sec:benchmark-protocols}
\noindent\textbf{Interactive World Generation.}
An interactive video world model generates observations from an initial image $I_0$, a scene description $s$, and user inputs $\mathcal U$, yielding $\hat V=\mathcal M(I_0,s,\mathcal U)$. Depending on the model's interface, $\mathcal U$ comprises requested text instructions $T$, discrete actions $A$, or camera poses $P$.

\noindent\textbf{Consistency Decomposition.}
Existing works examine aspects of consistency, including local geometric agreement or content recovery upon return~\citep{duan2025worldscore,ying2026wbench}. A systematic evaluation should account for what the inputs constrain, what the model is free to generate, and whether the generated content forms a coherent 3D scene. We evaluate three dimensions covering input content, newly generated regions, and 3D structure.

\begin{dimensioncards}
\dimensioncard{evidence}{Input preservation}{%
We test whether the model \textbf{preserves the scene shown in the initial image}. Distinctive static objects and structures are annotated as \emph{\textbf{anchors}}, providing landmarks that should remain in place across viewpoint changes and reappear when the camera returns. Evaluating anchors individually prevents a missing landmark from being overlooked in a scene-wide average dominated by other visible objects and scene content.}

\dimensioncard{persistence}{Generated-content persistence}{%
We evaluate \textbf{regions absent from the input image}. Their appearance is not prescribed, but the controls specify when the model should revisit them. We compare these revisits with the first generated observations to test whether the same content remains at corresponding locations when the camera revisits the region.}

\dimensioncard{coherence}{3D self-consistency}{%
We examine the generated video itself, using recovered cameras to test \textbf{cross-view agreement} and whether a \textbf{shared 3D reconstruction} explains observations beyond those used to build it. These tests assess internal coherence independently of the requested camera trajectory used to produce the video.}

\end{dimensioncards}
\noindent\textbf{Exploration-and-Revisit Protocols.}
Exploration-and-revisit trajectories test input preservation, generated-content persistence, and 3D self-consistency within one rollout. Six templates challenge preservation under changes in parallax and viewing direction: \textsuperscript{1}\textit{forward and backward}, \textsuperscript{2}\textit{lateral return}, \textsuperscript{3}\textit{yaw return}, \textsuperscript{4}\textit{arc return}, \textsuperscript{5}\textit{orbit around an anchor}, and \textsuperscript{6}\textit{peek and return}. In peek-and-return, the camera moves past an anchor into an unseen region, returns, and revisits that region on a second excursion. Each trajectory starts at $P_0$ and returns to the same camera pose at $P_r$. For \textcolor{evidence}{\textbf{input preservation}}, we compare the return view against the initial RGB image and reference depth. A cross-view query at $P_q$ uses registered RGB-D and annotated anchors to evaluate input-visible surfaces from another viewpoint. For \textcolor{persistence}{\textbf{generated-content persistence}}, we pair a first visit at $P_{\mathrm{first}}$ to a region outside the input view with a revisit at $P_{\mathrm{revisit}}$ after exploring elsewhere. The first generated observation provides the reference for comparing appearance and geometry across visits. For \textcolor{coherence}{\textbf{3D self-consistency}}, views sampled along the trajectory support comparisons of corresponding surfaces across views and reconstruction tests using held-out generated views.

\subsection{Reference-Guided Data Construction}
\label{sec:benchmark-construction}
\begin{figure}[t]
\centering
\includegraphics[width=\linewidth]{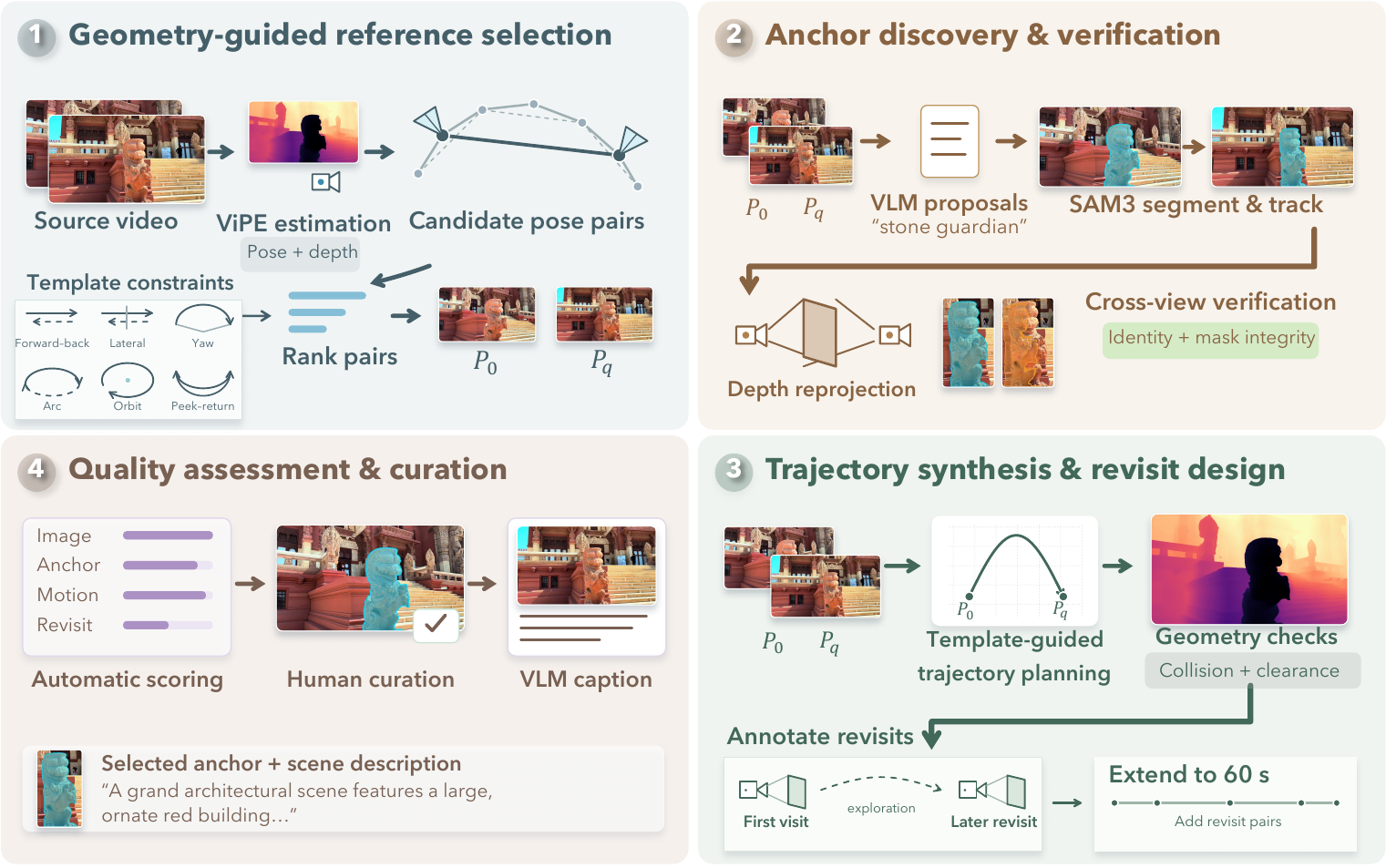}

\caption{\textbf{Data construction and annotation pipeline.} \textsuperscript{1}\textbf{Reference selection} uses geometry and template constraints. \textsuperscript{2}\textbf{Anchor verification} combines VLM proposals, segmentation, tracking, and cross-view surface checks. \textsuperscript{3}\textbf{Trajectory synthesis} defines geometrically valid exploration and revisits. \textsuperscript{4}\textbf{Quality review} includes ranking, human curation, and captioning.}
\label{fig:construction}

\end{figure}

We combine \textsuperscript{1}\textit{public datasets}, \textsuperscript{2}\textit{self-captured scenes}, and \textsuperscript{3}\textit{UE simulation} to cover diverse indoor and outdoor environments. Each case pairs \textbf{\textit{registered RGB-D observations}} from different viewpoints with a synthesized revisit trajectory that follows one of six templates and passes through both reference poses. This enables controlled exploration from existing observations without frame-by-frame RGB-D references. Figure~\ref{fig:construction} summarizes the four-stage construction and annotation pipeline, from selecting reference views to reviewing the completed cases.

\noindent\textbf{Geometry-Guided View Selection.}
We retain existing geometry and recover missing annotations with ViPE~\citep{huang2025vipe}. We select input--query pairs $(P_0,P_q)$ from registered poses using each template's rotation and translation requirements. Ranking favors sufficient viewpoint change while balancing coverage across templates and scene content.

\noindent\textbf{Anchor Discovery and Verification.}
A VLM~\citep{qwen2026qwen35} proposes bounded, distinctive static landmarks visible at both endpoints. SAM3~\citep{carion2025sam3} segments them independently at both endpoints. We filter masks by area, connectivity, and completeness, then track shortlisted objects to verify \textbf{\textit{cross-view identity}}. Back-projecting input anchors with depth tests their visibility and depth agreement in the query view, verifying \textbf{\textit{cross-view surface support}} for landmarks visible in both the input and query observations.

\noindent\textbf{Trajectory Synthesis and Revisit Design.}
We fit template parameters to synthesize trajectories through $P_0$ and $P_q$ that return to the initial pose. Paths are checked against scene geometry and revised for clearance. We select \textbf{\textit{temporally separated visits}} to newly revealed regions based on pose proximity, novelty relative to the input, and shared visible surfaces. Minute-long extensions increase travel distance while retaining the short case's \textbf{\textit{input, anchors, and reference poses}}. Additional revisit pairs test whether appearance and geometry persist in these regions after further exploration.

\noindent\textbf{Quality Assessment and Human Curation.}
We rank cases by image sharpness, anchor visibility, mask integrity, camera motion, and shared surface support across revisits. Reviewers inspect references, anchor overlays, projected support, and trajectories to correct or reject cases. A VLM produces factual scene captions from source frames to provide scene descriptions alongside the initial images as model inputs.

\subsection{Dataset Scale and Coverage}
\label{sec:benchmark-coverage}

We curate $240$ cases from approximately $110{,}000$ source samples across four public datasets, ScanNet++~\citep{yeshwanth2023scannetpp}, SpatialVID~\citep{wang2025spatialvid}, Sekai~\citep{li2025sekai}, and MiraData~\citep{ju2024miradata}. For simulation, we collect $48$ UE environments and render $365$ candidate videos, retaining $180$ cases after curation. We also record $85$ videos ourselves and retain $60$. Together, data annotation and \textbf{\textit{manual curation}} required over \textbf{\textit{300 hours}}.

\orbia{} contains \textbf{\textit{720 cases}} and \textbf{\textit{859 anchor evaluation targets}}, with \textbf{\textit{480 short cases}} and \textbf{\textit{240 paired minute-long extensions}}, standardized at $720$p and $16$ FPS. The short cases cover $120$ indoor and $360$ outdoor settings, including $300$ cases from real-world observations and $180$ from simulation. Short trajectories last approximately $22$ seconds with one revisit; their minute-long extensions include three revisits to test preservation through longer exploration and repeated returns. Source-specific construction procedures and dataset composition are detailed in Appendix~\ref{app:data-construction}.

\section{Evaluating 3D Consistency}
\label{sec:evaluation}
Given a generated video, we estimate depth and poses with DA3~\citep{lin2025da3} and evaluate the designated observations against registered RGB-D references or earlier generated views. Cross-view agreement and held-out rendering assess internal 3D consistency. The three dimensions differ in their reference: the input scene for input preservation, the model's own first observation for generated-content persistence, and the generated video itself for 3D self-consistency. The survey by \citet{kong2025worldmodelingsurvey} distinguishes generation, reconstruction, and task-oriented evaluation; here, we focus on complementary content and geometry tests during interactive exploration. We describe the core consistency metrics and long-horizon evaluation below. Detailed metric definitions and computation procedures are provided in Appendix~\ref{app:metrics}.

\begin{evaluationparagraphs}
\noindent\textcolor{evidence}{\raisebox{0.15ex}{\rule{0.45em}{0.45em}}}\hspace{0.4em}\dimensionlabel{evidence}{Input Preservation.}
To assess input preservation at two checkpoints, we compare the return at $P_r$ with the input at $P_0$ and the query at $P_q$ with \textbf{\textit{registered RGB-D references on input-visible surfaces}}. Surfaces revealed for the first time at $P_q$ have no reference and are excluded from the query comparison. \textsuperscript{1}\textit{Scene Appearance Fidelity} quantifies appearance fidelity at return and query views, while \textsuperscript{2}\textit{Anchor Appearance Fidelity} assesses the preservation of distinctive landmarks across viewpoints and on return. Both use DINO similarity~\citep{oquab2023dinov2}, PSNR, SSIM, and LPIPS on input-visible scene or anchor regions, with \textbf{\textit{equal weight per anchor}}. A missing or altered landmark is one of the most visible failures of a generated world, yet it often occupies a small part of the frame, so a scene-wide average can hide it. Scoring each anchor separately makes such failures directly measurable. \textsuperscript{3}\textit{Scene Geometry Fidelity} quantifies geometric fidelity to the observed scene, while \textsuperscript{4}\textit{Anchor Geometry Fidelity} assesses anchor geometry relative to the surrounding scene. Both measure depth AbsRel against the registered reference using a shared scene-level scale. \textsuperscript{5}\textit{Anchor Identity Consistency} tests anchor identity retention during exploration. SAM3.1~\citep{carion2025sam3} tracks input masks toward $P_q$. We average masked DINO cosine similarity to the input over four times and equally across anchors.

\noindent\textcolor{persistence}{\raisebox{0.15ex}{\rule{0.45em}{0.45em}}}\hspace{0.4em}\dimensionlabel{persistence}{Generated-Content Persistence.}
For each annotated pair $(P_{\mathrm{first}},P_{\mathrm{revisit}})$, \textbf{\textit{the first generated observation serves as the reference}} for a revisit at the same pose or an overlapping viewpoint. Regions outside the input view have no ground truth, and many different contents would be plausible. What must hold is that the model shows the same content when it returns, so the model's own earlier output serves as the reference. \textsuperscript{1}\textit{Revisit Appearance Consistency} measures appearance retention across revisits, while \textsuperscript{2}\textit{Revisit Depth Consistency} measures geometric consistency across visits. Both use \textbf{\textit{surfaces visible in both observations}}, matched by projecting first-visit depth into the revisit view with estimated cameras; we compare DINO features for appearance and compute AbsRel for depth. Matching through 3D aligns the same surfaces across visits, so the test also holds when the revisit viewpoint differs. At identical requested poses, we also compare images directly with DINO, PSNR, SSIM, and LPIPS, and depth maps at the shared estimator scale. \textsuperscript{3}\textit{Revisit Surface Agreement} measures whether both visits reconstruct the same 3D surfaces. We compute a bidirectional nearest-point F-score between aligned point clouds, with matching distance relative to scene depth.

\noindent\textcolor{coherence}{\raisebox{0.15ex}{\rule{0.45em}{0.45em}}}\hspace{0.4em}\dimensionlabel{coherence}{3D Self-Consistency.}
We assess whether generated views form \textbf{\textit{a coherent 3D scene}} using cross-view correspondences from recovered cameras. These tests do not use the requested trajectory. A model can follow the requested camera motion and still produce views that disagree in 3D, or produce a coherent scene along the wrong trajectory; control alignment and 3D self-consistency are therefore scored separately. \textsuperscript{1}\textit{Cross-View Depth Consistency} tests whether views agree on the depth of shared surfaces, while \textsuperscript{2}\textit{Cross-View Feature Consistency} tests whether matched surfaces retain consistent visual features across views. Both use valid, unoccluded correspondences established by bidirectional depth reprojection for frame pairs at short and medium temporal intervals, measuring depth AbsRel and DINO cosine similarity, respectively. \textsuperscript{3}\textit{Regional Scale Agreement} tests relative scale consistency across scene regions. We measure dispersion in observed-to-reprojected depth ratios across a $4\times4$ grid. \textsuperscript{4}\textit{Reconstruction} tests \textbf{\textit{reconstruction accuracy on unseen views}}. DA3 predicts a Gaussian scene~\citep{kerbl2023gaussians} from \textbf{\textit{64 context frames}}; we render \textbf{\textit{16 held-out views}} and compare them with the corresponding generated frames using PSNR, SSIM, and LPIPS for pixel-level and perceptual agreement. WorldLens~\citep{liang2026worldlens} likewise uses reconstructed scenes and re-rendering to assess driving-world fidelity. Pairwise tests check nearby frames, whereas held-out rendering tests whether one scene explains views outside its context, revealing inconsistency that accumulates along the trajectory.

\noindent\textcolor{muted}{\raisebox{0.15ex}{\rule{0.45em}{0.45em}}}\hspace{0.4em}\dimensionlabel{muted}{Consistency over Longer Exploration.}
We track \textsuperscript{1}\textit{visual quality}, \textsuperscript{2}\textit{control}, and \textsuperscript{3}\textit{local geometry} in \textbf{\textit{five-second windows}} for both short and minute-long videos. Temporal stability combines the \textbf{\textit{average performance}} with its \textbf{\textit{retention relative to the initial window}} to assess whether a model sustains its capabilities during continued exploration. The final score weights short- and long-video means by \textbf{\textit{30\%}} and \textbf{\textit{70\%}}, respectively.

\noindent We also report control alignment, using camera pose errors and action accuracy, and visual quality, using standard image and video quality measures. The overall score weights control, visual quality, the three consistency dimensions and temporal stability by 25\%, 25\%, 10\% each and 20\%. Appendix~\ref{app:metrics} details these metrics and evaluation procedures.

\end{evaluationparagraphs}

\section{Experiments}
\label{sec:experiments}

We compare \textbf{27} models: \textbf{19} camera- or action-conditioned \textit{world models} and \textbf{eight} \textit{TI2V models}. Adapters map requested trajectories to native camera conditions, discrete commands, or text instructions, starting from the same initial image. Table~\ref{tab:results} reports all six tiers and the overall score. Appendices~\ref{app:adapters} and~\ref{app:extended} provide adapters and component results.

\begin{table}[t]
\centering
\begin{tablebody}{1.5pt}{1.2}

\caption{\textbf{Comparison across six evaluation tiers.} \textbf{Gen. Persist.} denotes generated-content persistence. \textbf{Temp. Stab.} denotes temporal stability. \textbf{3D Repr.} indicates scene geometry used by the generator ($\checkmark$: present, $\times$: absent, --: unconfirmed). Models are sorted by \textbf{Overall} within each group. \ranklegend{orbiaRankFirst}{1st}, \ranklegend{orbiaRankSecond}{2nd}, \ranklegend{orbiaRankThird}{3rd} mark the top three scores per column and group.}\label{tab:results}
\fontsize{7}{8.5}\selectfont
\begin{tabularx}{\linewidth}{@{}lccc*{7}{>{\centering\arraybackslash}X}@{}}
\toprule
\rowcolor{orbiaTableHeader}
\resulthead{Model} & \resulthead{Size~(B)} & \resulthead{Steps} & \resulthead{3D\\Repr.} & \resulthead{Control\\Align.} & \resulthead{Visual\\Quality} & \resulthead{Input\\Preserv.} & \resulthead{Gen.\\Persist.} & \resulthead{3D Self-\\Consist.} & \resulthead{Temp.\\Stab.} & \resulthead{Overall}\\
\midrule
\rowcolor{orbiaTableHeader}
\multicolumn{11}{l}{\textbf{World Models}}\\
\modellogo{cuhksz}SolarWM-H3~\citeyearpar{huang2026solarwm} & $33$ & $4$ & $\times$ & \scoresecond{$\mathbf{79.93}$} & $74.76$ & \scorethird{$\mathbf{69.26}$} & \scorefirst{$\mathbf{70.28}$} & \scorethird{$\mathbf{68.84}$} & \scorefirst{$\mathbf{84.05}$} & \scorefirst{$\mathbf{76.32}$}\\
\rowcolor{orbiaTableStripe}
\modellogo{nvidia}Lyra2~\citeyearpar{shen2026lyra2} & $14$ & $4$ & $\checkmark$ & \scorefirst{$\mathbf{81.64}$} & $71.37$ & $68.79$ & \scoresecond{$\mathbf{70.25}$} & $61.51$ & $82.31$ & \scoresecond{$\mathbf{74.77}$}\\
\modellogo{alaya}EVOKE~\citeyearpar{yin2026evoke} & $14$ & $3$ & $\checkmark$ & $66.47$ & \scorefirst{$\mathbf{76.48}$} & $68.81$ & $66.43$ & $68.02$ & \scoresecond{$\mathbf{83.18}$} & \scorethird{$\mathbf{72.70}$}\\
\rowcolor{orbiaTableStripe}
\modellogo{jd}Echo-WM~\citeyearpar{zhang2026echowm} & $22$ & $4$ & $\times$ & $70.74$ & \scorethird{$\mathbf{75.02}$} & $65.91$ & $61.42$ & $65.26$ & $82.69$ & $72.24$\\
\modellogo{nvidia}Gen3C~\citeyearpar{ren2025gen3c} & $7$ & $35$ & $\checkmark$ & \scorethird{$\mathbf{75.51}$} & $67.55$ & \scorefirst{$\mathbf{75.23}$} & $68.58$ & $60.16$ & $78.96$ & $71.95$\\
\rowcolor{orbiaTableStripe}
\modellogo{skywork}Matrix game 3.5~\citeyearpar{qian2026matrixgame35} & $5$ & $3$ & $\checkmark$ & $64.18$ & $74.03$ & $66.93$ & $60.82$ & $54.61$ & $76.41$ & $68.07$\\
\modellogo{alaya}AlayaWorld~\citeyearpar{team2026alayaworld} & $15$ & $4$ & $\checkmark$ & $61.68$ & $71.04$ & \scoresecond{$\mathbf{70.17}$} & $56.62$ & $58.92$ & $76.31$ & $67.01$\\
\rowcolor{orbiaTableStripe}
\modellogo{deepmind}Genie3~\citeyearpar{parkerholder2025genie3} & -- & -- & -- & $55.26$ & $74.43$ & $61.01$ & $54.41$ & $61.25$ & \scorethird{$\mathbf{83.15}$} & $66.72$\\
\modellogo{robbyant}LingBot-World v2~\citeyearpar{gao2026lingbot2} & $14$ & $4$ & $\times$ & $58.48$ & \scoresecond{$\mathbf{75.34}$} & $61.54$ & $52.00$ & $57.11$ & $77.28$ & $65.98$\\
\rowcolor{orbiaTableStripe}
\modellogo{tencent}HY-WorldPlay~\citeyearpar{sun2025worldplay} & $8$ & $4$ & $\times$ & $45.26$ & $72.06$ & $66.84$ & \scorethird{$\mathbf{69.84}$} & \scorefirst{$\mathbf{70.34}$} & $77.32$ & $65.50$\\
\modellogo{nvidia}SANA-WM~\citeyearpar{zhu2026sanawm} & $2.6$ & $4+3$ & $\times$ & $59.39$ & $73.22$ & $63.19$ & $51.95$ & $54.10$ & $75.03$ & $65.08$\\
\rowcolor{orbiaTableStripe}
\modellogo{alibaba}HappyOyster & -- & -- & -- & $60.49$ & $68.78$ & $58.23$ & $46.31$ & $59.17$ & $78.25$ & $64.34$\\
\modellogo{nvidia}Cosmos3-Nano~\citeyearpar{nvidia2026cosmos3} & $16$ & $30$ & $\times$ & $35.70$ & $74.25$ & $66.78$ & $66.29$ & \scoresecond{$\mathbf{70.10}$} & $74.71$ & $62.75$\\
\rowcolor{orbiaTableStripe}
\modellogo{robbyant}LingBot-World v1~\citeyearpar{gao2026lingbot} & $14$ & $4$ & $\times$ & $49.12$ & $73.84$ & $59.56$ & $50.71$ & $58.93$ & $75.37$ & $62.73$\\
\modellogo{tencent}Voyager~\citeyearpar{huang2025voyager} & $13$ & $50$ & $\checkmark$ & $60.92$ & $68.05$ & $58.65$ & $43.72$ & $46.32$ & $78.04$ & $62.72$\\
\rowcolor{orbiaTableStripe}
\modellogo{amap}DreamX-World~\citeyearpar{team2026dreamx} & $5$ & $4$ & $\times$ & $48.99$ & $74.96$ & $60.96$ & $48.70$ & $48.73$ & $74.82$ & $61.79$\\
\modellogo{amap}FantasyWorld~\citeyearpar{fantasy2025world} & $14$ & $50$ & $\checkmark$ & $39.16$ & $70.70$ & $63.08$ & $60.17$ & $63.53$ & $72.51$ & $60.64$\\
\rowcolor{orbiaTableStripe}
\modellogo{amap}ABot-World~\citeyearpar{jiang2026abotworld} & $5$ & $4$ & $\times$ & $47.06$ & $66.97$ & $57.47$ & $45.22$ & $48.32$ & $76.32$ & $58.87$\\
\modellogo{tsinghua}Astra~\citeyearpar{zhu2025astra} & $1.3$ & $50$ & $\times$ & $28.45$ & $61.30$ & $59.38$ & $44.74$ & $59.91$ & $67.31$ & $52.30$\\
\midrule
\rowcolor{orbiaTableGroupHeader}
\multicolumn{11}{l}{\textbf{TI2V Models}}\\
\modellogo{bytedance}Seedance2.5~\citeyearpar{bytedance2026seedance25} & -- & -- & -- & \scorefirst{$\mathbf{43.23}$} & $76.97$ & \scorefirst{$\mathbf{72.84}$} & \scoresecond{$\mathbf{67.70}$} & \scoresecond{$\mathbf{72.54}$} & \scorefirst{$\mathbf{80.39}$} & \scorefirst{$\mathbf{67.44}$}\\
\rowcolor{orbiaTableGroupStripe}
\modellogo{kuaishou}Kling3.0~\citeyearpar{kling2026video3} & -- & -- & -- & \scoresecond{$\mathbf{38.58}$} & $77.38$ & \scoresecond{$\mathbf{69.14}$} & \scorethird{$\mathbf{63.32}$} & $71.30$ & \scorethird{$\mathbf{79.30}$} & \scoresecond{$\mathbf{65.23}$}\\
\modellogo{nvidia}Cosmos3-Super~\citeyearpar{nvidia2026cosmos3} & $64$ & $4$ & -- & $27.92$ & \scorefirst{$\mathbf{79.04}$} & \scorethird{$\mathbf{66.04}$} & \scorefirst{$\mathbf{69.75}$} & \scorefirst{$\mathbf{74.55}$} & \scoresecond{$\mathbf{79.73}$} & \scorethird{$\mathbf{63.72}$}\\
\rowcolor{orbiaTableGroupStripe}
\modellogo{tencent}HunyuanVideo1.5~\citeyearpar{tencent2025hy15} & $8.3$ & $12$ & $\times$ & \scorethird{$\mathbf{36.09}$} & $77.07$ & $63.12$ & $57.44$ & $71.31$ & $78.96$ & $63.27$\\
\modellogo{minimax}MiniMax-H3~\citeyearpar{minimax2026h3} & $33$ & $50$ & -- & $32.48$ & \scoresecond{$\mathbf{78.09}$} & $63.36$ & $55.79$ & $69.42$ & $77.51$ & $62.00$\\
\rowcolor{orbiaTableGroupStripe}
\modellogo{lightricks}LTX2.5~\citeyearpar{lightricks2026ltx25} & $22$ & $8+3$ & -- & $26.51$ & \scorethird{$\mathbf{77.96}$} & $64.81$ & $60.25$ & \scorethird{$\mathbf{71.64}$} & $77.30$ & $61.25$\\
\modellogo{alibaba}Wan2.7~\citeyearpar{wan2025wan} & -- & -- & -- & $31.31$ & $76.46$ & $65.96$ & $54.35$ & $64.19$ & $75.79$ & $60.55$\\
\rowcolor{orbiaTableGroupStripe}
\modellogo{meituan}LongCat~\citeyearpar{meituan2025longcat} & $13.6$ & $16$ & $\times$ & $23.76$ & $77.30$ & $58.89$ & $47.77$ & $64.11$ & $74.22$ & $57.18$\\
\bottomrule

\end{tabularx}

\end{tablebody}
\end{table}

\begin{figure}[t]
\centering
\includegraphics[width=\linewidth]{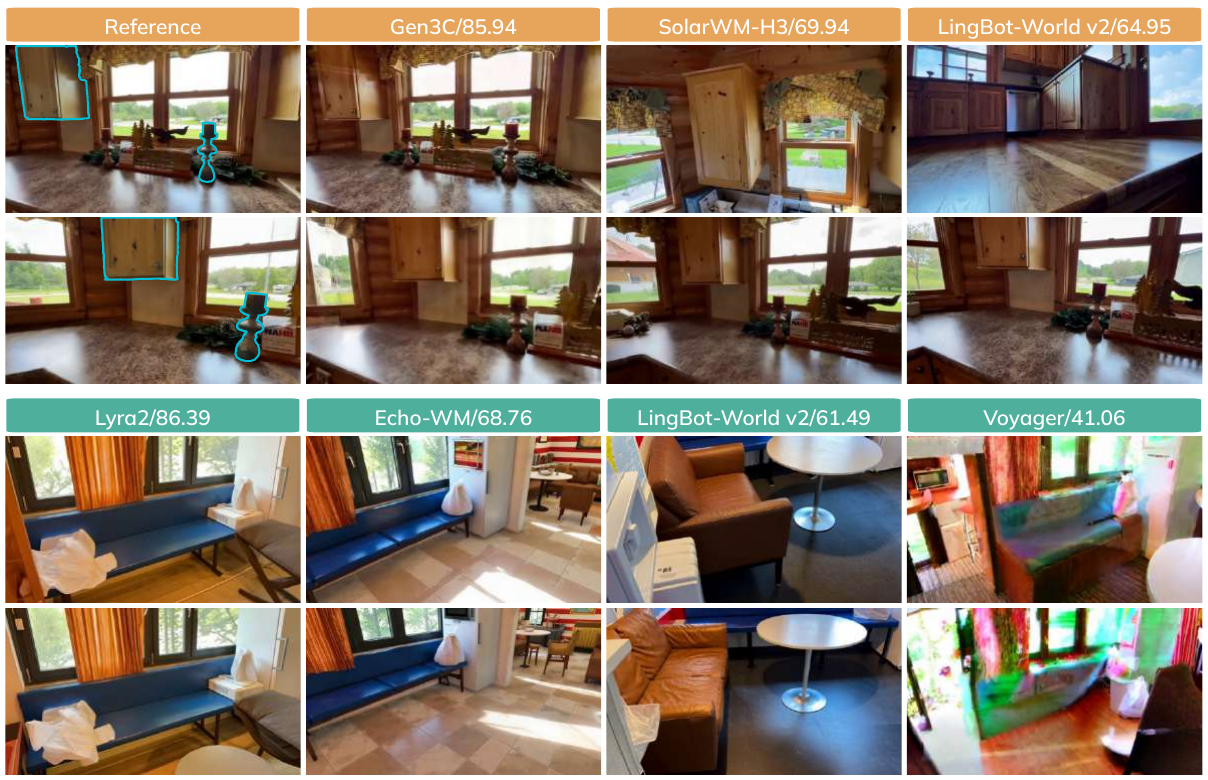}

\caption{\textbf{Input preservation and generated persistence expose different failures.} \textbf{Top two rows:} reference input $P_0$ and query $P_q$ versus model returns and queries. Cyan contours mark \textit{anchors}. \textbf{Bottom two rows:} first visits and revisits on another case. Headers report the evaluated models and their corresponding \textbf{case-level tier scores} for each example.}
\label{fig:main-preservation}

\end{figure}

\subsection{Overall Performance and Capability Gaps}
\noindent\textbf{SolarWM-H3 leads overall.}
SolarWM-H3 ranks first overall at $76.32$, with the strongest generated-content persistence and temporal stability among world models. Lyra2 and EVOKE rank second and third, leading control and visual quality, respectively. Gen3C leads input preservation, while HY-WorldPlay leads 3D self-consistency. Seedance2.5 leads the TI2V group overall, whereas Cosmos3-Super leads its visual quality, generated-content persistence, and 3D self-consistency. No model leads all six tiers.

\noindent\textbf{Several leading world models use explicit 3D memory.}
The 3D Repr.\ column in Table~\ref{tab:results} shows that three of the top five world models use explicit geometry: Lyra2, EVOKE, and Gen3C. Lyra2 uses per-frame geometry to retrieve relevant past frames and establish correspondences with target views~\citep{shen2026lyra2}. EVOKE stores observed geometry in an external world state bank and renders it from the current camera pose~\citep{yin2026evoke}. Gen3C maintains a 3D cache whose renderings guide generation along the requested camera trajectory~\citep{ren2025gen3c}. By linking past observations to 3D locations, these mechanisms let models reuse scene content at a requested viewpoint even after it leaves the recent video context. This may help explain their strong control and content-preservation results.

\noindent\textbf{Camera control is the largest gap between the two model groups.}
World models average $57.29$ in control, compared with $32.48$ for TI2V models, a gap of $24.80$ points. Even the best TI2V model, Seedance2.5 ($43.23$), falls well below Lyra2 ($81.64$). Control also has the largest standard deviation across world models, indicating substantial differences even among models with native camera or action interfaces. These results identify accurate execution of requested camera motion as a major obstacle to using current video generators as interactive worlds.

\noindent\textbf{Leading world models approach TI2V visual quality.}
TI2V models have higher mean visual quality ($77.53$ versus $71.80$), but EVOKE reaches $76.48$, only $1.05$ points below the TI2V mean. The best models in the two groups differ by $2.56$ points. Within the TI2V group, the standard deviation is just $0.75$, the smallest of its six tiers. Under the current metrics, quality distinguishes TI2V models less clearly than control and consistency.

\begin{table}[t]
\centering
\begin{tablebody}{4pt}{1.2}

\caption{\textbf{Tier 3: Detailed input preservation results.} Scores use a 0--100 scale ($\uparrow$). Models are ordered by \textbf{Tier Score} within each group. \ranklegend{orbiaRankFirst}{1st}, \ranklegend{orbiaRankSecond}{2nd}, \ranklegend{orbiaRankThird}{3rd} mark the top three scores per column and group.}
\label{tab:tier3-components}
\fontsize{7.5}{9}\selectfont
\begin{tabularx}{\linewidth}{@{}lc*{4}{>{\centering\arraybackslash}X}@{}}
\toprule
\rowcolor{orbiaTableHeader}
\textbf{Model} & \textbf{3D Repr.} & \textbf{Return} & \textbf{Query} & \textbf{Identity} & \textbf{Tier Score}\\
\midrule
\rowcolor{orbiaTableHeader}
\multicolumn{6}{l}{\textbf{World Models}}\\
\modellogo{nvidia}Gen3C & $\checkmark$ & \scorefirst{$\mathbf{81.24}$} & \scoresecond{$\mathbf{63.93}$} & $84.28$ & \scorefirst{$\mathbf{75.23}$}\\
\rowcolor{orbiaTableStripe}
\modellogo{alaya}AlayaWorld & $\checkmark$ & \scoresecond{$\mathbf{66.56}$} & $60.47$ & $86.71$ & \scoresecond{$\mathbf{70.17}$}\\
\modellogo{cuhksz}SolarWM-H3 & $\times$ & $54.73$ & \scorefirst{$\mathbf{65.30}$} & $89.06$ & \scorethird{$\mathbf{69.26}$}\\
\rowcolor{orbiaTableStripe}
\modellogo{alaya}EVOKE & $\checkmark$ & $57.72$ & $61.79$ & \scorethird{$\mathbf{89.25}$} & $68.81$\\
\modellogo{nvidia}Lyra2 & $\checkmark$ & \scorethird{$\mathbf{60.33}$} & $62.82$ & $85.19$ & $68.79$\\
\rowcolor{orbiaTableStripe}
\modellogo{skywork}Matrix game 3.5 & $\checkmark$ & $52.40$ & \scorethird{$\mathbf{62.86}$} & $86.89$ & $66.93$\\
\modellogo{tencent}HY-WorldPlay & $\times$ & $57.26$ & $55.88$ & \scoresecond{$\mathbf{91.04}$} & $66.84$\\
\rowcolor{orbiaTableStripe}
\modellogo{nvidia}Cosmos3-Nano & $\times$ & $54.91$ & $56.17$ & \scorefirst{$\mathbf{92.79}$} & $66.78$\\
\modellogo{jd}Echo-WM & $\times$ & $50.07$ & $62.32$ & $86.55$ & $65.91$\\
\rowcolor{orbiaTableStripe}
\modellogo{nvidia}SANA-WM & $\times$ & $47.17$ & $60.44$ & $82.89$ & $63.19$\\
\modellogo{amap}FantasyWorld & $\checkmark$ & $45.98$ & $56.64$ & $88.77$ & $63.08$\\
\rowcolor{orbiaTableStripe}
\modellogo{robbyant}LingBot-World v2 & $\times$ & $41.69$ & $56.85$ & $87.65$ & $61.54$\\
\modellogo{deepmind}Genie3 & -- & $45.53$ & $53.91$ & $85.98$ & $61.01$\\
\rowcolor{orbiaTableStripe}
\modellogo{amap}DreamX-World-5B & $\times$ & $42.54$ & $54.22$ & $88.34$ & $60.96$\\
\modellogo{robbyant}LingBot-World v1 & $\times$ & $42.92$ & $52.75$ & $85.27$ & $59.56$\\
\rowcolor{orbiaTableStripe}
\modellogo{tsinghua}Astra & $\times$ & $41.31$ & $52.92$ & $86.06$ & $59.38$\\
\modellogo{tencent}Voyager & $\checkmark$ & $40.07$ & $57.25$ & $79.09$ & $58.65$\\
\rowcolor{orbiaTableStripe}
\modellogo{alibaba}HappyOyster & -- & $42.64$ & $47.58$ & $88.02$ & $58.23$\\
\modellogo{amap}ABot-World & $\times$ & $40.11$ & $50.88$ & $83.61$ & $57.47$\\
\midrule
\rowcolor{orbiaTableGroupHeader}
\multicolumn{6}{l}{\textbf{TI2V Models}}\\
\modellogo{bytedance}Seedance2.5 & -- & \scorefirst{$\mathbf{74.16}$} & \scorefirst{$\mathbf{57.38}$} & $92.14$ & \scorefirst{$\mathbf{72.84}$}\\
\rowcolor{orbiaTableGroupStripe}
\modellogo{kuaishou}Kling3.0 & -- & \scoresecond{$\mathbf{63.64}$} & \scoresecond{$\mathbf{54.86}$} & \scorefirst{$\mathbf{93.69}$} & \scoresecond{$\mathbf{69.14}$}\\
\modellogo{nvidia}Cosmos3-Super & -- & $54.13$ & \scorethird{$\mathbf{54.43}$} & \scoresecond{$\mathbf{93.44}$} & \scorethird{$\mathbf{66.04}$}\\
\rowcolor{orbiaTableGroupStripe}
\modellogo{alibaba}Wan2.7 & -- & \scorethird{$\mathbf{54.58}$} & $54.09$ & \scorethird{$\mathbf{93.16}$} & $65.96$\\
\modellogo{lightricks}LTX2.5 & -- & $52.47$ & $53.50$ & $92.24$ & $64.81$\\
\rowcolor{orbiaTableGroupStripe}
\modellogo{minimax}MiniMax-H3 & -- & $48.04$ & $54.04$ & $91.11$ & $63.36$\\
\modellogo{tencent}HunyuanVideo1.5 & $\times$ & $45.69$ & $54.08$ & $92.60$ & $63.12$\\
\rowcolor{orbiaTableGroupStripe}
\modellogo{meituan}LongCat & $\times$ & $40.55$ & $51.09$ & $87.64$ & $58.89$\\
\bottomrule
\end{tabularx}
\end{tablebody}
\end{table}

\begin{table}[t]
\centering
\begin{tablebody}{2pt}{1.2}

\caption{\textbf{Tier 4: Detailed generated-content persistence results.} Scores use a 0--100 scale ($\uparrow$). Models are ordered by \textbf{Tier Score} within each group. \ranklegend{orbiaRankFirst}{1st}, \ranklegend{orbiaRankSecond}{2nd}, \ranklegend{orbiaRankThird}{3rd} mark the top three scores per column and group. \textbf{Same} and \textbf{Diff.} denote direct comparisons at identical and different requested poses. \textbf{Reproj.} denotes geometry-aligned comparison.}
\label{tab:tier4-components}
\fontsize{7.5}{9}\selectfont
\begin{tabularx}{\linewidth}{@{}lc*{9}{>{\centering\arraybackslash}X}@{}}
\toprule
\rowcolor{orbiaTableHeader}
 & \textbf{3D} & \multicolumn{5}{c}{\textbf{Revisit Appearance}} & \multicolumn{2}{c}{\componenthead{Revisit Depth}} & \componenthead{Revisit\\Surface} &\textbf{Tier} \\
\cmidrule(lr){3-7}\cmidrule(lr){8-9}\cmidrule(lr){10-10}
\rowcolor{orbiaTableHeader}
\textbf{Model} & \textbf{Repr.} & \componenthead{Same\\PSNR} & \componenthead{Same\\LPIPS} & \componenthead{Same\\DINO} & \componenthead{Diff.\\DINO} & \componenthead{Reproj.\\DINO} & \componenthead{Same\\AbsRel} & \componenthead{Reproj.\\AbsRel} & \componenthead{F-score} & \textbf{Score}\\
\midrule
\rowcolor{orbiaTableHeader}
\multicolumn{11}{l}{\textbf{World Models}}\\
\modellogo{cuhksz}SolarWM-H3 & $\times$ & $43.68$ & \scorethird{$\mathbf{47.45}$} & $62.88$ & \scorefirst{$\mathbf{87.33}$} & \scoresecond{$\mathbf{62.59}$} & \scoresecond{$\mathbf{82.40}$} & \scoresecond{$\mathbf{80.76}$} & \scorefirst{$\mathbf{59.54}$} & \scorefirst{$\mathbf{70.28}$}\\
\rowcolor{orbiaTableStripe}
\modellogo{nvidia}Lyra2 & $\checkmark$ & \scoresecond{$\mathbf{53.88}$} & \scorefirst{$\mathbf{54.30}$} & \scorefirst{$\mathbf{67.72}$} & $82.78$ & \scorefirst{$\mathbf{64.44}$} & $80.23$ & \scorethird{$\mathbf{78.73}$} & \scorethird{$\mathbf{58.12}$} & \scoresecond{$\mathbf{70.25}$}\\
\modellogo{tencent}HY-WorldPlay & $\times$ & $44.18$ & $44.82$ & $59.34$ & \scoresecond{$\mathbf{85.48}$} & \scorethird{$\mathbf{62.06}$} & \scorethird{$\mathbf{82.13}$} & \scorefirst{$\mathbf{84.24}$} & \scoresecond{$\mathbf{59.06}$} & \scorethird{$\mathbf{69.84}$}\\
\rowcolor{orbiaTableStripe}
\modellogo{nvidia}Gen3C & $\checkmark$ & \scorefirst{$\mathbf{55.64}$} & \scoresecond{$\mathbf{51.52}$} & \scoresecond{$\mathbf{64.19}$} & $81.86$ & $61.85$ & \scorefirst{$\mathbf{84.91}$} & $76.98$ & $53.12$ & $68.58$\\
\modellogo{alaya}EVOKE & $\checkmark$ & $47.53$ & $44.61$ & \scorethird{$\mathbf{63.19}$} & \scorethird{$\mathbf{84.12}$} & $59.83$ & $81.98$ & $74.36$ & $49.14$ & $66.43$\\
\rowcolor{orbiaTableStripe}
\modellogo{nvidia}Cosmos3-Nano & $\times$ & \scorethird{$\mathbf{53.56}$} & $42.64$ & $56.92$ & $81.25$ & $58.82$ & $81.11$ & $77.31$ & $52.85$ & $66.29$\\
\modellogo{jd}Echo-WM & $\times$ & $37.11$ & $34.83$ & $53.10$ & $82.07$ & $55.65$ & $76.78$ & $70.25$ & $45.20$ & $61.42$\\
\rowcolor{orbiaTableStripe}
\modellogo{skywork}Matrix game 3.5 & $\checkmark$ & $41.68$ & $42.79$ & $59.46$ & $78.42$ & $58.07$ & $74.27$ & $64.94$ & $42.67$ & $60.82$\\
\modellogo{amap}FantasyWorld & $\checkmark$ & $34.42$ & $32.52$ & $52.29$ & $78.65$ & $54.90$ & $77.16$ & $71.66$ & $42.84$ & $60.17$\\
\rowcolor{orbiaTableStripe}
\modellogo{alaya}AlayaWorld & $\checkmark$ & $34.91$ & $30.74$ & $53.41$ & $73.74$ & $53.54$ & $70.07$ & $63.97$ & $40.42$ & $56.62$\\
\modellogo{deepmind}Genie3 & -- & $36.73$ & $22.95$ & $47.16$ & $71.63$ & $47.68$ & $63.50$ & $75.41$ & $35.00$ & $54.41$\\
\rowcolor{orbiaTableStripe}
\modellogo{robbyant}LingBot-World v2 & $\times$ & $22.34$ & $16.70$ & $41.38$ & $73.30$ & $47.35$ & $66.59$ & $66.51$ & $34.38$ & $52.00$\\
\modellogo{nvidia}SANA-WM & $\times$ & $35.27$ & $24.40$ & $44.74$ & $76.55$ & $48.62$ & $71.64$ & $54.43$ & $29.63$ & $51.95$\\
\rowcolor{orbiaTableStripe}
\modellogo{robbyant}LingBot-World v1 & $\times$ & $20.83$ & $16.71$ & $41.39$ & $68.48$ & $46.90$ & $61.95$ & $68.50$ & $33.99$ & $50.71$\\
\modellogo{amap}DreamX-World-5B & $\times$ & $31.04$ & $22.14$ & $44.26$ & $76.47$ & $46.63$ & $62.31$ & $50.91$ & $23.70$ & $48.70$\\
\rowcolor{orbiaTableStripe}
\modellogo{alibaba}HappyOyster & -- & $42.68$ & $17.64$ & $39.55$ & $64.13$ & $42.51$ & $46.99$ & $64.42$ & $23.23$ & $46.31$\\
\modellogo{amap}ABot-World & $\times$ & $46.29$ & $20.84$ & $42.32$ & $57.58$ & $44.37$ & $63.19$ & $53.52$ & $21.12$ & $45.22$\\
\rowcolor{orbiaTableStripe}
\modellogo{tsinghua}Astra & $\times$ & $36.99$ & $14.19$ & $37.64$ & $54.08$ & $39.27$ & $74.58$ & $62.49$ & $19.26$ & $44.74$\\
\modellogo{tencent}Voyager & $\checkmark$ & $13.52$ & $14.44$ & $43.39$ & $61.92$ & $41.94$ & $61.93$ & $49.34$ & $25.61$ & $43.72$\\
\midrule
\rowcolor{orbiaTableGroupHeader}
\multicolumn{11}{l}{\textbf{TI2V Models}}\\
\modellogo{nvidia}Cosmos3-Super & -- & \scorethird{$\mathbf{36.85}$} & \scorefirst{$\mathbf{41.66}$} & \scorefirst{$\mathbf{60.33}$} & \scorefirst{$\mathbf{88.60}$} & \scorefirst{$\mathbf{61.41}$} & \scorefirst{$\mathbf{85.15}$} & \scoresecond{$\mathbf{82.96}$} & \scorefirst{$\mathbf{57.57}$} & \scorefirst{$\mathbf{69.75}$}\\
\rowcolor{orbiaTableGroupStripe}
\modellogo{bytedance}Seedance2.5 & -- & \scorefirst{$\mathbf{40.41}$} & \scoresecond{$\mathbf{37.59}$} & \scoresecond{$\mathbf{57.08}$} & \scoresecond{$\mathbf{86.79}$} & \scoresecond{$\mathbf{57.92}$} & \scoresecond{$\mathbf{79.65}$} & \scorefirst{$\mathbf{83.76}$} & \scoresecond{$\mathbf{55.56}$} & \scoresecond{$\mathbf{67.70}$}\\
\modellogo{kuaishou}Kling3.0 & -- & $34.59$ & \scorethird{$\mathbf{32.08}$} & \scorethird{$\mathbf{52.33}$} & \scorethird{$\mathbf{82.83}$} & \scorethird{$\mathbf{53.45}$} & $77.90$ & \scorethird{$\mathbf{82.64}$} & \scorethird{$\mathbf{46.89}$} & \scorethird{$\mathbf{63.32}$}\\
\rowcolor{orbiaTableGroupStripe}
\modellogo{lightricks}LTX2.5 & -- & \scoresecond{$\mathbf{37.47}$} & $29.93$ & $51.78$ & $78.80$ & $50.16$ & \scorethird{$\mathbf{78.43}$} & $76.91$ & $42.05$ & $60.25$\\
\modellogo{tencent}HunyuanVideo1.5 & $\times$ & $19.81$ & $20.71$ & $48.34$ & $80.75$ & $51.06$ & $69.26$ & $73.71$ & $40.99$ & $57.44$\\
\rowcolor{orbiaTableGroupStripe}
\modellogo{minimax}MiniMax-H3 & -- & $19.68$ & $17.32$ & $43.91$ & $77.27$ & $46.31$ & $70.33$ & $79.04$ & $37.73$ & $55.79$\\
\modellogo{alibaba}Wan2.7 & -- & $20.18$ & $21.02$ & $45.81$ & $74.97$ & $48.09$ & $73.87$ & $70.31$ & $34.15$ & $54.35$\\
\rowcolor{orbiaTableGroupStripe}
\modellogo{meituan}LongCat & $\times$ & $18.27$ & $14.95$ & $38.19$ & $67.91$ & $42.38$ & $65.31$ & $66.08$ & $25.23$ & $47.77$\\
\bottomrule
\end{tabularx}
\end{tablebody}
\end{table}

\noindent\textbf{TI2V models generate strong 3D self-consistency despite weak control.}
TI2V models average $69.88$ in 3D self-consistency, exceeding world models by $10.14$ points. Cosmos3-Super ($74.55$) also surpasses the best world model, HY-WorldPlay ($70.34$). Its strong depth, cross-view feature, and regional-scale scores show that this advantage extends beyond reconstruction fidelity (Table~\ref{tab:tier5-components}). Thus, the TI2V control gap does not stem simply from an inability to generate internally consistent scenes. The challenge is to make those scenes unfold along the requested camera trajectory.

\noindent\textbf{Models differ substantially in generated-content persistence.}
The group means are close for both input preservation ($64.30$ for world models and $65.52$ for TI2V) and generated-content persistence ($57.29$ and $59.55$). Gen3C leads input preservation at $75.23$, while SolarWM-H3, Lyra2, and HY-WorldPlay match or exceed the best TI2V persistence score. Within each group, however, generated-content persistence varies more than input preservation: their standard deviations are $9.22$ versus $4.79$ for world models and $6.80$ versus $3.91$ for TI2V. \textbf{Models differ more in their ability to maintain newly generated content than to preserve the input scene.}

\noindent\textbf{Leading world models match or exceed TI2V temporal stability.}
SolarWM-H3, EVOKE, and Echo-WM score $84.05$, $83.18$, and $82.69$ in temporal stability, respectively, all above the best TI2V score of $80.39$ from Seedance2.5. These world models generate autoregressively in causal chunks~\citep{huang2026solarwm,yin2026evoke,zhang2026echowm}, whereas TI2V generators such as LTX2.5 use bidirectional temporal attention within each clip~\citep{lightricks2026ltx25}. These causal models already match or exceed the temporal stability of bidirectional TI2V generators.

\subsection{Complementarity of 3D Consistency Metrics}

\noindent\textbf{Input preservation measures fidelity to the input scene.}
Gen3C leads input preservation and Return. Figure~\ref{fig:main-preservation} shows its advantage in both return and query views. In the first row, Gen3C's return view most closely matches the input reference, while SolarWM-H3 and LingBot-World v2 show substantial forgetting. In the second row, Gen3C preserves both the appearance and spatial placement of key anchors, including the cabinet and lamp, closely matching the query reference, while other models place these anchors in incorrect relative positions.

\begin{figure}[t]
\centering
\includegraphics[width=\linewidth]{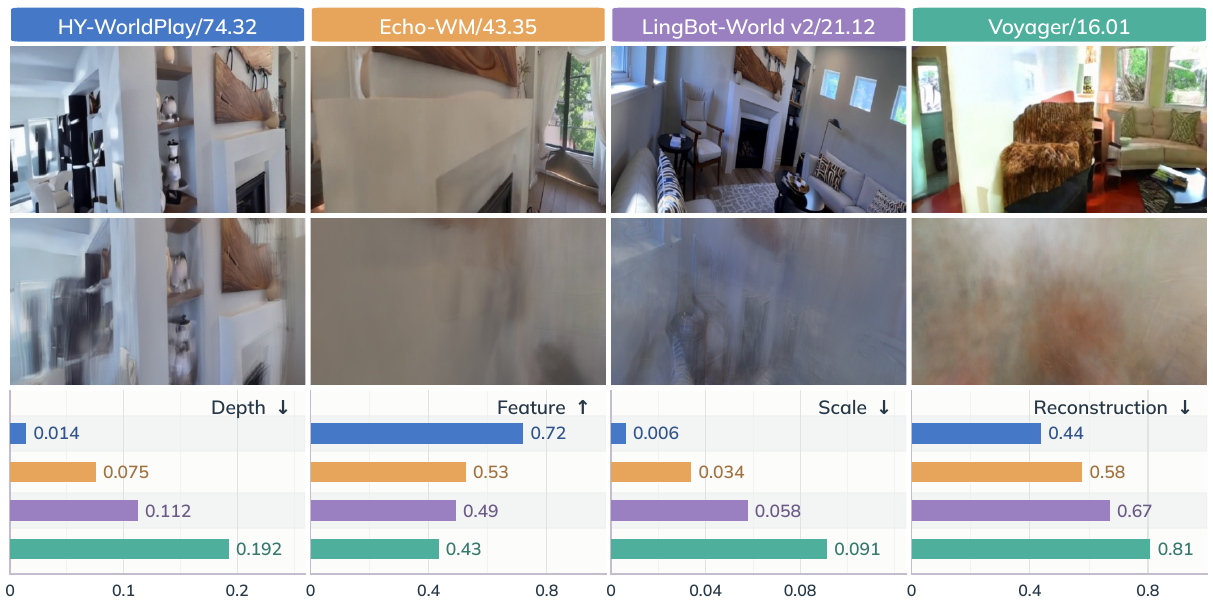}

\caption{\textbf{3D self-consistency reveals reconstruction failures.} \textbf{Top two rows:} held-out generated views and their Gaussian reconstructions. \textbf{Bottom:} \textit{depth, feature, scale, and reconstruction metrics}. Header and bar colors identify models. Header values summarize the \textbf{case-level tier scores} for the corresponding evaluated videos.}

\label{fig:main-geometry}
\end{figure}

\noindent\textbf{Input preservation requires complementary checks.}
SolarWM-H3 leads Query, while Cosmos3-Nano leads Identity (Table~\ref{tab:tier3-components}). Gen3C instead leads Return, showing that the three components favor different models. Query checks appearance and geometry at a new viewpoint, Identity tracks anchors during the input-to-query interval, and Return tests recovery of the initial scene after the full exploration trajectory. The three checks jointly assess input preservation during exploration and after return, covering failures that any one component alone would miss.

\noindent\textbf{Semantic similarity alone does not ensure visual fidelity.}
High semantic similarity between a first visit and revisit can coexist with poor perceptual fidelity (Figure~\ref{fig:main-preservation}, bottom two rows). MiniMax-H3, for example, scores $77.27$ on different-view DINO but only $17.32$ on normalized same-view LPIPS (Table~\ref{tab:tier4-components}). A revisit can match the scene's general content while changing textures and object appearance. Our tests therefore combine semantic, perceptual, and pixel-level comparisons to check both whether the same content reappears and whether its visual details are preserved.

\noindent\textbf{Appearance consistency does not establish spatial agreement.}
Lyra2 leads several revisit-appearance measures, whereas HY-WorldPlay leads reprojected depth consistency and SolarWM-H3 leads surface agreement (Table~\ref{tab:tier4-components}). The appearance tests compare how the revisited content looks; the depth and surface tests check whether it occupies the same 3D locations. Preserving visual details and preserving the 3D layout therefore require separate checks.

\noindent\textbf{Local cross-view agreement does not ensure coherent scene reconstruction.}
FantasyWorld ranks second among world models in cross-view feature consistency at both temporal intervals, but second to last in reconstruction PSNR (Appendix~\ref{app:tier5-results}). HY-WorldPlay likewise leads depth consistency and regional scale agreement without achieving the best reconstruction score. These ranking differences show that strong agreement between pairs of views does not necessarily mean that the generated views can be reconstructed into a coherent 3D scene. Reconstruction consequently provides a separate test of whether the video depicts one coherent scene throughout exploration.

\subsection{Relationships among Quality, Control, and Consistency}
\label{sec:analysis-cross-tier}
\begin{figure}[t]
\centering
\includegraphics[width=\linewidth]{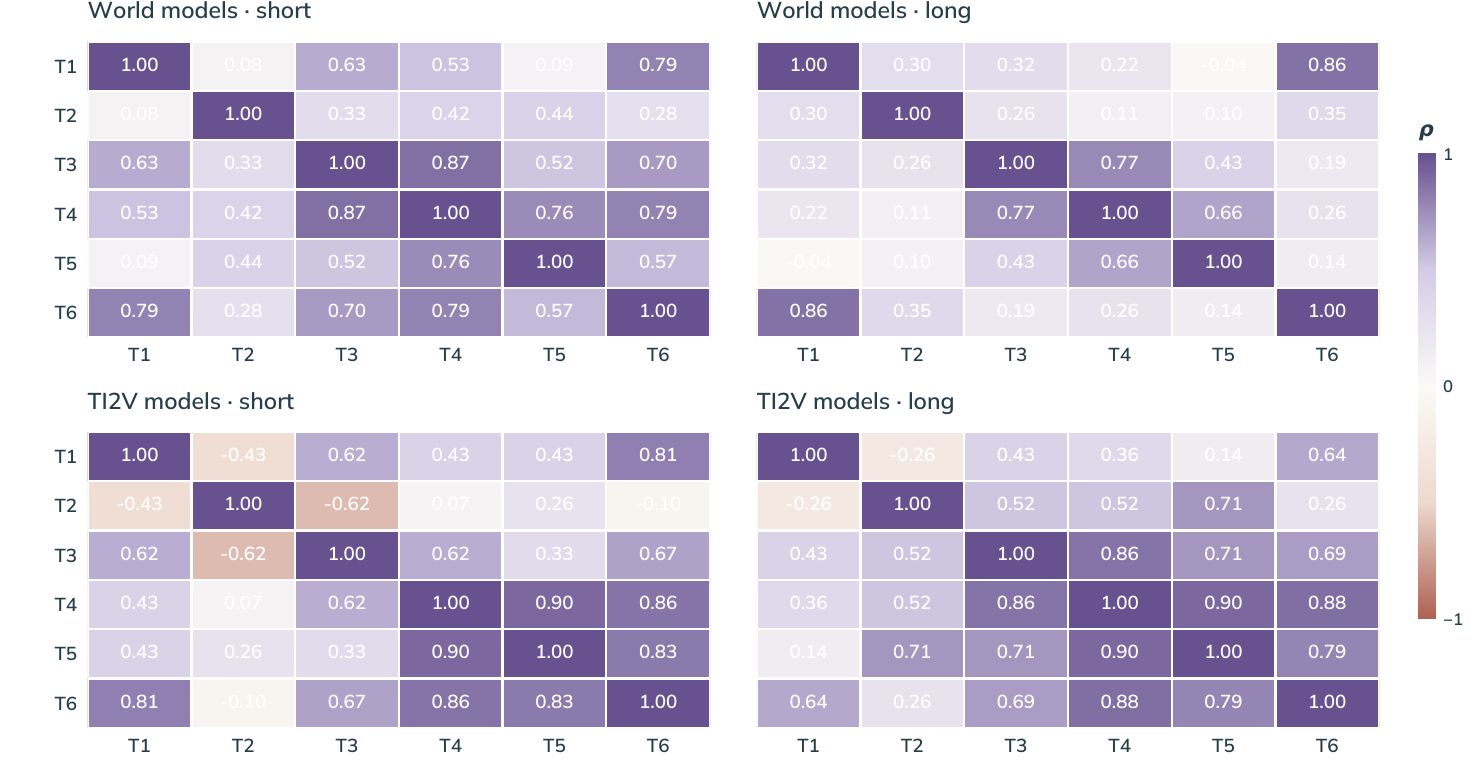}
\caption{\textbf{Cross-tier ranking correlations.} \textbf{Top:} \textit{world models}. \textbf{Bottom:} \textit{TI2V models}. \textbf{Left:} \textit{short videos}. \textbf{Right:} \textit{long videos}. Cells show \textbf{Spearman correlations} between model-level tier scores.}
\label{fig:analysis-correlations-short}
\label{fig:analysis-correlations-long}
\end{figure}

We examine Spearman correlations between tier scores. Figure~\ref{fig:analysis-correlations-short} reports model-level correlations separately for short and long videos within each model group.

\noindent\textbf{Visual quality does not ensure 3D consistency.}
LingBot-World v2 ranks second among world models in visual quality, but twelfth in both input preservation and generated-content persistence, and fourteenth in 3D self-consistency. Gen3C shows the reverse: seventeenth in visual quality but first in input preservation. Across all models, visual quality has a rank correlation of only $0.33$ with the mean of the three consistency tiers among world models and $0.21$ among TI2V models. Figure~\ref{fig:main-geometry} illustrates this discrepancy: LingBot-World v2 generates a detailed view but yields a less faithful reconstruction and weaker depth agreement than HY-WorldPlay. Visual quality alone therefore gives little assurance that a model preserves scene content or maintains a consistent 3D scene. This separation is consistent with WorldLens~\citep{liang2026worldlens}, which reports distinct strengths in visual fidelity, geometry, and downstream utility for driving world models.

\noindent\textbf{Strong visual quality does not guarantee faithful execution of the requested scene and camera motion.}
Quality and control have weak rank agreement among world models ($0.08$ for short videos and $0.30$ for long videos), and negative correlations among TI2V models ($-0.43$ and $-0.26$). Short-video TI2V quality also correlates negatively with input preservation ($-0.62$). In Table~\ref{tab:results}, Cosmos3-Super leads quality but ranks sixth in control, while Seedance2.5 leads input preservation but ranks seventh in quality. Higher visual quality does not imply more accurate camera motion, and in short TI2V videos it can coincide with poorer input preservation.

\noindent\textbf{Control and 3D self-consistency measure different capabilities.}
Lyra2 ranks first in control but seventh in 3D self-consistency, while HY-WorldPlay ranks sixteenth and first, respectively. Across world models, the rank correlation between the two tiers is only $0.09$ in short videos and $-0.04$ in long videos (Figure~\ref{fig:analysis-correlations-short}). A model can generate views that agree with one another even when the camera follows the wrong path. Internal geometric agreement and accurate execution of camera commands are thus separate requirements for interactive world generation.

\noindent\textbf{Preserving the input does not ensure preserving newly generated content.}
World-model input preservation and generated-content persistence have strongly aligned model rankings, but much weaker agreement across cases within each model. For short videos, the model-level correlation is $0.87$, compared with a mean within-model correlation of $0.35$; the same pattern holds for long videos. Strong models tend to perform well on both tasks, yet a successful input-preservation case can still lose newly generated content at revisit. A model's ability to recover the input scene therefore does not reliably predict whether it will preserve the regions it creates during exploration.

\noindent\textbf{Following the camera path is not enough to preserve content during long exploration.}
Among world models, the correlation between control and input preservation falls from $0.63$ in short videos to $0.32$ in long videos; for generated-content persistence, it falls from $0.53$ to $0.22$. Longer exploration requires models to preserve scene content across additional movements and revisits, even when they follow the requested camera path.

\subsection{Control and Consistency across Trajectory Templates}
\label{sec:analysis-template}
\begin{figure}[t]
\centering
\includegraphics[width=\linewidth]{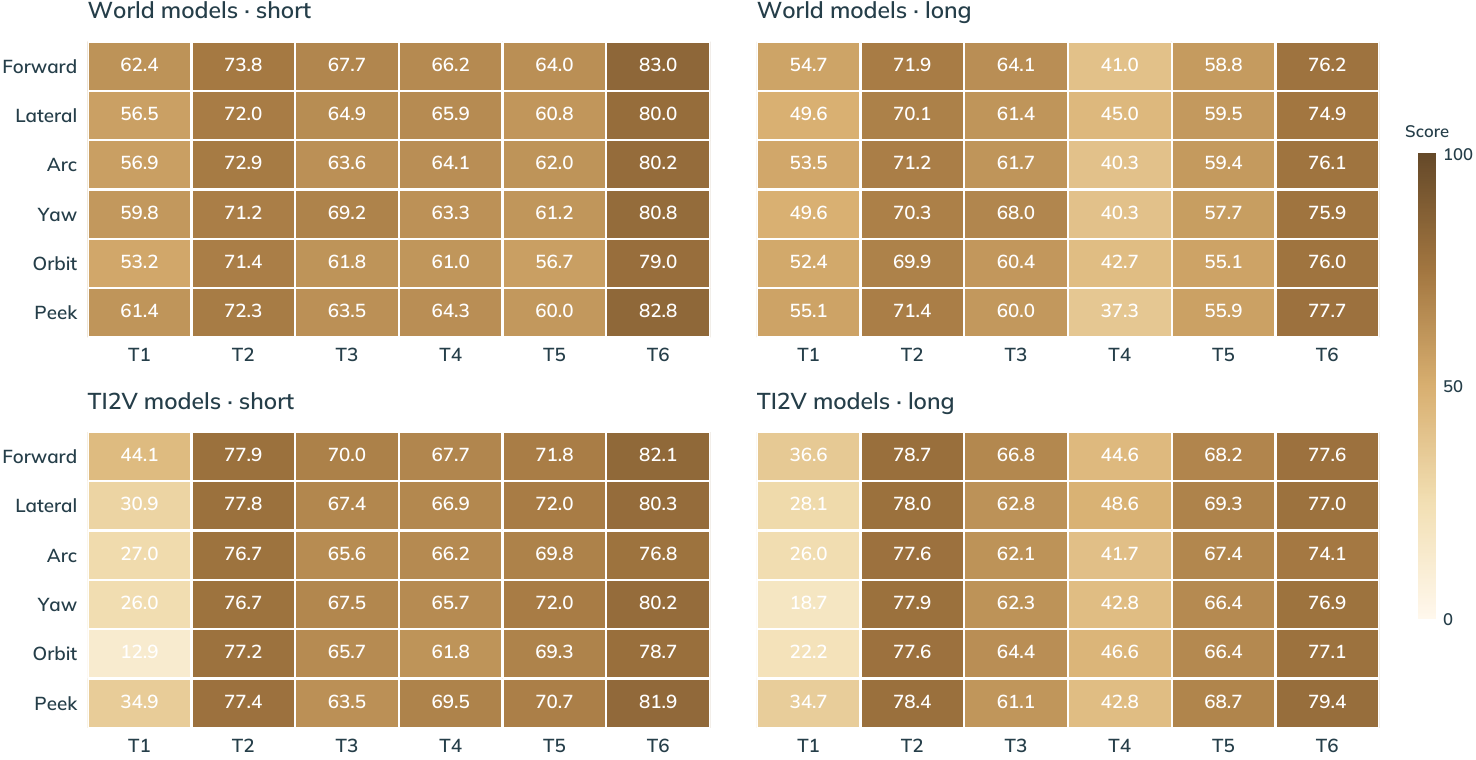}
\caption{\textbf{Performance across trajectory templates.} \textbf{Top:} \textit{world models}. \textbf{Bottom:} \textit{TI2V models}. \textbf{Left:} \textit{short videos}. \textbf{Right:} \textit{long videos}. Cells show \textbf{mean tier scores} with equal model weights within each evaluated model group.}
\label{fig:analysis-templates}
\end{figure}

Figure~\ref{fig:analysis-templates} compares all six trajectory templates using equal model weights within each group. The templates range from reversing a translation to circling an anchor or revealing an occluded region. These comparisons show which camera motions expose failures that are less apparent on a simple forward--backward path.

\begin{figure}[t]
\centering
\includegraphics[width=\linewidth]{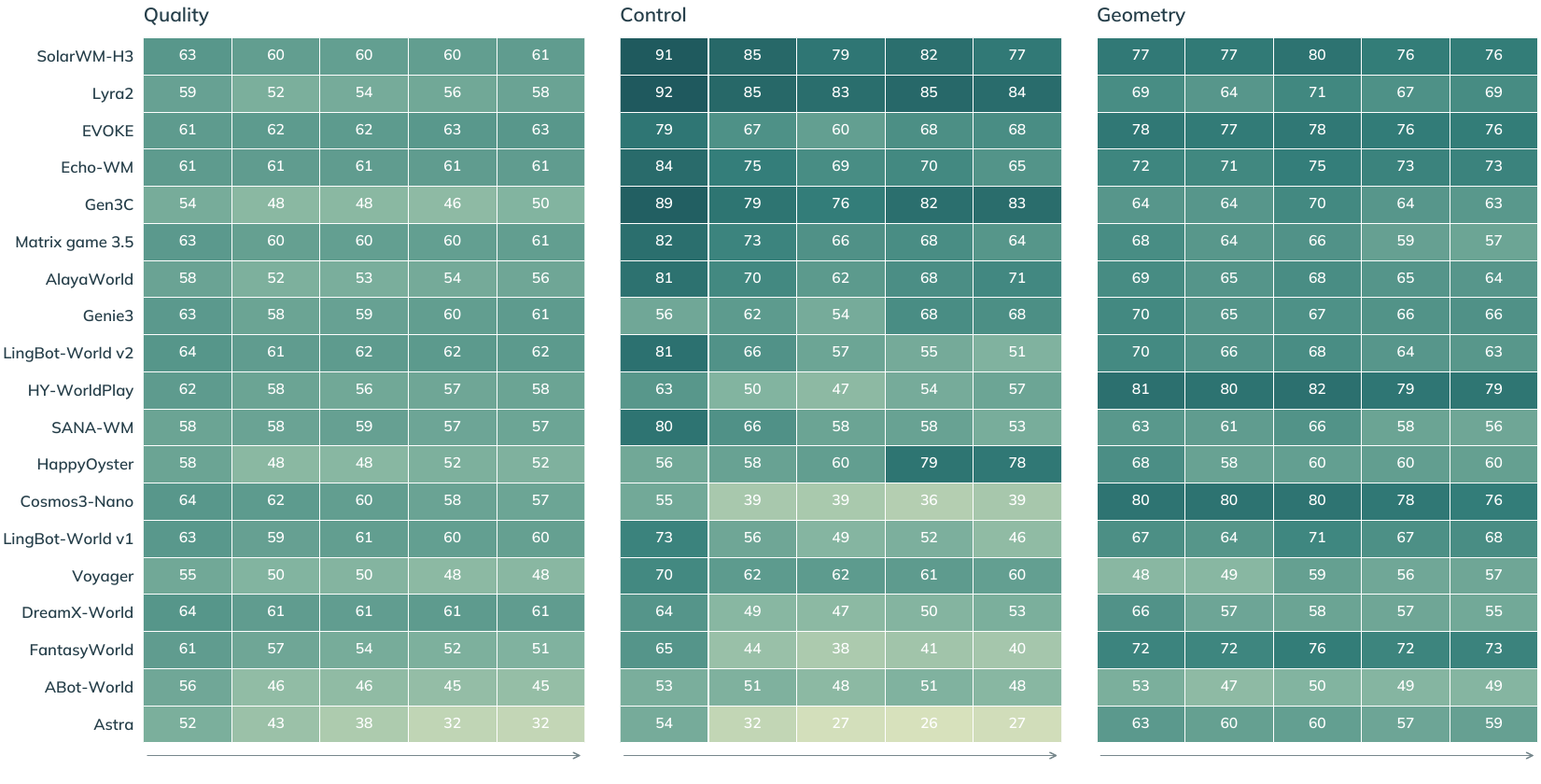}
\par
\includegraphics[width=\linewidth]{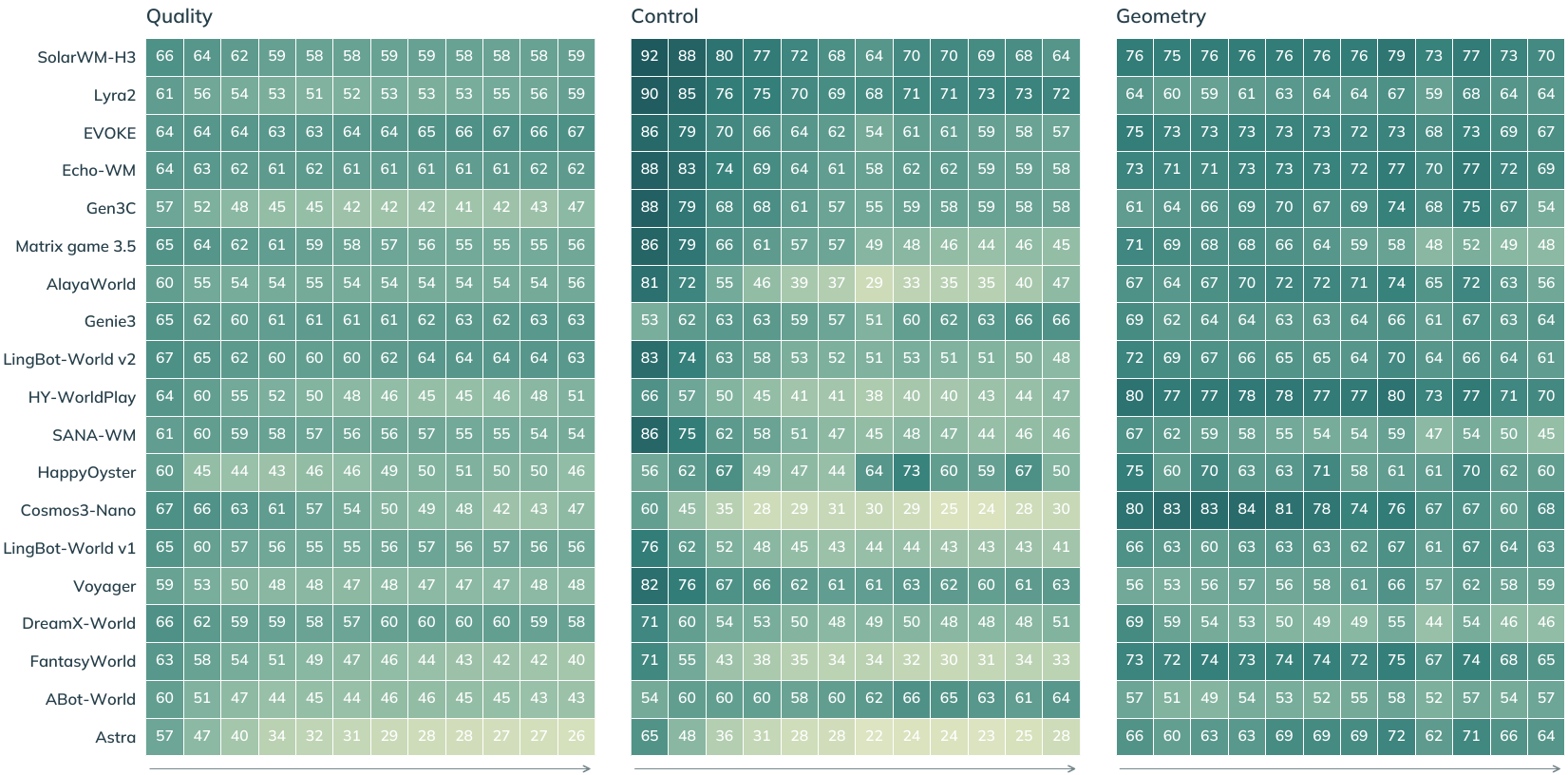}
\caption{\textbf{Temporal performance for world models.} \textbf{Top:} \textit{short videos}. \textbf{Bottom:} \textit{long videos}. Rows follow the model order in Table~\ref{tab:results}. \textit{Time increases from left to right}. Cells show mean window scores for \textbf{quality}, \textbf{control}, and \textbf{local geometry} on a shared 0--100 scale, rounded to integers. Darker colors indicate higher scores within each evaluated time window.}
\label{fig:analysis-time-world-absolute}
\end{figure}

\noindent\textbf{TI2V models fall behind in camera control on every trajectory template.}
TI2V models score below world models in control on all six templates, for both short and long videos. On short forward--backward trajectories, their mean control is $44.05$, compared with $62.41$ for world models. On orbit trajectories, the gap widens to $12.95$ versus $53.15$. Meanwhile, TI2V visual quality remains within $76.70$--$77.94$ across the short-video templates. These models can produce good-looking frames across camera motions, but translating a text instruction into the specified trajectory remains difficult, especially when position and viewing direction must change together.

\noindent\textbf{Orbit motion challenges both camera control and scene preservation.}
For short world-model videos, orbit has the lowest mean control, input-preservation, and generated-content-persistence scores among the six templates. Control falls from $62.41$ on forward--backward motion to $53.15$ on orbit, while input preservation and generated-content persistence reach $61.83$ and $60.97$. Circling an anchor repeatedly changes visible surfaces and occlusions. Simple reversal tests therefore underestimate the challenge of preserving a scene across different viewpoints.

\noindent\textbf{Complex trajectories can still end in forgetting despite stronger control.}
For long world-model videos, peek-and-return has the highest control score ($55.09$) but the lowest generated-content persistence ($37.32$). Models can therefore follow this camera motion yet fail to preserve the content revealed during the peek. Preserving newly revealed content through changes in visibility remains difficult even when camera control is comparatively strong.

\begin{figure}[t]
\centering
\includegraphics[width=\linewidth]{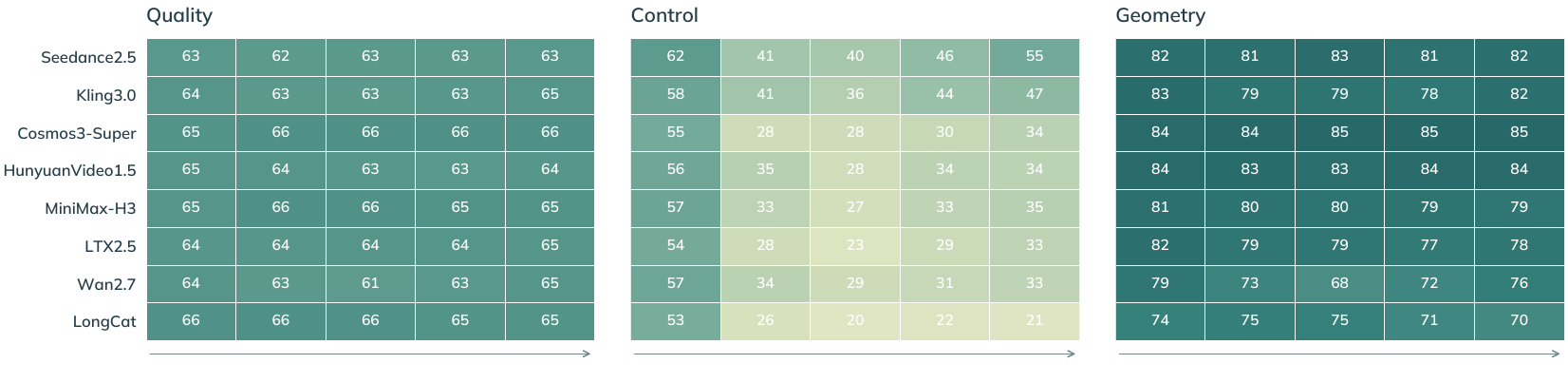}
\par
\includegraphics[width=\linewidth]{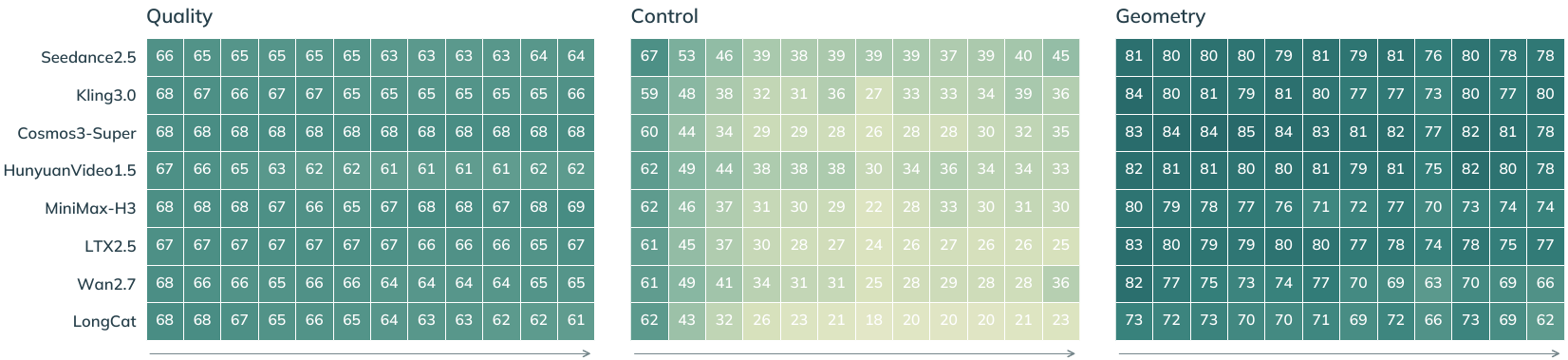}
\caption{\textbf{Temporal performance for TI2V models.} \textbf{Top:} \textit{short videos}. \textbf{Bottom:} \textit{long videos}. Rows follow the model order in Table~\ref{tab:results}. \textit{Time increases from left to right}. Cells show mean window scores for \textbf{quality}, \textbf{control}, and \textbf{local geometry} on a shared 0--100 scale, rounded to integers. Darker colors indicate higher scores within each evaluated time window.}
\label{fig:analysis-time-video-absolute}
\end{figure}

\subsection{Performance Degradation during Long Exploration}
\label{sec:analysis-temporal}
Figures~\ref{fig:analysis-time-world-absolute} and~\ref{fig:analysis-time-video-absolute} show temporal performance in visual quality, control, and local geometry for all models in successive five-second windows. Local geometry includes depth, feature, and regional-scale consistency.

\noindent\textbf{Control declines most over time in leading models from both groups.}
Across SolarWM-H3, Lyra2, EVOKE, Echo-WM, and Gen3C, control drops by an average of $26.98$ points from the first to the final five-second window of minute-long videos. For Seedance2.5, Kling3.0, and Cosmos3-Super, the corresponding decline is $23.25$ points. Control declines more than visual quality or local geometry in every one of these eight models. SolarWM-H3 falls from $91.91$ to $64.50$, and Cosmos3-Super from $60.26$ to $34.94$. Sustaining accurate camera motion remains difficult for both groups.

\noindent\textbf{TI2V visual quality remains stable while local geometry deteriorates.}
During minute-long generation, local geometry declines more than visual quality in seven of the eight TI2V models. The mean declines are $7.13$ points for local geometry and only $2.08$ for visual quality. Wan2.7, for example, loses $16.26$ points in local geometry but only $3.11$ in quality. Cosmos3-Super's quality improves slightly while its geometry falls by $5.34$ points. Visually stable frames therefore do not ensure that geometric agreement is maintained during long exploration.

\noindent\textbf{Long trajectories make newly generated content much harder to preserve.}
Figure~\ref{fig:analysis-templates} shows lower generated-content persistence (Tier~4) on long videos for every trajectory template in both model groups. On forward--backward trajectories, scores fall from $66.21$ to $40.96$ for world models and from $67.74$ to $44.58$ for TI2V models. On peek-and-return, they fall from $64.29$ to $37.32$ and from $69.52$ to $42.83$, respectively. The drop spans roughly $15$--$27$ points across the six templates. Long exploration exposes forgetting even on simple forward--backward paths.

\subsection{Performance Variation across Scene Sources}
\label{sec:analysis-source}
\begin{figure}[t]
\centering
\includegraphics[width=\linewidth]{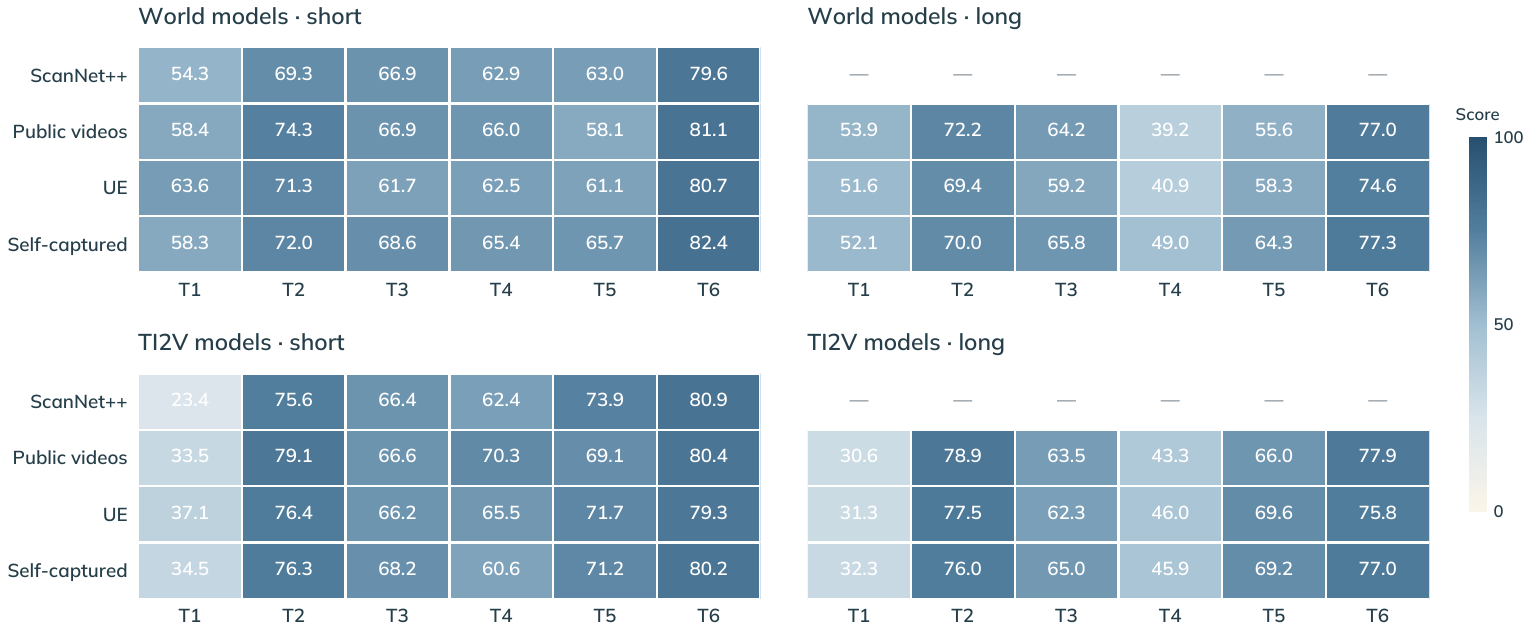}
\caption{\textbf{Performance across data sources.} \textbf{Top:} \textit{world models}. \textbf{Bottom:} \textit{TI2V models}. \textbf{Left:} \textit{short videos}. \textbf{Right:} \textit{long videos}. Cells show \textbf{mean tier scores} with equal model weights. Dashes indicate \textit{unavailable groups}.}
\label{fig:analysis-sources}
\end{figure}

Figure~\ref{fig:analysis-sources} compares ScanNet++ (with sensor depth and reconstructed meshes), a public video group (SpatialVID, Sekai, and MiraData), UE simulation, and self-captured scenes.

\noindent\textbf{Regular UE trajectories are easier to follow in short videos.}
UE scenes have the highest short-video control scores in both groups: $63.6$ for world models and $37.1$ for TI2V models. The trajectories are authored directly from the six templates, then previewed and refined in the simulator. Regular motion may make these paths easier to follow.

\noindent\textbf{The public video group yields the highest visual quality.}
Both model groups achieve their highest visual-quality scores on the public video group in both short and long videos. The scores are $74.3$ and $72.2$ for world models on short and long videos, and $79.1$ and $78.9$ for TI2V models. This advantage may reflect a closer match between these real-video scenes and the visual distributions learned by current video generators.

\subsection{Alignment with Human Judgments}
\label{sec:human-validation}
WorldLens~\citep{liang2026worldlens} complements quantitative metrics with human assessments of realism, physical plausibility, consistency, and safety. Here, we assess whether automatic scores reproduce human preferences in the overall ordering of models across the six evaluation tiers. We select 90 cases spanning six models and distribute the annotation tasks among 20 raters. Raters rank the available videos separately for all six tiers. Ties are allowed. The interface provides the input and query references, requested trajectory, generated videos, and revisit clips appropriate to each judgment.

We convert rankings into relative preference scores, assigning 1 for a win, 0.5 for a tie, and 0 for a loss, and average across raters and cases. Human and automatic scores are compared on the same available cases. Spearman correlations between human and automatic model-level scores range from $0.859$ to $0.921$ across the six tiers, with \textbf{temporal stability} showing the highest agreement. The automatic tiers thus order models in agreement with human judgments.

\let\orbiaConclusionBottomTitleSpace\bottomtitlespace
\def\bottomtitlespace{18\baselineskip}
\section{Conclusion}
\let\bottomtitlespace\orbiaConclusionBottomTitleSpace
\orbia{} provides 720 exploration-and-revisit cases across indoor and outdoor environments and six motion protocols to evaluate input preservation, generated-content persistence, and 3D self-consistency. By distinguishing observed content from newly generated regions, and combining appearance, geometry, and held-out reconstruction tests, the benchmark diagnoses complementary aspects of whether generated videos depict a persistent 3D world.

Our evaluation of 27 text-, action-, and camera-conditioned models shows that strong visual quality and accurate control do not ensure 3D consistency, while local cross-view agreement can coexist with poor scene reconstruction. Preserving the input scene also does not reliably predict whether newly generated content survives a revisit. Minute-long exploration further exposes declining control, content forgetting, and geometric deterioration that visually stable frames can conceal. These findings motivate evaluating world models through both sustained exploration and explicit revisits, with separate tests of content retention and spatial coherence. We hope \orbia{} supports more reliable comparisons and guides the development of models that maintain persistent spatial memory, follow camera commands, and preserve coherent scenes throughout exploration.

\section*{Author Contributions}
\label{sec:contributions}
\begin{itemize}[leftmargin=1.2em, itemsep=2pt, topsep=2pt, parsep=0pt]
\item \textbf{Project Leadership:} \\Dongyue Lu, Shengju Qian
\item \textbf{Benchmark \& Metric Design:} \\Dongyue Lu, Hongshan Chen, Hur Lim, Yanrong Wang, Yonggang Wu, Guian Fang, Shengju Qian
\item \textbf{Data Collection \& Curation:} \\Dongyue Lu, Hongshan Chen, Hur Lim, Yanrong Wang, Yonggang Wu, Guian Fang
\item \textbf{Simulation Data Construction (Unreal Engine):} \\Hongshan Chen, Hur Lim, Yanrong Wang, Yonggang Wu \\\emph{(equal contributions, listed in no particular order)}
\item \textbf{Model Integration \& Experiments:} \\Dongyue Lu, Guian Fang, Yue Liao, Wei Chow, Shengju Qian
\item \textbf{Writing \& Editing:} \\Dongyue Lu, Guian Fang, Yue Liao, Lingdong Kong, Tianxin Huang, Yu Yang
\item \textbf{Discussions \& Insights:} \\Lingdong Kong, Tianxin Huang, Yu Yang, Hongji Yang, Junchao Huang, Ao Liang, Yihao Wang, Zihan Wang, Rong Li, Hanlin Chen
\item \textbf{Advising:} \\Xin Wang, Wei Tsang Ooi, Mike Zheng Shou, Shuicheng Yan
\end{itemize}

\orbiaappendix
\vspace{0.5cm}
\clearpage
\begin{figure}[t]
\centering
\includegraphics[width=\linewidth]{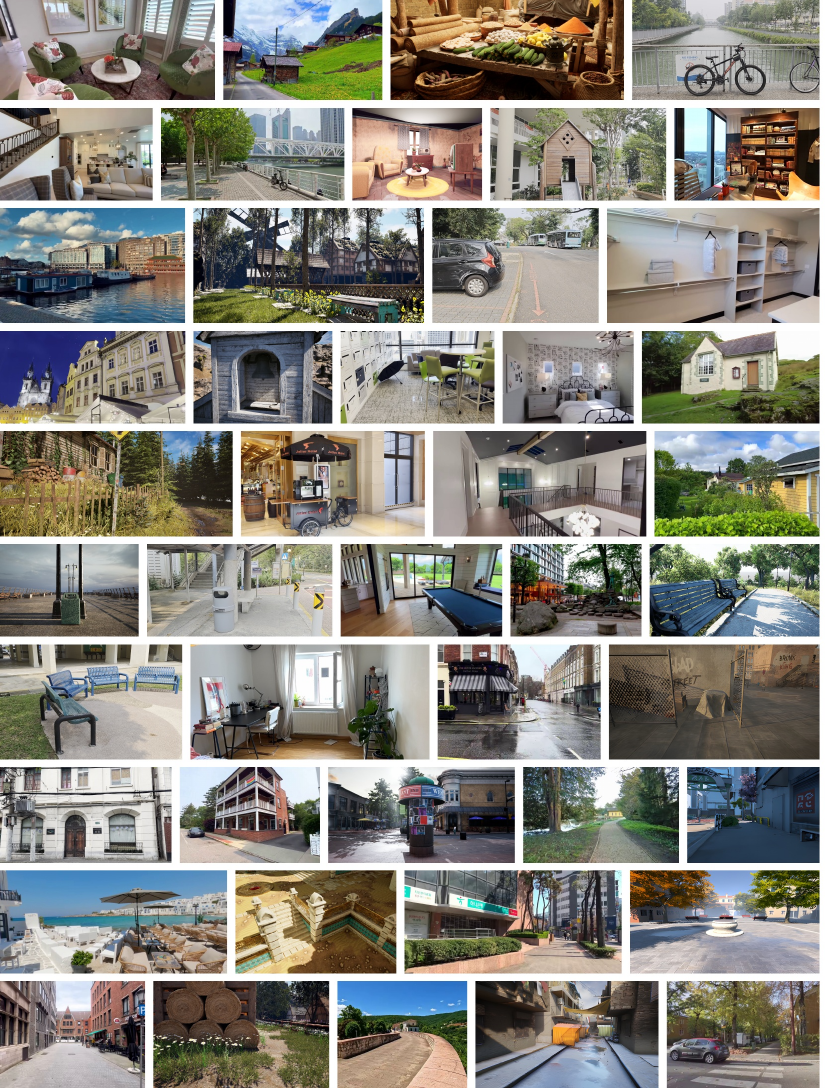}
\caption{\textbf{Dataset gallery.} Selected input images from the \orbia{} benchmark illustrate diverse indoor and outdoor environments.}
\label{fig:gallery}
\label{fig:gallery-collected}
\end{figure}
\clearpage

\begin{figure}[t]
\centering
\tightfigurecaption
\includegraphics[width=\linewidth]{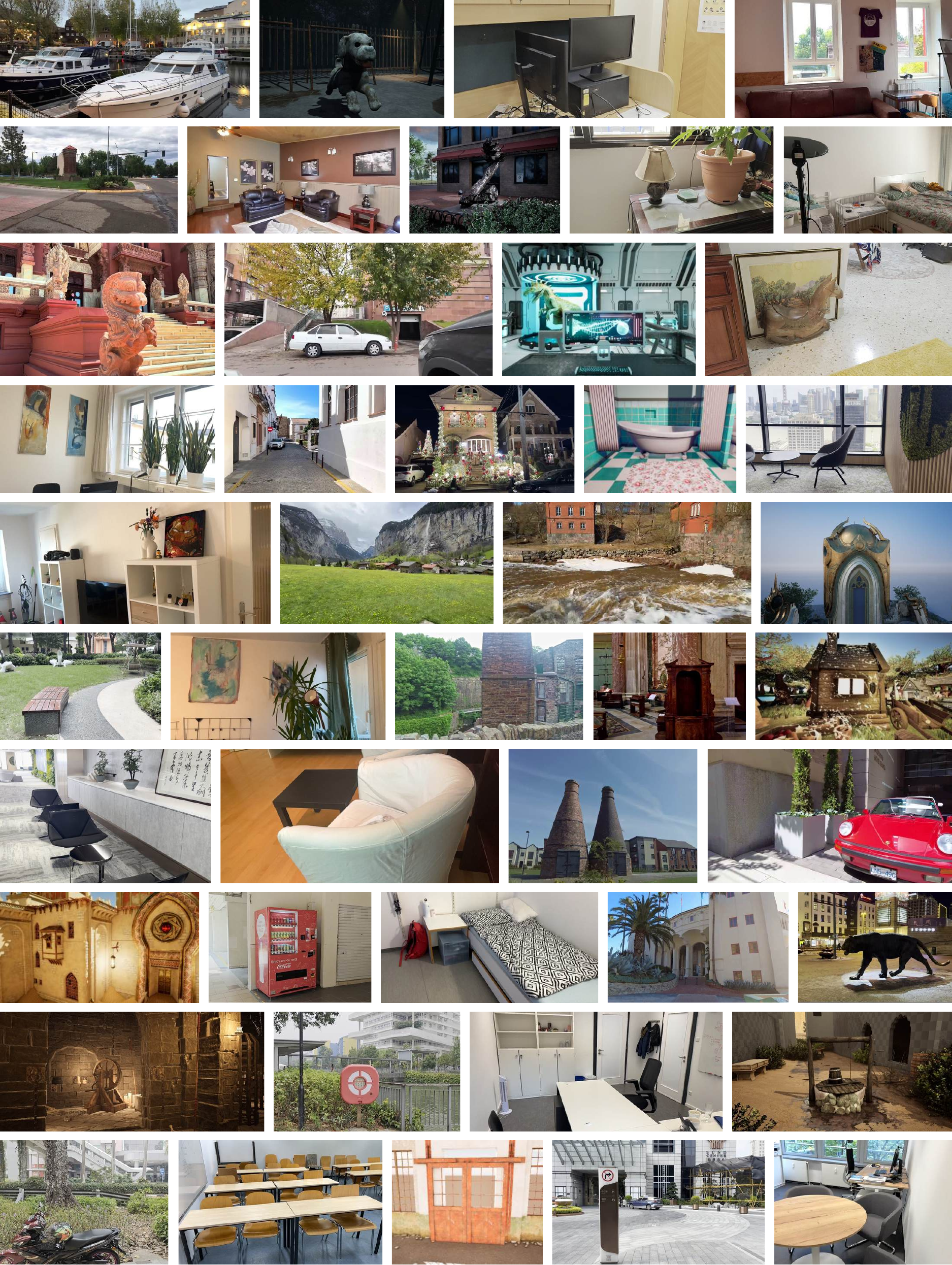}
\caption{\textbf{Dataset gallery.} Selected input images from the \orbia{} benchmark illustrate diverse indoor and outdoor environments.}
\label{fig:gallery-additional}
\end{figure}
\clearpage

\section{Dataset Overview and Annotated Examples}
\label{app:gallery}
This appendix describes the composition and reference annotations of \orbia{}. We report dataset statistics in Section~\ref{app:dataset-statistics}, describe the exploration templates in Section~\ref{app:trajectory-suite}, and present complete annotated examples in Section~\ref{app:annotated-examples}. 

\subsection{Dataset Statistics}
\label{app:dataset-statistics}
\orbia{} contains \textbf{$\mathbf{720}$ evaluation cases} and \textbf{$\mathbf{859}$ anchor evaluation targets}. The dataset combines varied scenes with registered reference views and repeatable camera trajectories, enabling comparisons of how models preserve observed content and maintain newly generated regions during exploration.

\noindent\textbf{Scene coverage.}{\ }
Public datasets cover indoor rooms, streets, and natural environments. Self-captured videos add outdoor landmarks, while UE provides virtual scenes with rendered depth and known camera poses. This combination supports evaluation across real and simulated environments. We scanned $110{,}406$ public videos and scenes, reduced them to $10{,}944$ machine-selected candidates and an $881$-case diversity pool, and retained $240$ public-data cases. Table~\ref{tab:data-sources} lists the raw sample counts and final source composition.

\noindent\textbf{Annotated references.}{\ }
Each case provides an input image and a registered query view, with depth and anchor masks. \textbf{Anchors} occupy a median of $5.2\%$ of the input image and $5.4\%$ of the query image, so changes to these objects can be overlooked in whole-image scores. Their masks and depth enable object-level appearance and geometry comparisons. At the query view, input preservation is evaluated on surfaces visible in the input; newly revealed regions are evaluated at later registered revisits to the corresponding scene surfaces.

\noindent\textbf{Exploration protocols.}{\ }
Each template contributes $80$ short and $40$ long cases. Forward--backward and lateral trajectories primarily change camera position, while yaw and orbit introduce larger orientation changes (Table~\ref{tab:template-rotation}). \textbf{Long extensions} preserve the short case's initial scene and references while adding more revisits and longer intervals between them (Table~\ref{tab:sequence-specifications}), allowing us to assess content preservation as exploration continues.

\subsection{Exploration-and-Revisit Templates}
\label{app:trajectory-suite}
\begin{figure}[t]
\centering
\includegraphics[width=\linewidth]{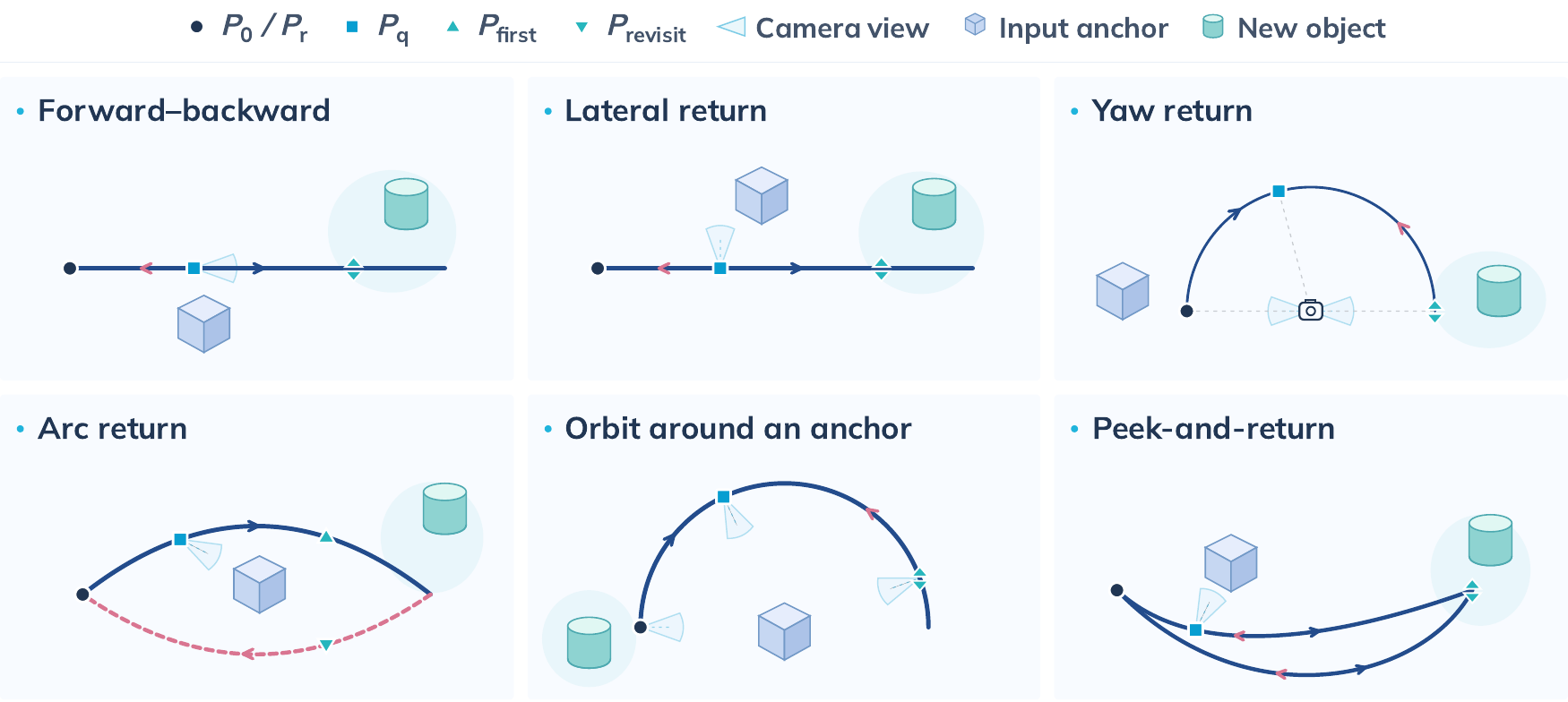}
\caption{\textbf{Exploration-and-revisit templates.} Blue arrows show outward motion and rose arrows show return motion. In yaw return, the camera stays at a fixed center and the symbols along the arc indicate viewing directions.}
\label{fig:routes}
\end{figure}

The six trajectory templates vary how the camera explores a scene and revisits previously observed regions (Figure~\ref{fig:routes}). Each case starts at $P_0$ and includes a registered query pose $P_q$ and a return pose $P_r$. The paired poses $(P_{\mathrm{first}},P_{\mathrm{revisit}})$ provide two observations of content revealed during exploration. Section~\ref{app:annotated-examples} shows the trajectories and reference views for annotated examples from each of the four scene source groups.

\begin{itemize}
    \item \textbf{Forward--backward.} The camera advances into the scene and retraces its path while maintaining its viewing direction. Changes in object size and occlusion test whether scene depth and object placement remain consistent as the camera approaches and retreats.

    \item \textbf{Lateral return.} The camera moves sideways and returns while keeping a similar viewing direction. Nearby and distant surfaces shift by different amounts, testing whether the model preserves their relative depth and recovers surfaces that become temporarily occluded.

    \item \textbf{Yaw return.} The camera rotates to reveal another part of the scene and then turns back, with little translation. This tests whether content remains recognizable after leaving the field of view and appearing again.

    \item \textbf{Arc return.} The camera follows a curved trajectory through the scene before returning. The changing position and viewing direction reveal different sides of objects, testing whether their appearance and geometry remain consistent across changing viewpoints along the curved trajectory.

    \item \textbf{Orbit around an anchor.} The camera moves around an anchor while looking toward it. Observing the same object from multiple directions tests whether its shape and position relative to surrounding objects remain consistent when viewed from different sides and at different points along the trajectory.

    \item \textbf{Peek-and-return.} The camera passes an anchor near the starting view, explores a previously unseen region, and returns. A second excursion revisits that region, providing the pair $(P_{\mathrm{first}},P_{\mathrm{revisit}})$ to test whether newly generated content is retained between visits.
\end{itemize}

\begin{table}[t]
\centering
\begin{tablebody}{3pt}{1.12}
\small
\caption{\textbf{Dataset composition.}}
\label{tab:data}
\label{tab:data-sources}
\begin{tabularx}{\linewidth}{@{}Xlrrrrrl@{}}
\toprule
\rowcolor{orbiaTableHeader}
\textbf{Source} & \textbf{Data Type} & \shortstack{\textbf{Raw Data}} & \multicolumn{2}{c}{\textbf{Cases}} & \multicolumn{2}{c}{\textbf{Anchors}} & \textbf{Geometry}\\
\cmidrule(lr){4-5}\cmidrule(lr){6-7}
& & & Short & Long & Short & Long & \\
\midrule
ScanNet++~\citep{yeshwanth2023scannetpp} & Real & $906$ & $60$ & $0$ & $106$ & $0$ & Measured\\
\rowcolor{orbiaTableStripe}
SpatialVID~\citep{wang2025spatialvid} & Real & $100{,}000$ & $134$ & $74$ & $177$ & $90$ & Estimated\\
Sekai~\citep{li2025sekai} & Real & $8{,}000$ & $25$ & $25$ & $35$ & $35$ & Estimated\\
\rowcolor{orbiaTableStripe}
MiraData~\citep{ju2024miradata} & Real & $1{,}500$ & $21$ & $21$ & $28$ & $28$ & Estimated\\
Our recordings & Self-captured & $85$ & $60$ & $30$ & $60$ & $30$ & Estimated\\
\rowcolor{orbiaTableStripe}
UE simulation & Simulated & $365$ & $180$ & $90$ & $180$ & $90$ & Rendered\\
\midrule
\rowcolor{orbiaTableHighlight}
\textbf{Total} & & $\mathbf{110{,}856}$ & $\mathbf{480}$ & $\mathbf{240}$ & $\mathbf{586}$ & $\mathbf{273}$ & \\
\bottomrule
\end{tabularx}
\end{tablebody}
\par
\protocoltableslayout
\begin{minipage}[t]{0.485\linewidth}

\caption{\textbf{Short and long protocols.} Median interval measures time between first visits and their paired revisits.}
\label{tab:sequence-specifications}
\centering\footnotesize
\tablelayout{3pt}{1.16}
\begin{tabularx}{\linewidth}{@{}Xcc@{}}
\toprule
\rowcolor{orbiaTableHeader}\textbf{Specification} & \textbf{Short} & \textbf{Long}\\
\midrule
RGB resolution & \multicolumn{2}{c}{$1280\times720$}\\
\rowcolor{orbiaTableStripe}Frame rate & \multicolumn{2}{c}{$16$ FPS}\\
Frame count & $357$ & $961$\\
\rowcolor{orbiaTableStripe}Duration & $22.25$ s & $60$ s\\
Revisit pairs & $1$ & $3$\\
\rowcolor{orbiaTableStripe}Median interval & $8.5$ s & $29.5$ s\\
\bottomrule
\end{tabularx}
\end{minipage}\hfill%
\begin{minipage}[t]{0.485\linewidth}

\caption{\textbf{Median input--query motion.} Translation is camera-center distance in the original reference-pose units.}
\label{tab:template-rotation}
\centering\footnotesize
\tablelayout{3pt}{1.16}
\begin{tabularx}{\linewidth}{@{}Xrr@{}}
\toprule
\rowcolor{orbiaTableHeader}\textbf{Template} & \textbf{Rotation} & \textbf{Translation}\\
\midrule
Forward--backward & $1.1^\circ$ & $1.36$\\
\rowcolor{orbiaTableStripe}Lateral return & $5.4^\circ$ & $1.47$\\
Yaw return & $31.3^\circ$ & $0.02$\\
\rowcolor{orbiaTableStripe}Arc return & $11.6^\circ$ & $1.57$\\
Orbit around an anchor & $47.5^\circ$ & $1.26$\\
\rowcolor{orbiaTableStripe}Peek-and-return & $11.2^\circ$ & $0.75$\\
\bottomrule
\end{tabularx}
\end{minipage}

\end{table}

\subsection{Annotated Dataset Examples}
\label{app:annotated-examples}
The following cards show model inputs, requested trajectories, and evaluator-only references. Query RGB, depth, and anchor masks are withheld from generation. We retain one figure for each source group: ScanNet++, public videos, UE, and self-captured videos. Additional trajectory examples appear in Appendix~\ref{app:additional-annotations}. Trajectory panels show the requested camera positions and trajectories in a top-down view of the scene.

\begin{figure}[tp]
\centering
\includegraphics[width=.92\linewidth]{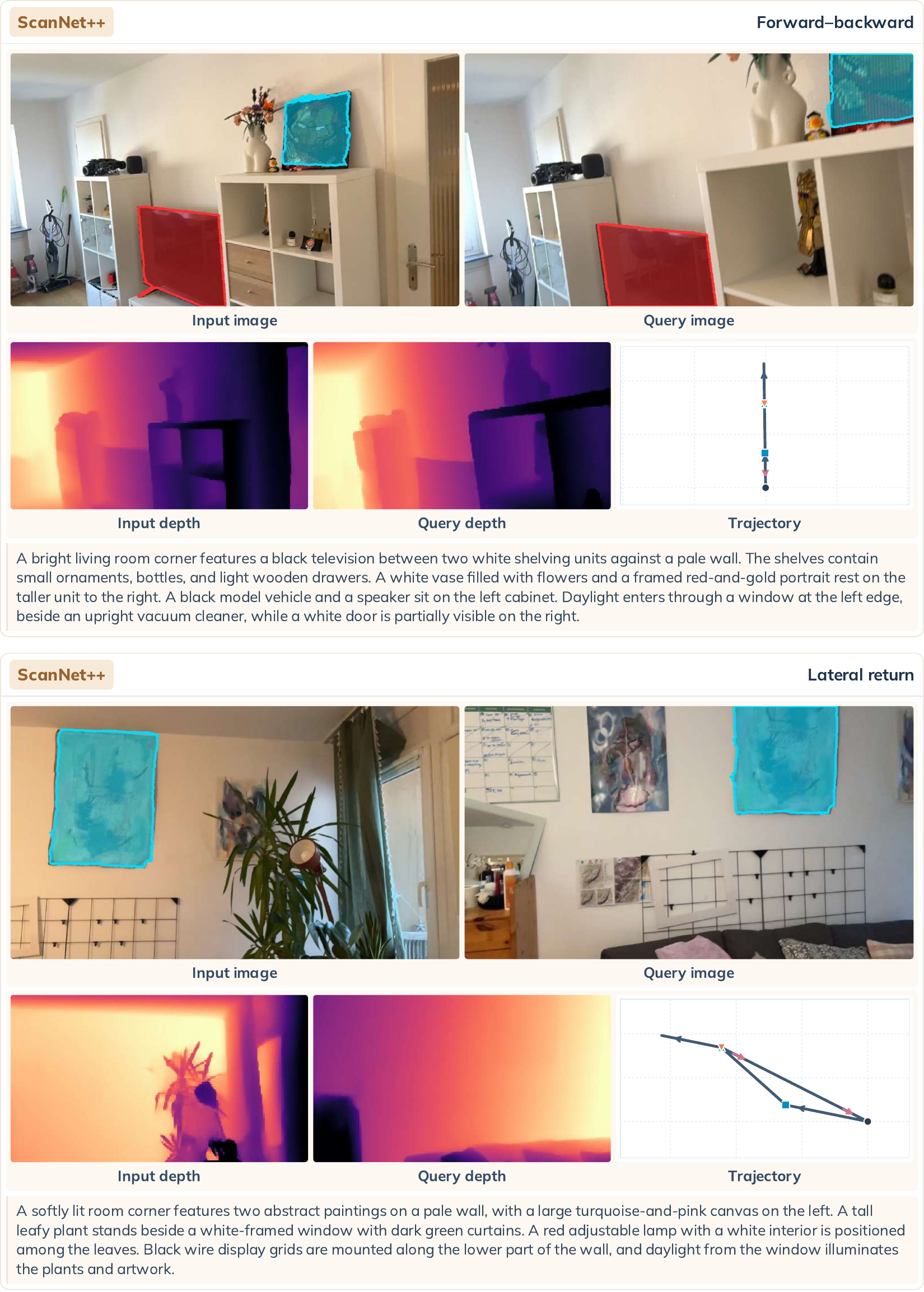}
\caption{Examples of the forward--backward (top) and lateral return (bottom) templates from ScanNet++.}
\label{fig:annotated-01}
\end{figure}

\begin{figure}[tp]
\centering
\includegraphics[width=.92\linewidth]{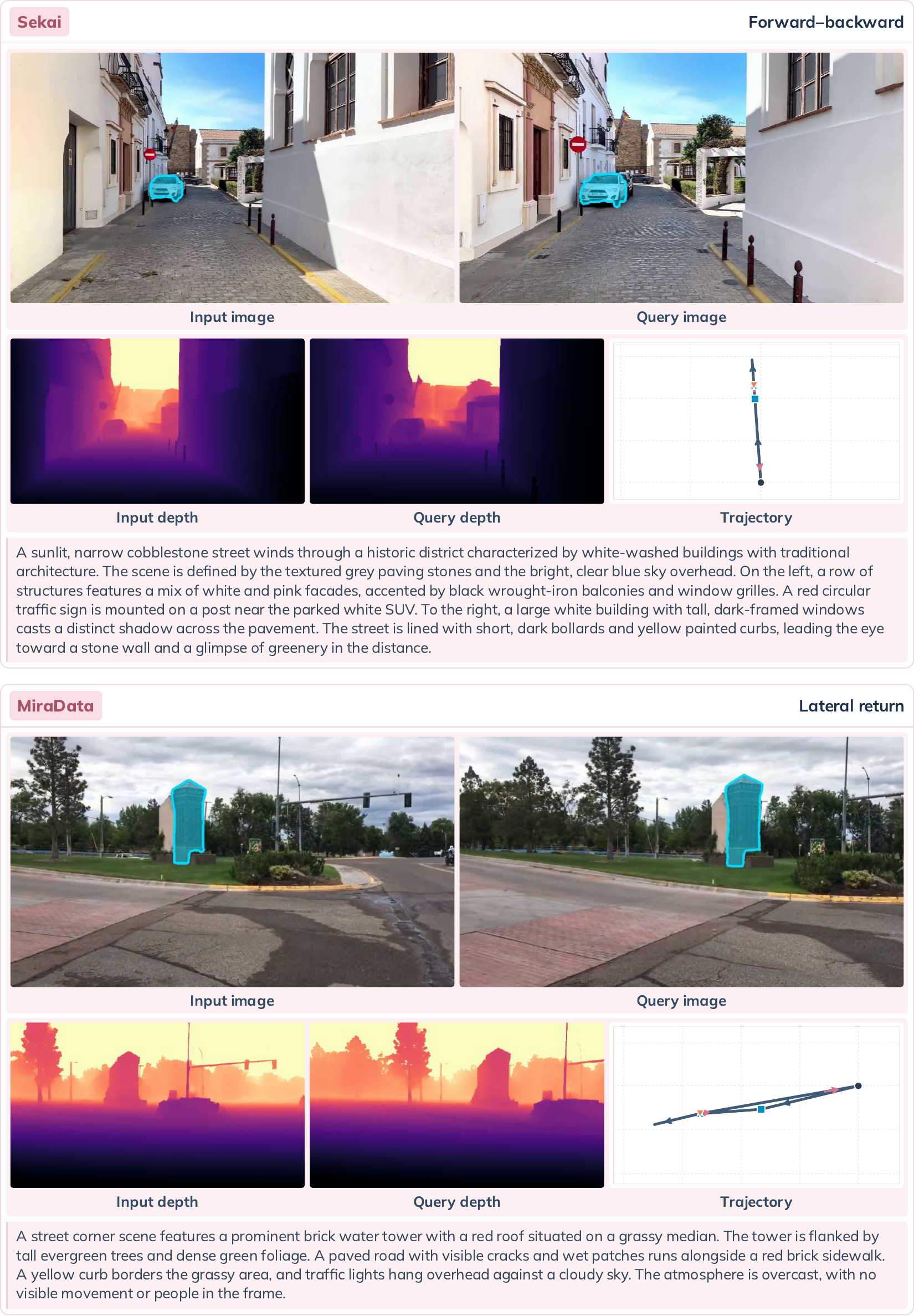}
\caption{Examples of the forward--backward template from Sekai (top) and the lateral return template from MiraData (bottom).}
\label{fig:annotated-04}
\end{figure}

\begin{figure}[tp]
\centering
\includegraphics[width=.92\linewidth]{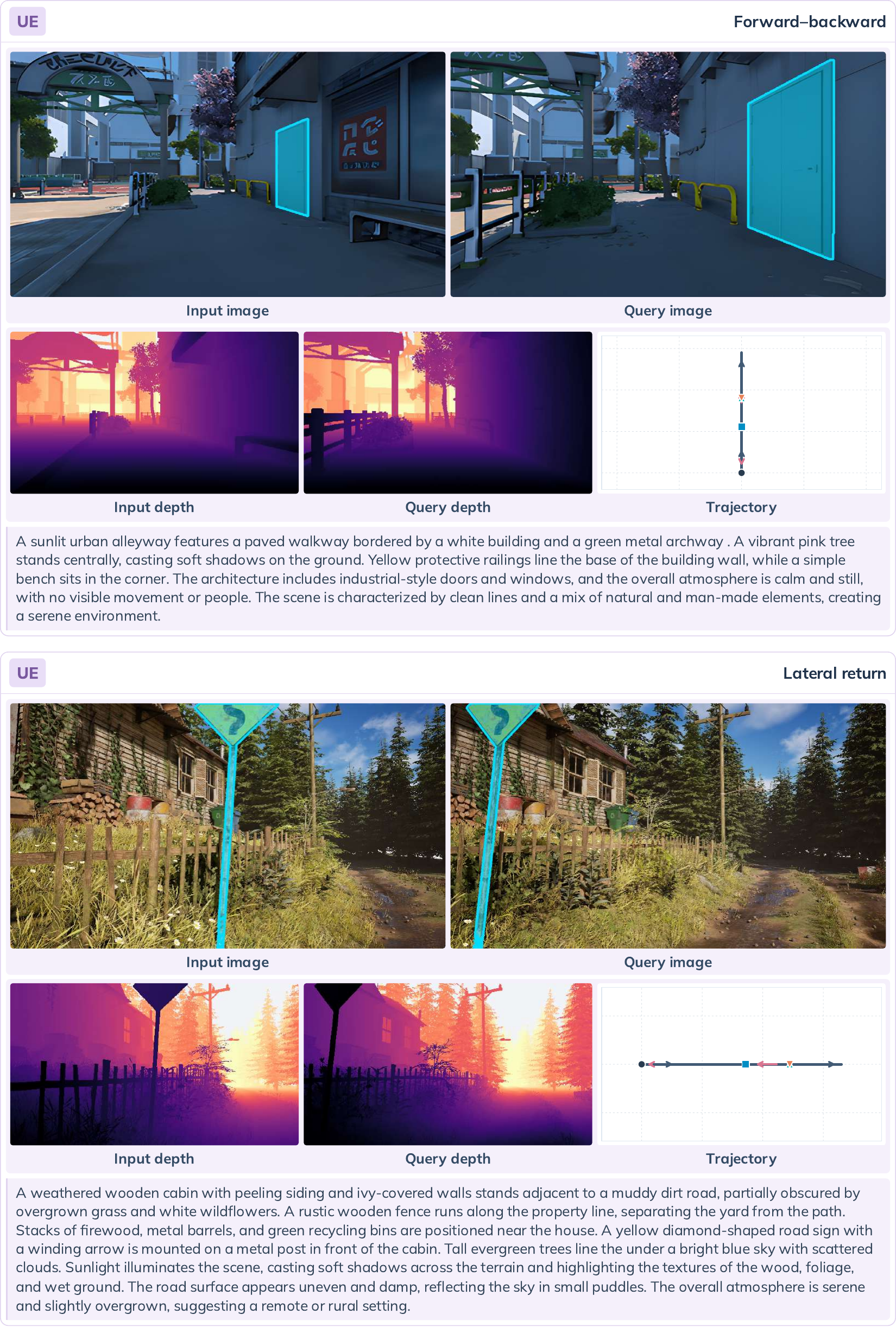}
\caption{Examples of the forward--backward (top) and lateral return (bottom) templates from UE.}
\label{fig:annotated-07}
\end{figure}

\begin{figure}[tp]
\centering
\includegraphics[width=.92\linewidth]{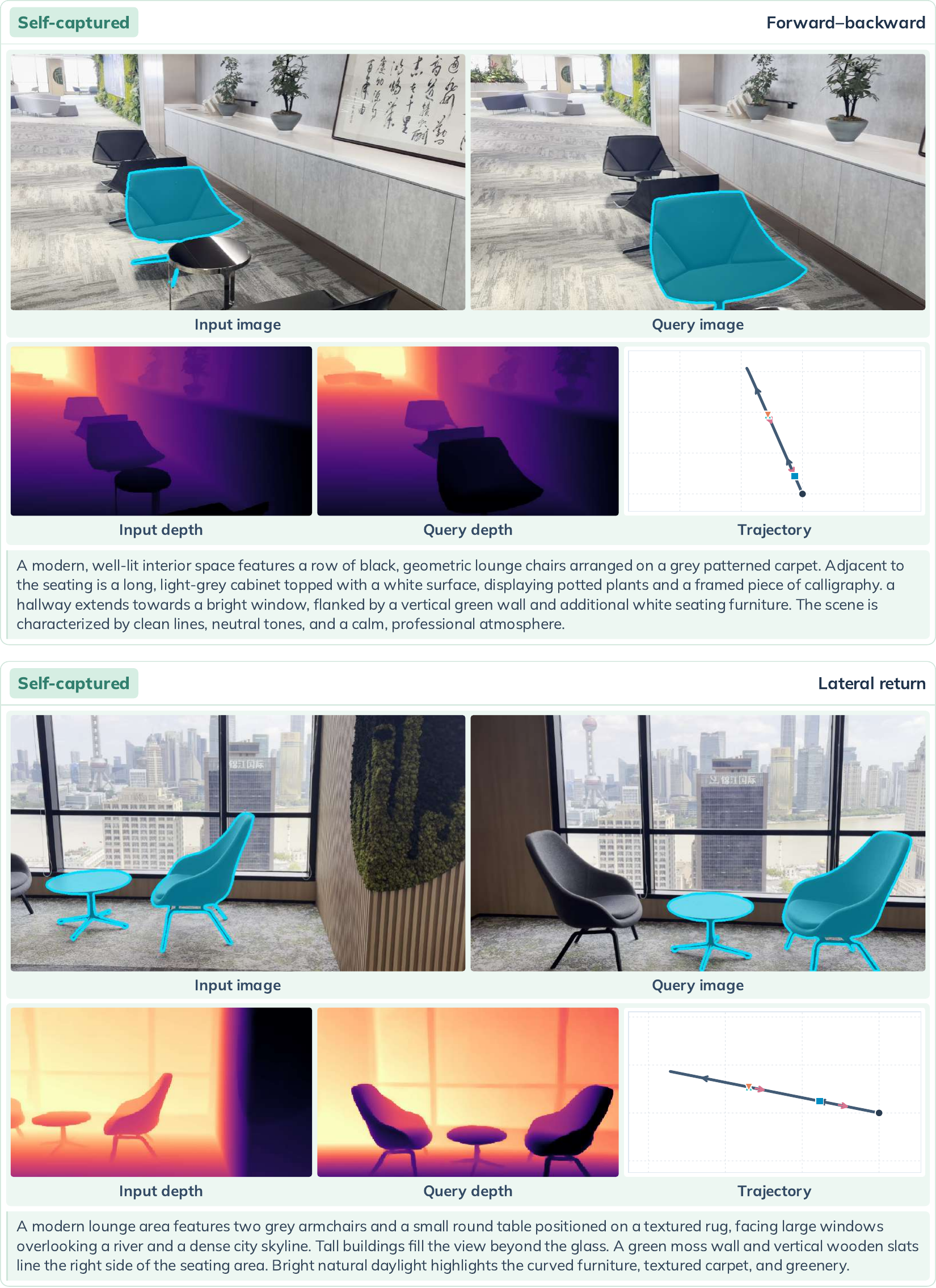}
\caption{Examples of the forward--backward (top) and lateral return (bottom) templates from self-captured videos.}
\label{fig:annotated-10}
\end{figure}

\section{Dataset Construction and Annotation}
\label{app:data-construction}

Constructing an \orbia{} case requires a pair of registered reference views, anchors that can be verified across those views, and a camera trajectory that supports exploration and revisits. We describe how these components are assembled from different sources, extended to one minute, and reviewed before release. Table~\ref{tab:data-sources} reports dataset statistics, and Appendix~\ref{app:annotated-examples} presents annotated examples.

\subsection{From Source Observations to Evaluation Cases}
\label{app:construction-overview}

Following Section~\ref{sec:benchmark-construction}, our pipeline connects \textbf{observed scene references} with \textbf{designed camera motion} in four stages. Two registered RGB-D views provide the input and query references at $P_0$ and $P_q$. A synthesized trajectory passes through these poses and returns to $P_0$ at $P_r$, enabling controlled exploration without recording every intermediate view.

\begin{constructionsteps}
\item \textbf{Select reference views.} Prepare depth and camera annotations, then identify input--query pairs compatible with the translation and rotation of a trajectory template. Candidate pairs should provide a meaningful viewpoint change while keeping a recognizable landmark visible in both views.
\item \textbf{Verify anchors across views.} Obtain an object mask at each endpoint and check that both masks identify the same static object or structure. Project the input anchor into the query view using depth and cameras to identify the surfaces that can be compared across views.
\item \textbf{Construct exploration and revisits.} Fit a smooth trajectory through the reference poses and check its motion against the available scene geometry. Select a first visit and a later revisit to a region outside the input view, recording the intervening exploration and the surfaces visible at both visits.
\item \textbf{Rank, review, and annotate.} Rank candidates by image, anchor, trajectory, and revisit quality. Human reviewers inspect the proposed annotations and trajectories, correct unsuitable choices, and select the final cases. A VLM produces a factual scene caption for model conditioning.
\end{constructionsteps}

\subsection{Construction from Different Data Sources}
\label{app:source-construction}

The sources differ in which components are available initially. Public videos require view selection from recorded camera motion. ScanNet++ additionally provides a semantic mesh for anchor and visibility annotation. For UE and self-capture, we choose landmarks and trajectory templates during acquisition. The procedures below adapt the common pipeline to these differences.

\subsubsection{Public Videos: View Selection and Anchor Verification}
\label{app:video-construction}

SpatialVID~\citep{wang2025spatialvid}, Sekai~\citep{li2025sekai}, and MiraData~\citep{ju2024miradata} broaden the range of real-world environments. Their recorded camera paths usually lack the revisits required by our protocols, so we extract suitable reference views and synthesize a new trajectory through them.

\begin{figure}[t]
\centering
\begin{promptbody}
\begin{tikzpicture}
\node[orbia vlm prompt=blue] {\begin{minipage}{\dimexpr\linewidth-24pt\relax}
\ttfamily\fontsize{9}{10}\selectfont\raggedright
\textbf{Input.} Five ordered source frames spanning $P_0$ to $P_q$.\par\medskip
\textbf{Anchor selection}\par 
Only for a usable scene, propose 3-8 visual anchors. A valid anchor must:\par\noindent- 
refer to one static physical object or one coherent bounded structure;\par\noindent-
refer to the same physical instance at $P_0$ and $P_q$;\par\noindent- 
be clearly visible and independently identifiable at both endpoints;\par\noindent- 
preferably be a substantial foreground or middle-distance target, rather than a small distant detail, while not occupying almost the entire image;\par\noindent- 
look coherent and bounded enough to segment as one target;\par\noindent-
be distinctive enough not to be confused with nearby instances.

\smallskip
Reject a target that is dynamic, ambiguous, unbounded, severely occluded, too small or distant, easily interchangeable with repeated details, or not the same physical instance at both endpoints. Trees and other vegetation are not valid anchors even when one individual tree is prominent. Prefer larger, closer, more complete and more distinctive targets. Do not judge mask connectivity, holes, or tracking; downstream segmentation performs those checks.

\smallskip
\textbf{Centered-anchor soft preference}\par 
After satisfying identity, staticness, two-endpoint visibility, completeness, and distinctiveness, rank otherwise valid anchors by visual usefulness. Prefer foreground or middle-distance targets with moderate-to-large visible area whose visible mask is near the central region of each image. $P_0$ is the model input, so give somewhat greater priority to a substantial, centered and complete $P_0$ view; $P_q$ must still show the same instance clearly and completely, and should also be reasonably centered when possible. Order candidates from strongest to weakest under these preferences.

\smallskip
Centering is only a tie-breaking preference, never a reason to choose a person, moving object, ambiguous region, unbounded surface, severe occlusion, or a different physical instance. Do not reject an otherwise usable pair merely because no valid anchor is perfectly centered.

\smallskip
\textbf{Segmentation prompts}\par 
For each anchor, provide 2-4 synonymous noun phrases for SAM3.1. Each phrase must contain 1-4 words, name the same physical target, and contain no spatial directions, coordinates, boxes, mask IDs, or multiple semantic objects. Prompt variants must remain at the same semantic target and scale: for example, a viewing platform cannot use ground or dirt path as a synonym.

\smallskip
\textbf{Output contract}\par 
Return exactly one JSON object and no other text. It must contain:\par\noindent-
scene\_hazards: an object containing exactly these boolean fields: shot\_boundary, endpoint\_no\_coview, low\_light\_or\_blur, severe\_weather\_or\_visibility\_loss, salient\_person\_or\_crowd, moving\_traffic\_dominant, persistent\_text\_overlay, open\_or\_generic\_scene, no\_distinctive\_shared\_anchor, severe\_anchor\_occlusion;\par\noindent- 
pair\_usable: boolean, true exactly when every scene hazard is false;\par\noindent- 
pair\_reason: one short explanation of the scene-level decision;\par\noindent-
candidates: an empty list for a rejected pair, otherwise 3-8 objects.

\smallskip
Each candidate object must contain only label, anchor\_kind, and sam\_prompts. anchor\_kind must be "instance" or "compact\_structure". sam\_prompts must contain 2-4 synonymous 1-4 word noun phrases for the same target. Do not output endpoint descriptions, boxes, coordinates, mask IDs, markdown, or additional keys.
\end{minipage}};
\end{tikzpicture}
\end{promptbody}
\caption{\textbf{VLM prompt for initial anchor selection.} The preceding task-purpose and initial scene-gate instructions are omitted here. The full prompt also rejects unsuitable scenes before proposing anchors.}
\label{fig:prompt-anchor}
\end{figure}

\coloredstep{sourcevideo}{Prepare geometry and rank view pairs.} We retain SpatialVID's released depth and cameras and recover missing geometry for Sekai and MiraData with ViPE~\citep{huang2025vipe}. SpatialVID's sparse geometry indices are mapped to the corresponding video frames, while the Sekai/MiraData adapter uses per-frame geometry. We filter registered pose pairs by each template's translation and rotation requirements. Ranking favors motion compatible with the template and sufficient viewpoint change, while balancing template coverage and distributing candidates across different scenes and times within each video. Ranked alternatives allow another pair to be examined when the first fails annotation.

\coloredstep{sourcevideo}{Propose and segment landmarks.} A VLM~\citep{qwen2026qwen35} examines five frames spanning the candidate pair, including both endpoints and three approximately evenly spaced intermediate frames. Figure~\ref{fig:prompt-anchor} presents the anchor-selection instructions. It proposes static, bounded landmarks that are visually distinctive and recognizable at both endpoints. SAM3~\citep{carion2025sam3} segments each proposal independently in the two reference images. We check mask area, connectivity, and completeness, remove duplicate proposals, and track shortlisted objects between views to verify that the endpoint masks identify the same physical landmark.

\coloredstep{sourcevideo}{Filter fragmented masks and identity mismatches.} The SpatialVID strict anchor profile accepts endpoint masks covering $0.3\%$--$50\%$ of the image and requires tracking agreement of at least $0.8$ IoU at both endpoints. Mask integrity is assessed through connected components and internal gaps. The largest component must contain at least $85\%$ of the mask, with at most one significant component and an internal-gap ratio no greater than $20\%$. A component is significant when it contains at least the larger of $16$ pixels and $0.2\%$ of the mask area. These checks remove fragmented regions and accidental merges of neighboring objects. Tracking then checks that independently segmented endpoint masks follow the same object through the video.

\coloredstep{sourcevideo}{Verify shared anchor surfaces.} We back-project pixels inside the input mask into 3D and project them into the query view. Visibility and depth agreement determine which parts of the query anchor correspond to surfaces observed in the input. This selects anchors that support a meaningful cross-view comparison even when the query reveals substantial new scene content. The resulting masks and projection overlays are retained for human inspection.
For each projected sample, we check positive camera depth, image bounds, and agreement with the query depth. When several samples project to the same pixel, the nearest surface determines visibility. The retained pixels identify the part of an anchor that the input actually reveals. This distinction matters when a viewpoint change exposes another side of the same object: its identity can remain clear while only some of its surfaces have an input reference.

\coloredstep{sourcevideo}{Synthesize the trajectory and select revisits.} We fit the chosen template through $P_0$ and $P_q$, preserving both reference poses and the final return. Available depth supports checks of camera motion and proximity to observed surfaces. Revisit selection considers temporal separation, camera-pose proximity, and shared surfaces outside the input view. Curved trajectories permit different viewpoints of the same region, while peek trajectories use repeated visits near the outward waypoint. Reviewers inspect the paired views together with their geometric overlap before accepting them.

Algorithm~\ref{alg:shared-surfaces} summarizes the shared-surface checks and event-pair annotation. Tolerances and acceptance criteria follow the geometric checks configured separately for each data source.

\begin{constructionalgorithm}[!t]
\begin{minipage}{\linewidth}
\refstepcounter{constructionalgorithm}\label{alg:shared-surfaces}
\hrule\smallskip
\textbf{Algorithm \theconstructionalgorithm\quad Shared-surface and revisit annotation}
\par\smallskip\hrule\smallskip
\small
\textbf{Input:} Registered depth, intrinsics $K_i$, camera-to-world poses $P_i$, endpoint anchor masks $A_0,A_q$, scene geometry $G$, and a requested camera trajectory through the input, query, and revisit poses.\par
\textbf{Output:} Query anchor support $M_q$, a first-visit/revisit pair, shared unseen surface masks, and geometric diagnostics.
\par\smallskip
\textbf{A. Input--query shared anchor surfaces}
\begin{enumerate}
\algorithmspacing
\item Back-project valid input-anchor pixels into world coordinates. For camera-axis depth, $X=P_0[D_0(u)K_0^{-1}\tilde u;1]$; calibrated source adapters handle other depth representations.
\item Transform $X$ to the query camera, $x_q=P_q^{-1}X$, and project to $v=\pi(K_qx_q)$. Discard samples behind the camera or outside the image. Resolve competing projected samples by nearest depth.
\item Retain samples inside $A_q$ with valid query evidence and depth agreement within the source-specific tolerance. Their query pixels form $M_q$. Record support size and depth residuals for anchor verification.
\end{enumerate}
\textbf{B. Shared surfaces outside the input view}
\begin{enumerate}
\setcounter{enumi}{3}
\algorithmspacing
\item For a designated outward visit $i$, choose a later return-branch pose $j$ with small translation and rotation error, respecting the template's event order and temporal-separation requirements.
\item Raycast $G$ at $P_0,P_i,P_j$. Quantize valid first-hit surface points on a common world lattice to obtain visible surface sets $V_0,V_i,V_j$ for the input, outward-visit, and revisit camera poses, respectively.
\item Remove cells within exclusion radius $r$ of input-visible cells and intersect the remaining sets:
\[
U_i=V_i\setminus\mathcal N_r(V_0),\qquad
U_j=V_j\setminus\mathcal N_r(V_0),\qquad S=U_i\cap U_j.
\]
Here $\mathcal N_r$ uses distances between voxel cells, rather than only their centers.
\item Map $S$ back to first-hit pixels in both event views. Record shared-pixel fractions among valid geometry hits, valid-support counts, and pose errors. Apply the source-specific validity checks and coverage grading before retaining the pair for review.
\end{enumerate}
\textit{Geometry:} $G$ is the scene mesh or a depth-derived proxy. Visibility and unseen-surface labels refer to this available geometry at the input, outward-visit, and revisit camera poses used in the annotation.
\par\smallskip\hrule
\end{minipage}
\end{constructionalgorithm}

\subsubsection{ScanNet++: Mesh-Guided Reference Construction}
\label{app:scannet-construction}

ScanNet++~\citep{yeshwanth2023scannetpp} provides indoor RGB-D observations, calibrated cameras, and semantic instance meshes. These annotations allow anchors and cross-view visibility to be derived directly from the reconstructed scene. The visibility operations in Algorithm~\ref{alg:shared-surfaces} use this mesh, with sensor-depth validity checks for registered reference pixels.

\coloredstep{sourcescannet}{Select registered RGB-D endpoints.} We use the iPhone RGB-D stream and calibrated poses to find view pairs compatible with each template. Selection checks image sharpness, valid depth, and camera displacement. The input view must contain a clearly identifiable static instance that remains visible from the query pose.
RGB frames are center-cropped and resized together with their camera calibration. We evaluate candidate pairs using the selected instance's visible surface area, allowing the rest of the query image to reveal new parts of the room. When several anchors are present, each retains its own masks and geometric checks so that a well-observed object can be selected independently of less suitable instances.

\coloredstep{sourcescannet}{Render instance masks and visible surfaces.} We render anchor masks from the semantic mesh, assigning each pixel to the first visible surface along its camera ray. Anchors must have a recognizable silhouette, reliable depth, and enough common surface area for cross-view comparison. We exclude unbounded walls, floors, and ceilings, as well as transparent objects and masks that ambiguously combine several instances. Projecting the input-visible anchor surface into $P_q$ identifies the part of the query mask used for input-preservation evaluation.

\coloredstep{sourcescannet}{Plan and check the camera trajectory.} Translation and rotation are interpolated through the registered endpoints. The mesh supports checks of camera clearance, near-plane intersections, floor and ceiling distance, and abrupt motion. Validation examines the designated evaluation poses, $32$ additional uniformly spaced poses, and the segments between sampled positions. Failed trajectories are replanned with different intermediate waypoints. Mesh visibility also identifies regions outside the initial view, from which we select temporally separated visits sharing visible surfaces.

\coloredstep{sourcescannet}{Associate revisits with scene surfaces.} We record the mesh surfaces visible from $P_0$ and compare this set with the surfaces visible at candidate revisit poses. Surfaces shared by two visits but absent from the input provide locations for generated-content evaluation. Ray visibility handles occlusion as the camera moves through the room. Pixels with invalid depth or unreliable mesh boundaries are excluded from the comparison. This produces reference masks tied to particular scene surfaces and lets reviewers inspect where the camera leaves the input-visible region and later returns.

\subsubsection{UE Simulation: Controlled Capture and Rendered References}
\label{app:ue-construction}

UE provides synchronized images and geometric annotations along prescribed trajectories. We collect $48$ virtual environments and render $365$ candidate videos, retaining $180$ cases after curation. Acquisition starts with a static scene actor and a trajectory template selected for the scene layout and its landmark.

\begin{figure}[t]
\centering
\includegraphics[width=\linewidth]{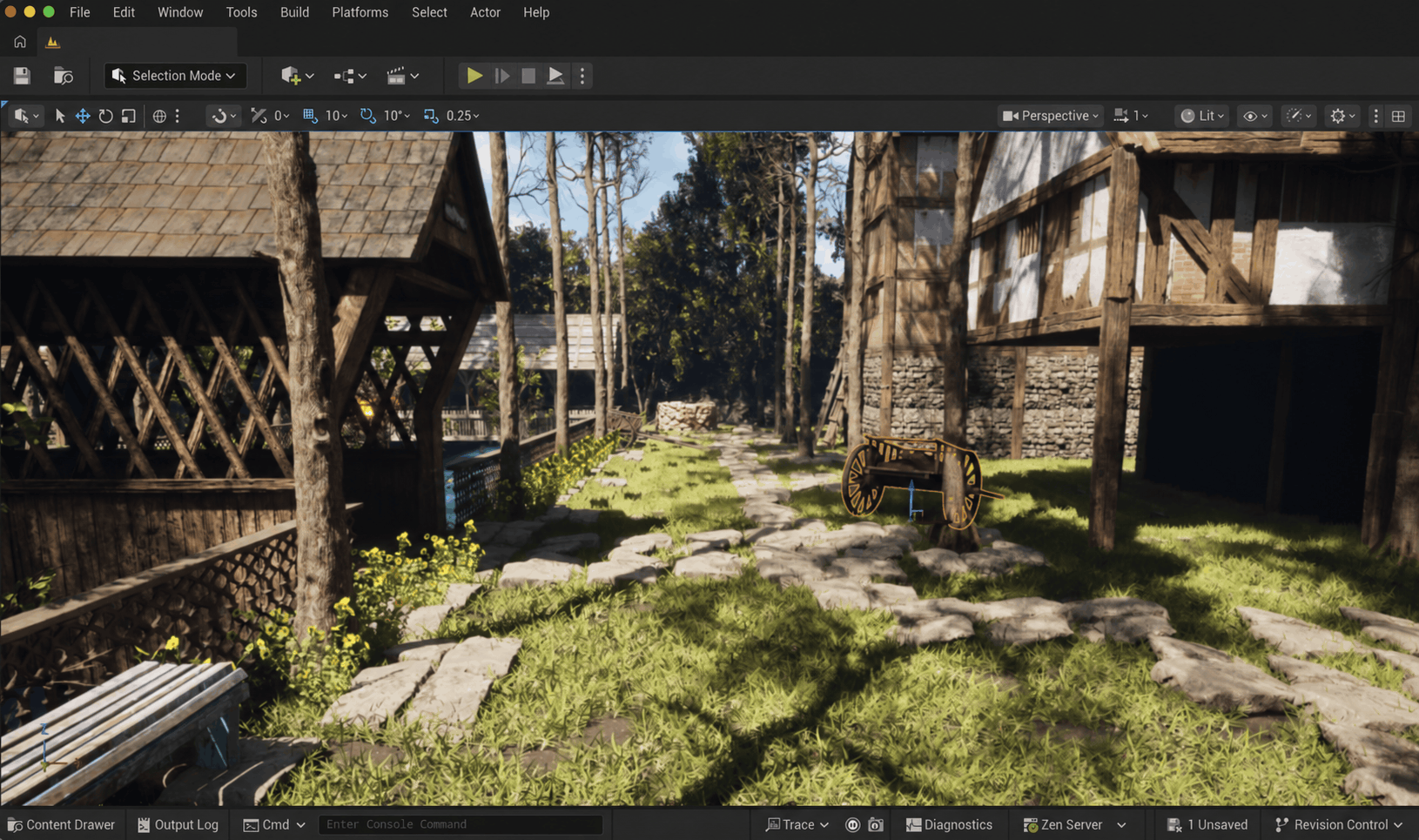}
\caption{\textbf{Anchor selection in the UE editor.} A cart is selected as a bounded static landmark, with its scene Actor highlighted in yellow. Its location and bounds guide the initial camera framing and trajectory design.}
\label{fig:ue-anchor-editor}
\end{figure}

\noindent\textbf{Anchor selection and controlled rendering.}{\ }
We design each capture around a selected landmark and an exploration template. Scene selection, camera placement, and motion are reviewed together so that the resulting video provides both a recognizable anchor and meaningful changes in the surrounding view.

\begin{figure}[tp]
\centering
\includegraphics[width=.9\linewidth]{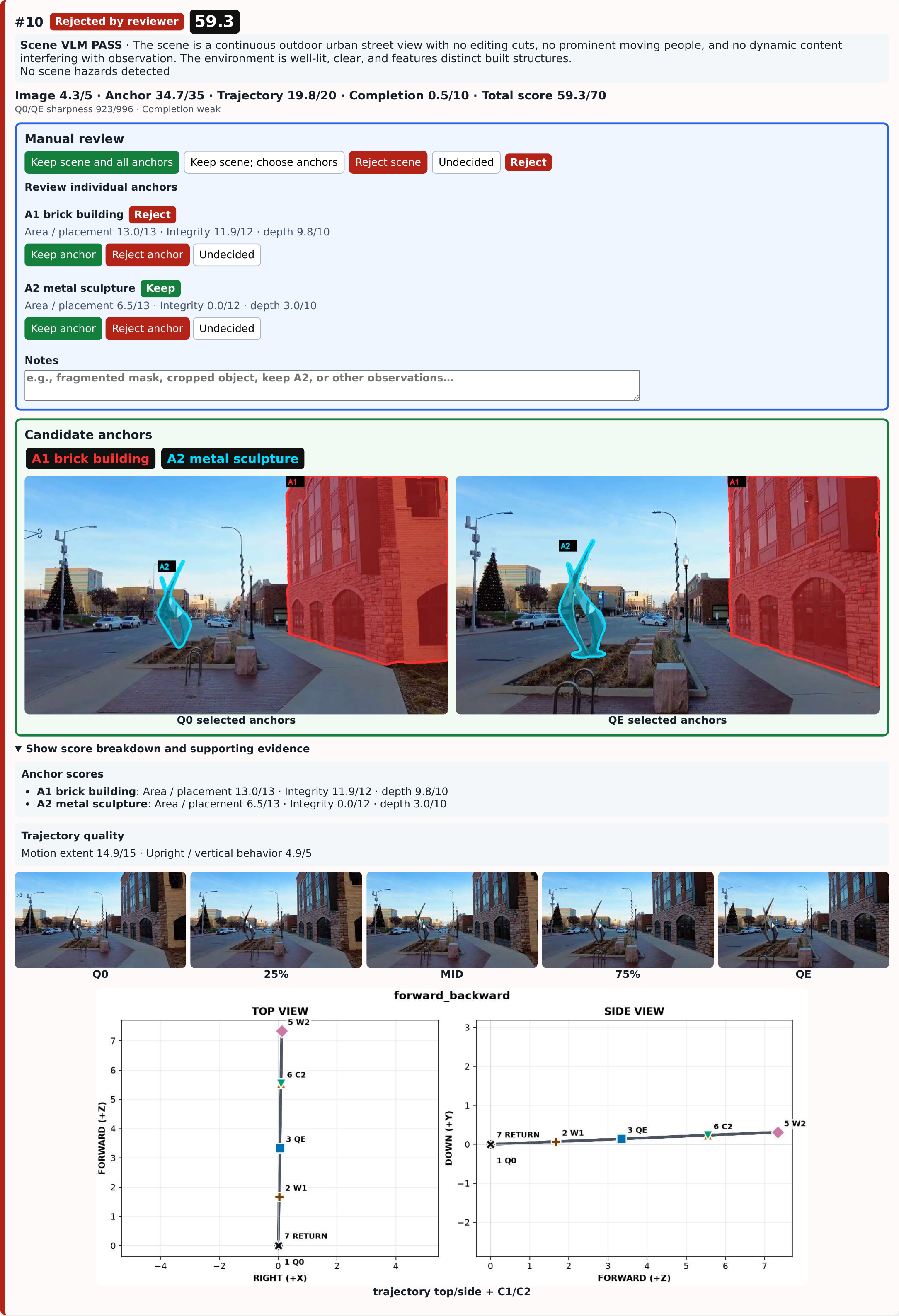}
\caption{\textbf{Manual review of a MiraData forward--backward case.} The interface combines scene review, annotation scores, per-anchor decisions, reference overlays, sampled frames, and camera trajectories.}
\label{fig:qa-miradata}
\end{figure}

\coloredstep{sourceue}{Choose the scene and anchor.} Annotators inspect a virtual environment and select a static object or structure with clear boundaries and a distinctive appearance. Figure~\ref{fig:ue-anchor-editor} illustrates this selection in the UE editor. We favor bounded landmarks that remain recognizable under viewpoint changes, while avoiding thin repetitive geometry, foliage, transparent surfaces, and large unbounded regions. The anchor is associated with its scene Actor, or a group of Actors forming one semantic object, so that its identity remains fixed throughout rendering. Its location and bounds guide camera placement. The initial view should show the anchor clearly together with enough surrounding scene content to support exploration around the selected landmark and into the surrounding scene.

\coloredstep{sourceue}{Assign a trajectory and set its scale.} We choose a template suited to the scene layout while balancing coverage across the six motion types. The anchor's size, camera field of view, and an authored starting view determine the initial framing and a suitable travel distance or rotation range. Forward and lateral paths require room to translate, yaw requires a useful angular sweep, and arc and orbit paths require clearance around the landmark. Peek paths place the anchor near the initial view and extend into a region revealed during exploration.

\coloredstep{sourceue}{Preview and refine the motion.} Candidate paths are checked in UE for camera collisions and inspected through sparse renders of the input, intermediate exploration, query region, and return. Reviewers assess anchor scale, foreground occlusion, background content, and whether the motion visibly follows the intended template. Collision checks alone are insufficient because some visible geometry has no collision enabled. We therefore inspect the rendered views and adjust the starting position, travel direction, radius, or viewing direction when a trajectory is obstructed or poorly framed.

\coloredstep{sourceue}{Render the accepted sequence.} The camera path is placed in a Level Sequence and rendered with Movie Render Queue. RGB, depth, and object IDs share the same camera sequence, and evaluated camera poses and intrinsics are recorded for geometric annotation. Object IDs provide the selected anchor's visible mask at each frame; depth is converted to metric camera-axis $Z$. We allow assets and lighting to settle before capture, control exposure, and disable motion blur and depth of field to preserve clear reference images. The rendered video is then reviewed for smooth motion, loading artifacts, and anchor visibility before reference selection.

\noindent\textbf{Post-processing captured videos.}{\ }
We process the captured RGB videos together with their rendered depth, object masks, and camera poses to construct benchmark cases. This stage selects the input and query references, identifies visits to newly revealed regions, and maps the accepted observations to the benchmark timeline.

\coloredstep{sourceue}{Check reference-view quality.} For the query, candidates must contain a visible anchor covering at least $0.5\%$ of the image and valid anchor depth. Projected input surfaces must explain at least $40\%$ of the query anchor. Reviewers choose among valid candidates using the RGB images, masks, and projected-surface overlays.
The default query search also requires the anchor's image-area fraction to retain at least half its value in the input. This avoids choosing a distant or heavily occluded observation in which the landmark is too small to evaluate. 

\coloredstep{sourceue}{Select visits to newly revealed regions.} Rendered depth and cameras identify surfaces absent from the input and visible at both candidate visits. The general selector requires more than $8$ seconds between visits, a camera-center distance within $1$ meter, and an orientation difference within $15^\circ$. Arc trajectories allow a larger positional difference when both views observe the same region. Peek trajectories use repeated visits near the outward waypoint. Candidates are ranked by newly observed surface content, shared visible surfaces, and temporal separation.

\coloredstep{sourceue}{Align the selected observations.} We map RGB, depth, masks, and camera poses to the $357$-frame, $16$-FPS short protocol using the same source-frame indices. The mapping preserves the input, return, and selected evaluation events.

\subsubsection{Self-Capture: Landmark-Guided Acquisition}
\label{app:selfcapture-construction}

\begin{figure}[t]
\centering
\begin{promptbody}
\begin{tikzpicture}
\node[orbia vlm prompt=red] {\begin{minipage}{\dimexpr\linewidth-24pt\relax}
\ttfamily\fontsize{9}{10}\selectfont\raggedright
\textbf{Input.} Five ordered raw source frames, without annotation overlays or quality scores.\par\medskip
Review the five ordered video frames: $P_0$, 25\%, middle, 75\%, and $P_q$.

\smallskip
First classify the scene as indoor or outdoor. Then decide whether the scene should be accepted or rejected.

\smallskip
Reject the scene if any of the following is true:

\smallskip
\noindent- The frames cross a shot or editing boundary, or do not show one continuous scene.\par\noindent-
One or more people are central, prominent, being followed, presenting to the camera, or visibly moving through the main part of the scene across multiple frames. A single such person is enough to reject the scene, even when the surrounding room or building occupies more pixels.\par\noindent-
Moving people, vehicles, or other dynamic content strongly interfere with observation of the static scene.\par\noindent- 
The image is materially dark, blurred, foggy, obscured, or difficult to inspect.\par\noindent- 
Heavy rain, snow, lens contamination, glare, or other weather significantly reduces visibility.\par\noindent- 
Any subtitle, channel name, logo, or watermark is overlaid on the video. Carefully inspect all four corners in every frame. Fixed screen-position text is an overlay even when it is small, faint, or semi-transparent. Physical text on signs, products, or buildings is not an overlay.\par\noindent-
The scene is mainly open or natural scenery such as road, ground, snow, water, mountain, forest, or generic vegetation, without clear distinctive built objects.

\smallskip
Do not reject only because:\par\noindent- 
one or two people remain small and peripheral in every frame;\par\noindent-
a vehicle is clearly parked;\par\noindent- 
physical text appears on an object or building;\par\noindent- 
vegetation appears around an otherwise distinctive built scene.

\smallskip
If any rejection condition is present, set scene\_status to "reject". Otherwise set it to "pass".

\medskip
Return exactly one JSON object and no markdown:\par
{\ttfamily\footnotesize
\{\par
"environment": "indoor",\par
"environment\_confidence": "high",\par
"scene\_hazards": \{\par
"shot\_boundary\_or\_no\_coview": false,\par
"central\_or\_prominent\_people": false,\par
"dynamic\_content\_dominant": false,\par
"low\_light\_or\_visibility\_loss": false,\par
"severe\_weather\_or\_lens\_contamination": false,\par
"subtitle\_or\_watermark": false,\par
"open\_or\_nature\_dominated": false\par
\},\par
"scene\_status": "pass",\par
"reason": "brief evidence-based reason",\par
\}\par
}

\end{minipage}};
\end{tikzpicture}
\end{promptbody}
\caption{\textbf{VLM prompt for dedicated scene filtering.} The illustrative description value in the JSON schema is shortened for display. The decision instructions are reproduced in full.}
\label{fig:prompt-scene}
\end{figure}

We record $85$ outdoor videos around selected landmarks to broaden scene coverage beyond the public collections. The intended anchor and template are specified before recording, and $60$ cases are retained after review.

\coloredstep{sourceself}{Record the scene and recover geometry.} Acquisition follows the assigned exploration template around a distinctive static landmark. We estimate depth and cameras with DA3~\citep{lin2025da3}, preserving the correspondence between source timestamps, RGB frames, and geometric annotations. The input is selected near the beginning of the recording, with adjustments for blur or exposure.

\coloredstep{sourceself}{Annotate the designated landmark.} Query candidates are selected according to the known template. SAM3 segments the anchor independently at the input and query views, and reviewers verify identity and choose the reference pair. Checks consider mask area, completeness, valid depth, and the query surfaces recovered by projecting the input anchor. The nominal projected-surface threshold is $40\%$ of the query mask.
Here, the acquisition record already supplies the intended landmark. Review therefore concentrates on choosing sharp reference frames and confirming that segmentation follows that landmark in both views. The reviewer checks both endpoint images against the source recording to confirm that the masks follow the same physical landmark in both reference views.

\coloredstep{sourceself}{Regularize motion and select revisits.} We synthesize a smooth trajectory through the selected reference poses, preserving the final return while removing irregularities in handheld motion. Estimated scene geometry supports selection of visits to newly revealed regions. Reviewers inspect the trajectory and paired views, retaining the selected anchors, reference frames, and geometric diagnostics. 

\subsection{Extending Exploration to One Minute}
\label{app:long-construction}

Long cases retain the short case's input image, anchors, and registered reference views, providing a paired setting for examining degradation during extended exploration. Each extension contains $961$ camera poses over $60$ seconds at $16$ FPS and three pairs of generated-region revisits.

\begin{figure}[t]
\centering
\begin{promptbody}
\begin{tikzpicture}
\node[orbia vlm prompt=green] {\begin{minipage}{\dimexpr\linewidth-24pt\relax}
\ttfamily\fontsize{9}{10}\selectfont\raggedright
\textbf{Input.} Five ordered source frames: $P_0$, 25\%, middle, 75\%, and $P_q$.\par\medskip
Write one dense English scene caption from the five ordered frames.

\smallskip
Describe only persistent visible architecture, terrain, vegetation, water, furniture, materials, lighting, weather, atmosphere, and stable spatial relations supported across the frames. Do not mention people, animals, human actions, moving vehicles, transient subjects, or any other temporary event. A stationary parked vehicle may be mentioned only when it is clearly persistent. Never mention the camera, viewpoint, framing, video, trajectory, temporal change, future frames, generation, rendering, games, animation, media, or these instructions. Do not use image-plane directions such as left, right, centre, foreground, or background. Do not speculate beyond visible evidence. Begin directly with scene language such as "A ... scene where ..."; do not write "This image/video...". Use 80-150 English words.

\smallskip
Return exactly one JSON object and no markdown: \{"dense\_caption":"..."\}
\end{minipage}};
\end{tikzpicture}
\end{promptbody}
\caption{\textbf{VLM prompt for final scene captions.} This separate captioning pass describes the persistent architecture, objects, and surroundings used to condition model generation on the initial scene.}
\label{fig:prompt-caption}
\end{figure}

\begin{constructionsteps}
\item \textbf{Preserve the registered references.} The standard extension copies the short trajectory from $P_0$ through $P_q$, including its pose sequence. This keeps the initial exploration and cross-view checkpoint fixed. Some peek cases place $P_q$ at the intermediate retraction between two excursions. Their extension lengthens the first excursion and adjusts the query time while preserving the registered $P_0$ and $P_q$ poses.
\item \textbf{Extend motion within the same template.} We increase travel distance or angular coverage while preserving the template's characteristic motion. Timing follows the short trajectory's median nonzero translation or angular speed. Forward and lateral paths explore farther before returning; yaw expands its sweep to $180^\circ$--$300^\circ$; and arc paths retain their curved shape. Orbit preserves its anchor-centered motion, while peek preserves the outward movement into an unseen region and the subsequent return to the starting view.
\item \textbf{Register three separated revisits.} We select three locations along the extended trajectory and revisit their camera poses after intervening exploration, with at least $5$ seconds between paired observations and a final return to $P_0$. Checks verify pose agreement within each pair, and geometric overlays show the shared surfaces and their visibility from the input and query views.
\item \textbf{Validate and revise the extension.} We check reference-pose preservation, the final return, revisit alignment, and motion continuity. Endpoint depth supports surface-clearance checks along the trajectory, using a radius of $1.5\%$ of the median endpoint depth and a sampling step no greater than half this radius. Failed trajectories are revised by adjusting their direction, heading, or template parameters. Reviewers inspect the resulting path and revisit regions before accepting the extended case into the final evaluation set.
\end{constructionsteps}

\subsection{Quality Ranking and Human Review}
\label{app:human-qa}

Automatic scoring prioritizes candidates for inspection. Human review then resolves choices that require scene understanding, such as whether a mask isolates a meaningful landmark or two visits show the intended region.

\begin{table}[t]
\caption{\textbf{Annotation-quality ranking for public-video candidates.} The SpatialVID scoring scheme assigns up to $70$ points.}
\label{tab:annotation-quality-ranking}
\centering\small
\tablelayout{4pt}{1.18}
\begin{tabularx}{\linewidth}{@{}p{0.14\linewidth}p{0.20\linewidth}cX@{}}
\toprule
\rowcolor{orbiaTableHeader}\textbf{Category} & \textbf{Component} & \textbf{Points} & \textbf{Factors considered}\\
\midrule
Image & Endpoint sharpness & $5$ & Sharpness of input and query RGB.\\
\rowcolor{orbiaTableStripe}Anchor & Area and placement & $13$ & Visible mask area at both endpoints, balance between their areas, and image-space placement of the two endpoint masks.\\
& Integrity and shape & $12$ & Largest connected component, fragmentation, internal gaps, substantial border contact, aspect ratio, and solidity.\\
\rowcolor{orbiaTableStripe}& Depth consistency & $10$ & Fraction of projected samples with consistent depth, and fraction falling inside the opposite-view mask.\\
Trajectory & Motion extent & $15$ & Maximum rotation from $P_0$ for yaw, or maximum horizontal excursion for the other translation-based trajectory templates.\\
\rowcolor{orbiaTableStripe}& Upright and vertical behavior & $5$ & Camera tilt, vertical-to-horizontal excursion ratio, and the vertical slope of the camera path between consecutive poses.\\
Revisit & Shared unseen surfaces & $10$ & Shared surface fraction outside the input view at both visits, support validity, and pose mismatch.\\
\midrule
\textbf{Total} & & $\mathbf{70}$ & \\
\bottomrule
\end{tabularx}
\end{table}

\begin{constructionsteps}
\item \textbf{Rank annotation quality.} The large candidate pool makes exhaustive manual inspection costly: each case requires checking reference images, anchor masks, depth, and camera motion. We therefore use automatic quality scores to prioritize promising candidates, followed by human inspection and final selection. The score considers four aspects: \textsuperscript{1}\emph{image sharpness}; \textsuperscript{2}\emph{anchor quality}, including area, placement, mask integrity, and depth consistency; \textsuperscript{3}\emph{trajectory quality}, including motion extent and uprightness or vertical stability; and \textsuperscript{4}\emph{revisit quality}, including shared unseen surfaces and pose agreement. These criteria favor clear reference views, well-defined landmarks, and informative exploration with verifiable revisits. Table~\ref{tab:annotation-quality-ranking} gives the component weights in the $70$-point scheme. For cases with multiple reliable anchors, the highest-scoring anchor determines the anchor term. Scores guide review order, while reviewers make the final acceptance and anchor-selection decisions.
\item \textbf{Review scene suitability.} We apply a dedicated VLM review after anchor, segmentation, and geometric checks have reduced the candidate pool. Early anchor search processes thousands of candidate frames, so postponing detailed scene review avoids spending additional inference on candidates that fail those checks. The anchor-selection prompt already includes an initial scene gate; this later pass rechecks scene suitability independently of anchor proposals and numerical quality scores (Figure~\ref{fig:prompt-scene}). Separating this pass allows detailed review to concentrate on candidates that survive annotation and geometry checks.
The scene reviewer receives five ordered raw frames and checks for shot changes, prominent moving subjects, poor visibility, severe weather or lens contamination, and intrusive subtitles or watermarks. It also records indoor or outdoor scene type. Anchor overlays, trajectory plots, and numerical scores are excluded from this review to keep the judgment focused on scene content. The pass/reject decision is recorded separately from the numerical ranking. Passing cases are ordered by their automatic scores within the template and indoor/outdoor groups for manual inspection.
\item \textbf{Inspect and correct annotations.} Figure~\ref{fig:qa-miradata} shows the manual review interface for a MiraData case; the Sekai orbit example is shown in Figure~\ref{fig:qa-sekai} in Appendix~\ref{app:additional-annotations}. The review interface presents input and query images, anchor overlays, projected input surfaces, alternative query frames, trajectory plots, and revisit pairs. Reviewers check \emph{anchor identity and mask completeness}, \emph{template fit and camera motion}, and \emph{the region shared across revisits}. They can replace an anchor, change the query selection, or reject a case. Accepted manual choices are preserved during export.
Top and side trajectory views help identify paths that rise through a ceiling, approach an observed wall, or make an abrupt turn. Reference overlays show missing object parts, neighboring objects included in a mask, and surfaces that become occluded at the query. The review record stores the chosen frame, retained anchor IDs, acceptance decisions, and comments or failure tags. These choices determine the final reference frames, anchor masks, and annotations exported for evaluation.
\item \textbf{Curate diversity and add captions.} We remove rejected candidates, group duplicates by source provenance, and balance trajectory templates while avoiding repeated use of the same scene content. A VLM annotates factual scene captions from sampled source frames for model conditioning, using the dedicated instructions in Figure~\ref{fig:prompt-caption}. The final record stores the selected references, anchors, review decisions, and source bindings. Annotation and manual curation required over $300$ hours across the collection.
\end{constructionsteps}

\subsection{Released Case Package}
\label{app:construction-records}

Each case separates model-facing inputs from evaluator-only references, with shared metadata linking both sets to the scene and specifying the camera poses and observation times used for evaluation.

\noindent\textbf{Model inputs.} The public package contains an initial RGB image, a scene caption, and a requested camera trajectory with intrinsics. Images are standardized to $1280\times720$, with matching geometric annotations, and camera poses are expressed relative to the initial view. Short and long cases contain $357$ and $961$ poses, respectively, at $16$ FPS.

\noindent\textbf{Evaluation references.} A separate reference package contains query RGB, registered depth, anchor masks and identities, and camera calibration. These annotations support comparisons at the query and revisit events and are withheld from generation. Query images are also excluded from generated-camera recovery and reconstruction context.

\noindent\textbf{Event and source metadata.} Each case records the frame indices and timestamps of the query, return, and first-visit/revisit pairs, together with their source-frame bindings. Long cases retain a link to the paired short case, preserving their common input and registered references.

\section{Evaluation Protocol and Metrics}
\label{app:metrics}

This appendix details the metrics used in each of the six evaluation tiers, including their definitions, implementation, and tier-level aggregation. We also explain score normalization and aggregation into tier scores and the overall score.

\subsection{Tier 1: Control Alignment}
\label{app:tier1}

Control alignment measures the agreement between requested and recovered camera motion. Let $P_t=[R_t\mid\mathbf p_t]$ and $\hat P_t=[\hat R_t\mid\hat{\mathbf p}_t]$ be their camera-to-world poses at frame $t$. A hat denotes a quantity recovered from generation; $E$ denotes an error and $S$ a similarity or normalized score. We align the initial poses and fit a single nonnegative scale to all camera positions recovered from the entire generated video.

\numberedmetriccard{metriccontrol}{Translation and Rotation Error} We measure the spatial and angular accuracy of the recovered camera trajectory.

\coloredstep{metriccontrol}{Translation Error.} We measure the mean distance between corresponding camera positions, normalized by the total length of the requested camera trajectory over the evaluated video:
\begin{equation}
 E_{\mathrm{trans}}=\frac{1}{NL}\sum_{t=1}^{N}\|s^*\hat{\mathbf p}_t-\mathbf p_t\|_2.
\end{equation}
Here $N$ is the number of evaluated poses, $L$ is the total requested path length, and $s^*$ minimizes the sum of squared position errors over the sequence. This alignment resolves the recovered trajectory's global scale, while the remaining error measures its spatial deviation. Pure-yaw trajectories are excluded from translation evaluation.

\coloredstep{metriccontrol}{Rotation Error.} We average the angular distance between camera orientations:
\begin{equation}
 E_{\mathrm{rot}}=\frac{1}{N}\sum_{t=1}^{N}d_R(\hat R_t,R_t).
\end{equation}
The geodesic distance $d_R$ is the angle of the relative rotation $R_t^\top\hat R_t$, reported in degrees. Both pose errors use observations at matching timestamps, so they measure the execution of the requested motion throughout the video.

\numberedmetriccard{metriccontrol}{Discrete Action Accuracy}
We extract local motion directions from the requested and recovered cameras and compare their direction sets at request-active intervals. Both sequences use identical extraction settings.

\coloredstep{metriccontrol}{Move Accuracy.} We measure camera-local displacement over two frames. An interval is active when its speed exceeds 1\% of the sequence's mean speed. Directions are selected by unit-displacement components with magnitude above 0.5, using their signs to distinguish opposite directions. Multiple directions may be active together.

\coloredstep{metriccontrol}{Look Accuracy.} We obtain the local rotation vector over four frames. Pitch and yaw components with magnitude above $0.4^\circ$ determine the active look directions, with their signs distinguishing opposite rotations. These intervals correspond to the benchmark's 16\,fps timeline.

For action type $a\in\{\mathrm{Move},\mathrm{Look}\}$, let $\mathcal U_a$ be the request-active intervals and $B^a_t,\hat B^a_t$ the requested and recovered direction sets. Accuracy requires an exact match of the full set:
\begin{equation}
 S_a=\frac{100}{|\mathcal U_a|}\sum_{t\in\mathcal U_a}\mathbf{1}[\hat B^a_t=B^a_t].
\end{equation}
We compute accuracy within each video and then average valid video scores. An action without requested activity is omitted for that case.

\textbf{\textit{Tier score.}} We convert translation and rotation errors to percentage scores and combine the applicable pose and action scores, assigning equal weight to each component evaluated for that video.

\subsection{Tier 2: Visual Quality}
\label{app:tier2}

Visual quality measures frame appearance and temporal smoothness. We adopt five components from VBench and WBench~\citep{huang2024vbench,ying2026wbench}, each expressed on a 0--100 scale:

\numberedmetriccard{metricquality}{Aesthetic Quality} The LAION aesthetic predictor~\citep{schuhmann2022laion5b} scores sampled frames on a 0--10 scale. We average the frame scores and multiply by ten.

\numberedmetriccard{metricquality}{Imaging Quality} MUSIQ~\citep{ke2021musiq} measures perceptual image quality. We average its native 0--100 scores over the sampled frames in each video, giving every sampled frame the same weight.

\numberedmetriccard{metricquality}{Temporal Flickering} We measure the mean absolute RGB difference between adjacent sampled frames and invert it as $S_{\mathrm{flicker}}=100(1-\mathrm{MAE}/255)$. Frames are sampled every two original frames, corresponding to 8\,fps for the 16\,fps videos. Higher scores indicate smaller frame-to-frame appearance changes.

\numberedmetriccard{metricquality}{Motion Smoothness} AMT predicts an intermediate frame from its two neighbors. Agreement with the observed middle frame measures local temporal continuity; we multiply the normalized consistency score by 100. Triplets use consecutive frames $(t,t+1,t+2)$, with starting indices spaced eight frames apart.

\numberedmetriccard{metricquality}{Human Preference Score} HPSv3 scores 20 uniformly sampled frames for short videos and 54 for minute-long videos, using an empty text prompt. Following WBench, we map the mean raw reward $\bar r$ as
\begin{equation}
 S_{\mathrm{HPS}}=100\,\operatorname{clip}\!\left(\frac{\bar r+5.21}{13.82},0,1\right).
\end{equation}
The fixed endpoints $-5.21$ and $8.61$ are shared across models.

\textbf{\textit{Tier score.}} The five components receive equal weight within each video, and videos are averaged equally regardless of frame count. Table~\ref{tab:quality-breakdown} reports the component scores.

\subsection{Tier 3: Input Preservation}
\label{app:tier3}

We measure scene and anchor fidelity at Return and Query, and anchor identity during exploration.
\label{app:tier42}\label{app:tier43}

\numberedmetriccard{metricinput}{Scene Appearance Fidelity}
We compare the generated frame $J$ with its reference $I$ using PSNR, SSIM, LPIPS, and DINO. The reference is the input image at Return and the registered query image at Query. We score valid pixels in region $\Omega$, restricting Query to surfaces visible in the input.

Regional DINO uses the comparison region's bounding crop. At Query, DINO instead matches patch features within a one-patch neighborhood and averages the best cosine similarities.

\numberedmetriccard{metricinput}{Scene Geometry Fidelity}
Let $D$ and $\bar D$ denote reference and generated depth, and let $\Omega_D\subseteq\Omega$ contain pixels with valid depth in both. We align generated depth to the reference with a median scale $a^*=\operatorname{median}_{u\in\Omega_D}D(u)/\bar D(u)$, then compute the following scale-aligned AbsRel over all scene pixels with valid reference and generated depth:
\begin{equation}
 E_{\mathrm{scene}}=\frac{1}{|\Omega_D|}\sum_{u\in\Omega_D}
 \frac{|a^*\bar D(u)-D(u)|}{D(u)+\epsilon}.
 \label{eq:metric-scene-depth}
\end{equation}
Here $\epsilon$ is a small positive constant. Scaling corrects global depth ambiguity but leaves local geometric errors intact, so errors on individual objects and other visible surfaces remain measurable.

\numberedmetriccard{metricinput}{Anchor Appearance Fidelity}\label{app:anchor-fidelity}
Let $A_k$ be the reference mask for anchor $k$. We evaluate it within $\Omega_k=\Omega\cap A_k$, using the same four appearance measurements as for the scene. For $N_A$ anchors and each measurement $m$, the case score is the equal-weight mean
\begin{equation}
 m_{\mathrm{anchor}}=\frac{1}{N_A}\sum_{k=1}^{N_A} m(I,J;\Omega_k).
\end{equation}
We average each measurement equally across anchors.

\numberedmetriccard{metricinput}{Anchor Geometry Fidelity}
We measure depth fidelity within each anchor using the AbsRel definition in Eq.~\ref{eq:metric-scene-depth}. For anchor $k$, the comparison region is $V_k=\Omega_k\cap\Omega_D$. We average errors over valid pixels within each anchor, then weight anchors equally. All anchors reuse the scene-level scale $a^*$, preserving relative scale differences between objects and their surroundings.

\numberedmetriccard{metricinput}{Anchor Identity Consistency}
We track each input anchor with SAM3.1~\citep{carion2025sam3} and compare its DINO descriptor with the input descriptor at one-quarter, one-half, three-quarters, and the end of the input-to-query interval. Descriptors are formed by mask-area-weighted pooling of DINO patches followed by unit normalization. With input descriptor $f_{k,0}$ and tracked descriptor $\hat f_{k,t}$, identity consistency is
\begin{equation}
 S_{\mathrm{id}}=\frac{1}{N_A}\sum_{k=1}^{N_A}
 \frac{1}{|\mathcal T_k|}\sum_{t\in\mathcal T_k}
 f_{k,0}^\top\hat f_{k,t}.
\end{equation}
Here $\mathcal T_k$ contains valid scoring times for anchor $k$. Tracking samples every fourth frame and includes all four scoring times; samples with insufficient mask or feature support are omitted.

\textbf{\textit{Tier score.}} Each checkpoint averages appearance and depth equally, with equal scene and anchor weights within each modality. Appearance averages normalized PSNR, SSIM, LPIPS, and DINO. After converting identity similarity to a percentage, we combine the three components:
\begin{equation}
 S_{\mathrm{T3}}=0.30S_{\mathrm{return}}+0.40S_{\mathrm{query}}+0.30S_{\mathrm{identity}}.
\end{equation}

\subsection{Tier 4: Generated-Content Persistence}
\label{app:tier4}

We compare generated frames at the registered first-visit and revisit times, using one pair for short cases and three for long cases. The selected pairs revisit surfaces outside the input view.

\numberedmetriccard{metricrevisit}{Revisit Appearance Consistency}
We compare DINO features of geometrically corresponding patches using Eq.~\ref{eq:metric-feature-consistency}. The mean cosine similarity measures appearance retention on co-visible surfaces.

For identical requested poses, we also compare the images directly using PSNR, LPIPS, and DINO similarity between patches at corresponding image locations. For different requested poses, the direct comparison uses global DINO descriptors pooled over the complete first-visit and revisit images.

\numberedmetriccard{metricrevisit}{Revisit Depth Consistency}
We project first-visit depth into the revisit camera and compute AbsRel on co-visible surfaces using Eq.~\ref{eq:metric-depth-consistency}. For identical requested poses, we additionally compare depth directly at corresponding pixels. Both comparisons use the same recovered global scale for the first-visit and revisit observations.

\numberedmetriccard{metricrevisit}{Revisit Surface Agreement}
For same- and different-pose revisit pairs, we back-project both depth maps into point clouds in the recovered coordinates. Precision $p_\tau$ is the fraction of first-visit points with a revisit neighbor within distance $\tau$; recall $r_\tau$ measures the reverse. Surface agreement is their harmonic mean:
\begin{equation}
 F_\tau=\frac{2p_\tau r_\tau}{p_\tau+r_\tau}.
\end{equation}
We use $\tau=\max(0.02,0.05d_{\mathrm{med}})$, where $d_{\mathrm{med}}$ is the median depth across both clouds. Points are sampled at stride eight and voxel-downsampled at $\tau/2$, retaining the sequence alignment. Nonempty clouds with no matches receive zero.

\textbf{\textit{Tier score.}} Registered pairs are averaged equally within each case. For same-pose pairs, the direct score weights RGB fidelity, patch DINO similarity, and direct depth consistency equally, with PSNR and LPIPS sharing the RGB contribution. Different-pose pairs use global DINO. The two branches are pooled by their valid case counts to obtain the direct-comparison score $S_{\mathrm{direct}}$ across the registered revisit pairs.

The geometric score averages normalized reprojected DINO, depth consistency, and surface F-score equally. The tier combines direct and geometric scores as
\begin{equation}
 S_{\mathrm{T4}}=\tfrac12(S_{\mathrm{direct}}+S_{\mathrm{geometry}}).
\end{equation}

\subsection{Tier 5: 3D Self-Consistency}
\label{app:tier5}

This tier tests whether generated views agree with a common 3D scene. Short- and medium-range frame pairs are fixed on the video timeline. Recovered depth and cameras establish correspondences between their shared visible surfaces.

\numberedmetriccard{metricgeometry}{Cross-View Depth Consistency}
We measure relative depth disagreement on surfaces visible in both views:
\begin{equation}
 E_D(s\to t)=\frac{1}{|\Omega_{s\to t}|}
 \sum_{u\in\Omega_{s\to t}}
 \frac{|\hat D_{s\to t}(u)-\hat D_t(v(u))|}
 {\hat D_t(v(u))+\epsilon}.
 \label{eq:metric-depth-consistency}
\end{equation}
Here $\hat D_t$ is the estimated depth at frame $t$, $\hat D_{s\to t}(u)$ is the source point's depth in the target camera, and $v(u)$ is its projected pixel. The set $\Omega_{s\to t}$ contains valid correspondences after image-bound and occlusion checks; $\epsilon$ is a small positive constant. Each frame pair receives the mean of the two directional errors. Lower error indicates stronger agreement about the depth of shared surfaces.

We average the two directional errors for each frame pair, then average pairs within the short- and medium-range groups. These groups use offsets of 8 and 32 frames, with at most 48 and 16 pairs, respectively. Pixels are sampled at stride four, with a relative target-occlusion tolerance of $0.10$. Short-range pairs measure local depth stability; medium-range pairs cover larger viewpoint changes.

\numberedmetriccard{metricgeometry}{Cross-View Feature Consistency}
We compare DINOv2~\citep{oquab2023dinov2} descriptors at geometrically matched locations:
\begin{equation}
 S_\phi(s\to t)=\frac{1}{|\mathcal C_{s\to t}|}
 \sum_{(a,b)\in\mathcal C_{s\to t}}
 \phi(\hat I_s,a)^\top\phi(\hat I_t,b),
 \label{eq:metric-feature-consistency}
\end{equation}
Here $\hat I_t$ is the generated RGB frame, $\phi(I,a)$ is the unit-normalized DINO descriptor for patch $a$ in image $I$, and $\mathcal C_{s\to t}$ contains unique patch pairs induced by geometric correspondences.

We evaluate at most 16 pairs in each temporal group and average both projection directions. Each matched patch pair contributes once, preventing dense pixel correspondences from overweighting the same feature pair.

\numberedmetriccard{metricgeometry}{Regional Scale Agreement}
This metric measures whether different image regions imply a consistent relative depth scale. We divide the target image into a $4\times4$ grid and compute the median ratio $\rho_j=\operatorname{median}(\hat D_t/\hat D_{s\to t})$ in each sufficiently supported cell $j$. The score is
\begin{equation}
 E_{\mathrm{scale}}=
 \operatorname{median}_{j\in\mathcal J}
 \left|\log\rho_j-\operatorname{median}_{k\in\mathcal J}\log\rho_k\right|.
\end{equation}
Here $\mathcal J$ contains sufficiently supported cells; lower values indicate more consistent regional scale.

Subtracting the median log-ratio removes a common multiplicative offset between the two depth maps. The remaining spread measures disagreement among regions. For example, if foreground furniture and a background wall require different relative scales, their cell ratios separate and the score increases. The median statistics reduce the influence of a few extreme depth samples; cells without enough correspondences are excluded.

\numberedmetriccard{metricgeometry}{Reconstruction}\label{app:tier41}
This metric evaluates whether a Gaussian scene reconstructed from generated views reproduces other views of the same video. DA3 predicts the scene from 64 context frames, which we render at 16 disjoint held-out cameras. We compare each rendered image with its corresponding generated frame using PSNR, SSIM, and LPIPS.

Context and held-out frames are uniformly sampled from the estimated sequence. DA3~\citep{lin2025da3,kerbl2023gaussians} predicts Gaussians in a feedforward pass using context RGB, recovered poses, and intrinsics. Each fidelity measure is averaged over the 16 held-out views on pixels with finite positive rendered depth.

\textbf{\textit{Tier score.}} Depth, feature, regional-scale, and reconstruction scores receive equal weight:
\begin{equation}
 S_{\mathrm{T5}}=\tfrac14(S_{\mathrm{depth}}+S_{\mathrm{feature}}+S_{\mathrm{scale}}+S_{\mathrm{GS}}).
\end{equation}
Depth and feature scores each average their normalized short- and medium-range components equally. The reconstruction score averages normalized PSNR, SSIM, and LPIPS equally.

\subsection{Tier 6: Temporal Stability}
\label{app:tier6}

We evaluate both short and minute-long videos in five-second windows, measuring both the average performance and its retention relative to the initial window. Each window uses the score mappings and full-sequence camera alignment defined for the preceding tiers. We evaluate three dimensions:

\coloredstep{metriclong}{Quality.} HPS, MUSIQ, and aesthetic quality.

\coloredstep{metriclong}{Control.} translation, rotation, Move Accuracy, and Look Accuracy over applicable components.

\coloredstep{metriclong}{Geometry.} cross-view depth, feature, and regional-scale consistency. Depth and feature each combine their two temporal groups with equal weight for the short- and medium-range comparisons.

We retain the final complete five-second window. Actions are scored on request-active intervals within each window. Each indicator requires a valid initial window and at least one later window.

\numberedmetriccard{metriclong}{Performance Level and Retention}
Let $s_{m,b}$ be the percentage score of indicator $m$ in window $b$, with $b=0$ denoting the initial window and $\mathcal B_m$ the valid later windows. The average level includes all valid windows:
\begin{equation}
 A_m=\frac{s_{m,0}+\sum_{b\in\mathcal B_m}s_{m,b}}{1+|\mathcal B_m|}.
\end{equation}
Retention measures the mean positive decline from the initial score:
\begin{equation}
 R_m=100-\frac{1}{|\mathcal B_m|}\sum_{b\in\mathcal B_m}\max(0,s_{m,0}-s_{m,b}).
\end{equation}
Later improvements contribute zero decline. Combining level and retention captures both the average scores achieved during exploration and the amount of decline relative to the initial five-second window.

\textbf{\textit{Tier score.}} Within each video, we average indicators equally within each dimension to obtain $A_d$ and $R_d$. The dimension score is
\begin{equation}
 S^{\mathrm{long}}_d=0.30A_d+0.70R_d.
\end{equation}
Tier~6 weights quality, control, and geometry equally within each video. We average videos separately by horizon, then combine the short- and long-video means with weights of 30\% and 70\%.

\subsection{Evaluation Pipeline and Score Aggregation}
\label{app:metric-setup}

The evaluator recovers geometry and constructs Gaussian and feature representations for metric comparison. Measurements are normalized and combined into tier scores and the overall score.

\coloredstep{metriclong}{Recover cameras and depth.}
We decode each generated video on its requested timeline, sample frames at stride two, and include the registered query, return, and revisit frames. DA3 Giant~1.1~\citep{lin2025da3} estimates poses, intrinsics, depth, and confidence at processing resolution 504. Sequences with at most 240 sampled frames are processed jointly. Longer sequences use 120-frame streaming chunks with 30-frame overlap and SALAD loop closure. Chunk alignment places the estimates in a common coordinate system. Each recovered observation is stored with its video-frame index for subsequent event and cross-view comparisons.

\coloredstep{metriclong}{Predict Gaussians and render held-out views.}
From the sampled sequence, we select 64 context frames and 16 disjoint held-out frames. The context RGB images and recovered cameras are passed to DA3's feedforward Gaussian prediction. Images are resized to $504\times504$ with matching intrinsics, and held-out cameras are expressed in the reconstruction's coordinates and scale. We render RGB and depth at the 16 held-out cameras with gsplat, using a black background. These renders and their corresponding generated frames supply the reconstruction measurements in Tier~5.

\coloredstep{metriclong}{Extract appearance and identity features.}
We extract DINOv2~\citep{oquab2023dinov2} global and patch descriptors from the selected images. Camera and depth correspondences identify patch pairs for cross-view and revisit comparisons. Registered reference masks define scene and anchor regions for input preservation; SAM3.1~\citep{carion2025sam3} tracks input anchors to the four identity-scoring times. PSNR, SSIM, and LPIPS compare the image pairs specified by each metric. Visual-quality predictors operate on their own frame and triplet schedules, as described in Tier~2.

\coloredstep{metriclong}{Compute event and temporal measurements.}
Recovered cameras are aligned to the requested trajectory for pose and action scoring. Depth, DINO correspondences, and point clouds provide the geometric measurements; query and return observations use registered references, while revisit measurements compare the selected generated observations. The same video estimates support five-second window measurements for Tier~6. We first reduce frame, anchor, and frame-pair observations to case-level components, then apply the score mappings and aggregation below.

\noindent\textbf{Normalization and aggregation.}\ \label{app:metric-normalization}
We convert metric components to a common 0--100 scale, with higher scores indicating better performance. Unit-range similarities are scaled directly; errors and PSNR use fixed linear mappings calibrated with equal weight per model. For Tiers~3--5, we normalize case-level measurements, average each component across valid cases, and apply the tier weights defined above. Tier~6 completes its temporal aggregation within each video before averaging across videos. The final six tier scores are combined as
\begin{equation}
 S_{\mathrm{overall}}=0.25S_{\mathrm{T1}}+0.25S_{\mathrm{T2}}
 +0.10S_{\mathrm{T3}}+0.10S_{\mathrm{T4}}+0.10S_{\mathrm{T5}}+0.20S_{\mathrm{T6}}.
\end{equation}

\raggedbottom
\section{Model Adapters and Generation Settings}
\label{app:adapters}

We group the evaluated models by the control interface used in our experiments: \textbf{camera-conditioned}, \textbf{action-conditioned}, and \textbf{text-conditioned}. For each model, we describe below its generation setting. The benchmark provides an initial image $I_0$, a scene caption, and a camera trajectory $\{P_t\}$. Adapters convert the trajectory to the model's coordinate convention and temporal sampling. Image resizing or cropping is accompanied by the corresponding intrinsic transformation when the model accepts calibration. For chunked generators, we extend the terminal camera pose to the required length and trim the extra output frames. Models with fixed generation windows continue from their own generated images or retained context. Evaluated videos use the benchmark's 16-FPS timeline; when a model uses a different native frame rate, its controls or text timings follow that native clock before output conversion.

\subsection{Camera-Conditioned Models}
\label{app:camera-models}

These models receive numerical camera conditions derived from $\{P_t\}$, expressed relative to the initial camera. We use the camera interface directly when an interactive model also offers an action-to-camera front end. This applies to Echo-WM and Matrix game 3.5; HY-WorldPlay also receives camera poses and derives its auxiliary action labels internally. Cosmos3-Nano's continuous camera increments belong to this group, although the native implementation labels these continuous numerical camera conditions as actions.

\adaptermodel{adaptercamera}{cuhksz}{SolarWM-H3}~\citep{huang2026solarwm} generates videos autoregressively with camera conditioning. We use 4 denoising steps, 5 latent frames per chunk, and a 6-chunk attention window. The requested trajectory enters its camera-conditioning path with the native \texttt{logd4} translation encoding and fixed camera intrinsics. The input image and caption initialize generation, and the history cache is carried across rollout chunks.

\adaptermodel{adaptercamera}{nvidia}{Lyra2}~\citep{shen2026lyra2} combines video generation with geometric conditioning. We use the DMD-distilled checkpoint with 4 steps at $832\times480$. Requested poses are rebased to $P_0$, converted to world-to-camera matrices, and passed through the custom-trajectory interface with pose scale 1. Intrinsics follow the image resize, while the native geometry pipeline aligns estimated depth to MoGe scale. Short runs generate 401 frames and retain the first 357.

\adaptermodel{adaptercamera}{alaya}{EVOKE}~\citep{yin2026evoke} uses camera and geometric conditions for autoregressive scene exploration. We use the Stage3 post-distillation checkpoint in its image-conditioned mode, with 3 steps, guidance scale 1, and $640\times384$ output. The initial image, caption, and requested cameras enter the native camera and geometry branches. Short sequences use 10 rollout chunks; longer sequences extend the same rollout with the corresponding camera windows.

\adaptermodel{adaptercamera}{jd}{Echo-WM}~\citep{zhang2026echowm} is a causal video generator with unified camera conditioning. We use the Flash checkpoint with 4 steps, 3 latent frames per chunk, and $1280\times704$ output. The adapter bypasses its Action DSL and supplies initial-camera-relative c2w poses directly to UCPE, applying the native translation factor $1/30$ and 8-frame camera sampling. Short and long runs share the same causal history mechanism, with a 19-frame latent cache and 7 sink frames carried from one rollout chunk to the next during generation.

\adaptermodel{adaptercamera}{nvidia}{Gen3C}~\citep{ren2025gen3c} uses a persistent 3D cache to condition subsequent generated views. We use 35 denoising steps at $1280\times704$. The input image initializes the scene and the requested camera sequence drives the native autoregressive generation path. Inference proceeds in 121-frame windows with 1 shared boundary frame, retaining the 3D cache across windows.

\adaptermodel{adaptercamera}{skywork}{Matrix game 3.5}~\citep{qian2026matrixgame35} combines autoregressive generation with geometric patch memory. We use the distilled first-person model with 3 denoising steps and $1280\times704$ output. The camera adapter supplies c2w extrinsics and image-aligned pixel intrinsics directly. The rollout consumes consecutive camera windows and retains its native memory of generated views.

\adaptermodel{adaptercamera}{alaya}{AlayaWorld}~\citep{team2026alayaworld} uses ViGeo to condition generation on scene geometry and camera motion. We use v1.1 Stage3, combining the Stage2b base with the Stage3 LoRA, at $960\times544$ with 4 steps and guidance scale 1. The adapter supplies poses relative to $P_0$ with scale 1 and intrinsics adjusted to the resized input. Short videos use 12 rollout rounds; the same interface processes the extended trajectories.

\adaptermodel{adaptercamera}{robbyant}{LingBot-World v2}~\citep{gao2026lingbot2} generates camera-conditioned video with causal attention and cached history. We use the causal-fast checkpoint with 4 denoising steps, 4 latent frames per chunk, and the 480p configuration. Requested poses and intrinsics enter its native camera preprocessing, including its relative-motion normalization. For short cases, terminal padding produces 365 frames, of which the first 357 are retained.

\adaptermodel{adaptercamera}{tencent}{HY-WorldPlay}~\citep{sun2025worldplay} supports autoregressive generation from camera and action conditions. We use its AR-distilled checkpoint with 4 steps, chunk size 4, and $848\times480$ output. The adapter supplies sampled c2w poses from the requested trajectory; the native pipeline converts them to its camera representation and derives the associated action labels. Thus, our input uses the numerical camera interface. Short generation produces 365 frames before trimming to 357; native 24-FPS output is played at 16 FPS.

\adaptermodel{adaptercamera}{nvidia}{SANA-WM}~\citep{zhu2026sanawm} provides streaming video generation with camera conditioning. We use the streaming checkpoint, 4 generator steps followed by 3 refiner steps, guidance scale 1, and flow shift 8. The requested poses are rebased to $P_0$ and converted into the native ray and Pl\"ucker conditions. The $1280\times720$ input is cropped to $1280\times704$ to satisfy spatial divisibility, with intrinsics updated accordingly. Streaming generation retains the model's causal decoding and refinement stages.

\adaptermodel{adaptercamera}{tencent}{Voyager}~\citep{huang2025voyager} generates new views conditioned on an image and camera motion. We use 50 steps, and $768\times512$ output. The requested trajectory is divided into 49-frame generation windows with a 1-frame overlap. Each window receives the corresponding camera segment and starts from the preceding generated last frame, extending the model to the benchmark's short and long durations.

\adaptermodel{adaptercamera}{robbyant}{LingBot-World v1}~\citep{gao2026lingbot} is evaluated with its Fast camera-control checkpoint and 4-step causal sampling. The adapter provides the initial image, scene caption, requested poses, and intrinsics to the native camera pipeline. Short runs retain the full available history; long runs use a 21-frame local attention window and 1 sink frame. The evaluated batches use the recorded 720p and 480p configurations, with the same pose-conditioning interface.

\adaptermodel{adaptercamera}{nvidia}{Cosmos3-Nano}~\citep{nvidia2026cosmos3} accepts continuous camera-motion conditions. We use 30 denoising steps. Consecutive requested poses are converted to local SE(3) increments, encoded as 3 translation components scaled by 10 and a 6-dimensional rotation representation. These 9-dimensional vectors are supplied directly to the camera-action input. Generation adds 60 frames per chunk, conditioned on the preceding generated last frame, at an inference frame rate of 16 FPS.

\adaptermodel{adaptercamera}{amap}{DreamX-World}~\citep{team2026dreamx} is evaluated with the 5B autoregressive checkpoint and 4 denoising steps at $1280\times704$. We use 3 latent frames per block, and the native 4-step schedule. Requested poses and normalized intrinsics pass through the official chunk-relative camera helper. Each chunk's cameras are expressed relative to the preceding chunk endpoint, while calibration follows the benchmark input image.

\adaptermodel{adaptercamera}{amap}{FantasyWorld}~\citep{fantasy2025world} generates camera-conditioned video from an initial image. We use 50 denoising steps, and $592\times336$ output. The requested trajectory is partitioned into 81-frame windows with a 1-frame overlap. Each window receives camera conditions for its trajectory segment and starts from the preceding window's final generated frame, continuing generation from the model's own output.

\adaptermodel{adaptercamera}{tsinghua}{Astra}~\citep{zhu2025astra} uses relative camera motion for blockwise video generation. We use 50 steps, and $832\times480$ output. The adapter converts the requested trajectory to consecutive world-to-camera increments at the native 4-frame sampling interval; this interface has no intrinsic conditioning. Camera windows advance with the history origin, and long runs extend the temporal indexing while retaining the native recent-history context. Short runs retain 357 frames from a 385-frame generation.

\subsection{Action-Conditioned Models}
\label{app:action-models}

ABot-World, Genie3, and HappyOyster receive discrete movement and look commands. We compile these commands from the requested camera trajectory and execute them through the model's streaming or browser interface.

\noindent\textbf{Pose-to-action conversion.}{\ }
We smooth reconstructed camera paths while retaining the initial, query, return, and revisit waypoints. The adapter regularizes positions and continuous-sign quaternions with a nominal 0.2-Hz smoothing setting. For camera-to-world poses $P_t$, the local increment $\Delta P_t=P_t^{-1}P_{t+1}$ supplies translation directions and yaw/pitch increments. After removing near-stationary motion, we normalize translation directions and use signed horizontal and forward components above 0.5 to identify movement keys. A direction is selected when its net vote exceeds half the action window. Movement and turning can be active together; yaw-return cases disable translation. We use model-specific turning responses to convert requested yaw and pitch into discrete action blocks or timed key presses.

\noindent\textbf{Control evaluation.}{\ }
For ABot-World, Genie3, and HappyOyster, converting camera trajectories into discrete actions introduces differences in motion scale, timing, and turning response. Their results therefore serve primarily as diagnostic evaluations under the adapted action interface.

These interfaces cannot accept camera poses or directly specify the requested displacement and rotation at each frame. Comparing DA3-recovered poses with the reference trajectory would therefore mix directional control with differences in action quantization and motion scale. For these 3 models, we use only \textbf{Move Accuracy} and \textbf{Look Accuracy} in the reported control score. DA3 recovers motion from the generated video, and we compare its movement and look direction sets with those derived from the original requested trajectory, following Appendix~\ref{app:tier1}. Accuracy is computed over request-active intervals within each video and then averaged across videos. Tier~1 averages the 2 action scores; the control dimension of Tier~6 uses the same components within each temporal window.

\begin{figure}[t]
\centering
\tightpromptcaption
\begin{promptbody}
\begin{tikzpicture}
\node[orbia t2v prompt=orange] {\begin{minipage}{\dimexpr\linewidth-24pt\relax}
\ttfamily\fontsize{9}{10.5}\selectfont\raggedright
\textbf{Segment 1 (15.0 seconds).} initial image $I_0$.\par\smallskip
$\langle$Scene caption$\rangle$\par\smallskip
Use one continuous shot with a fixed focal length.\par
The times below are local to this video segment.\par
Move at a relaxed, steady walking pace, easing smoothly into stops and reversals.\par
From 0.0 to 7.5 seconds, move steadily to the right.\par
From 7.5 to 15.0 seconds, reverse direction and move steadily to the left.\par
By 15.0 seconds in this segment, smoothly finish the specified camera movement and come to a stop.\par
Preserve the scene layout and objects throughout, including when revisiting. Do not cut or zoom.\par
\end{minipage}};
\end{tikzpicture}
\end{promptbody}
\caption{\textbf{Kling3.0 prompt for short-horizon lateral return.} The scene caption is represented by a placeholder.}
\label{fig:kling-prompt-short}
\end{figure}

\begin{figure}[t]
\centering
\tightpromptcaption
\begin{promptbody}
\begin{tikzpicture}
\node[orbia t2v prompt=teal] {\begin{minipage}{\dimexpr\linewidth-24pt\relax}
\ttfamily\fontsize{9}{10.5}\selectfont\raggedright
\textbf{Segment 1 (15.0 seconds).} initial image $I_0$.\par\smallskip
$\langle$Scene caption$\rangle$\par\smallskip
Use one continuous shot with a fixed focal length.\par
The times below are local to this video segment.\par
Move at a relaxed, steady walking pace, easing smoothly into stops and reversals.\par
From 0.0 to 15.0 seconds, move steadily to the right.\par
Preserve the scene layout and objects throughout, including when revisiting. Do not cut or zoom.\par
\par\medskip
\textbf{Segment 2 (15.0 seconds).} preceding generated last frame.\par\smallskip
$\langle$Scene caption$\rangle$\par\smallskip
Use one continuous shot with a fixed focal length.\par
Continue from the supplied image.\par
The times below are local to this video segment.\par
Move at a relaxed, steady walking pace, easing smoothly into stops and reversals.\par
From 0.0 to 5.0 seconds, move steadily to the right.\par
From 5.0 to 15.0 seconds, reverse direction and move steadily to the left.\par
Preserve the scene layout and objects throughout, including when revisiting. Do not cut or zoom.\par
\par\medskip
\textbf{Segment 3 (10.0 seconds).} preceding generated last frame.\par\smallskip
$\langle$Scene caption$\rangle$\par\smallskip
Use one continuous shot with a fixed focal length.\par
Continue from the supplied image.\par
The times below are local to this video segment.\par
Move at a relaxed, steady walking pace, easing smoothly into stops and reversals.\par
From 0.0 to 10.0 seconds, move steadily to the left.\par
By 10.0 seconds in this segment, smoothly finish the specified camera movement and come to a stop.\par
Preserve the scene layout and objects throughout, including when revisiting. Do not cut or zoom.\par
\end{minipage}};
\end{tikzpicture}
\end{promptbody}
\caption{\textbf{Kling3.0 prompt for long-horizon lateral return.} The same scene caption precedes each segment. Continuation requests use the preceding generated last frame.}
\label{fig:kling-prompt-long}
\end{figure}

\begin{table}[t]
\centering
\begin{tablebody}{3pt}{1.2}

\caption{\textbf{Tier 1: Detailed control alignment results.} Scores use a 0--100 scale ($\uparrow$). Models are ordered by Tier Score within each group. \ranklegend{orbiaRankFirst}{1st}, \ranklegend{orbiaRankSecond}{2nd}, \ranklegend{orbiaRankThird}{3rd} mark the top three scores per column and group.}
\label{tab:control-breakdown}
\fontsize{7.5}{9}\selectfont
\middlecolumns
\begin{tabularx}{\linewidth}{@{}lc*{5}{>{\centering\arraybackslash}X}@{}}
\toprule
\rowcolor{orbiaTableHeader}
\textbf{Model} & {\bfseries\shortstack{3D\\Repr.}} & \tablehead{Translation\\Score} & \tablehead{Rotation\\Score} & \tablehead{Move\\Accuracy} & \tablehead{Look\\Accuracy} & {\bfseries\shortstack{Tier\\Score}}\\
\midrule
\rowcolor{orbiaTableHeader}
\multicolumn{7}{l}{\textbf{World Models}}\\
\modellogo{nvidia}Lyra2 & $\checkmark$ & \scorefirst{$\mathbf{79.89}$} & \scorethird{$\mathbf{75.61}$} & \scorefirst{$\mathbf{83.46}$} & \scoresecond{$\mathbf{87.59}$} & \scorefirst{$\mathbf{81.64}$}\\
\rowcolor{orbiaTableStripe}
\modellogo{cuhksz}SolarWM-H3 & $\times$ & \scorethird{$\mathbf{71.12}$} & \scorefirst{$\mathbf{84.43}$} & \scorethird{$\mathbf{74.90}$} & \scorefirst{$\mathbf{89.27}$} & \scoresecond{$\mathbf{79.93}$}\\
\modellogo{nvidia}Gen3C & $\checkmark$ & \scoresecond{$\mathbf{72.56}$} & $74.52$ & $70.11$ & \scorethird{$\mathbf{84.86}$} & \scorethird{$\mathbf{75.51}$}\\
\rowcolor{orbiaTableStripe}
\modellogo{jd}Echo-WM & $\times$ & $60.71$ & $74.74$ & $64.29$ & $83.21$ & $70.74$\\
\modellogo{alaya}EVOKE & $\checkmark$ & $63.92$ & $61.26$ & $65.96$ & $74.75$ & $66.47$\\
\rowcolor{orbiaTableStripe}
\modellogo{skywork}Matrix game 3.5 & $\checkmark$ & $46.41$ & \scoresecond{$\mathbf{77.14}$} & $51.94$ & $81.22$ & $64.18$\\
\modellogo{alaya}AlayaWorld & $\checkmark$ & $61.74$ & $56.93$ & $59.59$ & $68.45$ & $61.68$\\
\rowcolor{orbiaTableStripe}
\modellogo{tencent}Voyager & $\checkmark$ & $49.68$ & $61.42$ & $61.35$ & $71.22$ & $60.92$\\
\modellogo{alibaba}HappyOyster & -- & $(42.64)$ & $(38.00)$ & $61.11$ & $59.86$ & $60.49$\\
\rowcolor{orbiaTableStripe}
\modellogo{nvidia}SANA-WM & $\times$ & $51.39$ & $57.35$ & $56.16$ & $72.65$ & $59.39$\\
\modellogo{robbyant}LingBot-World v2 & $\times$ & $50.94$ & $43.22$ & \scoresecond{$\mathbf{79.45}$} & $60.30$ & $58.48$\\
\rowcolor{orbiaTableStripe}
\modellogo{deepmind}Genie3 & -- & $(52.47)$ & $(35.42)$ & $68.97$ & $41.55$ & $55.26$\\
\modellogo{robbyant}LingBot-World v1 & $\times$ & $49.03$ & $32.31$ & $70.08$ & $45.08$ & $49.12$\\
\rowcolor{orbiaTableStripe}
\modellogo{amap}DreamX-World-5B & $\times$ & $45.98$ & $46.95$ & $57.06$ & $45.96$ & $48.99$\\
\modellogo{amap}ABot-World & $\times$ & $(43.76)$ & $(46.14)$ & $62.13$ & $31.99$ & $47.06$\\
\rowcolor{orbiaTableStripe}
\modellogo{tencent}HY-WorldPlay & $\times$ & $49.67$ & $48.21$ & $59.43$ & $23.72$ & $45.26$\\
\modellogo{amap}FantasyWorld & $\checkmark$ & $30.05$ & $46.90$ & $36.51$ & $43.17$ & $39.16$\\
\rowcolor{orbiaTableStripe}
\modellogo{nvidia}Cosmos3-Nano & $\times$ & $34.79$ & $38.92$ & $47.42$ & $21.65$ & $35.70$\\
\modellogo{tsinghua}Astra & $\times$ & $18.68$ & $33.16$ & $18.82$ & $43.12$ & $28.45$\\
\midrule
\rowcolor{orbiaTableGroupHeader}
\multicolumn{7}{l}{\textbf{TI2V Models}}\\
\modellogo{bytedance}Seedance2.5 & -- & \scorefirst{$\mathbf{44.08}$} & \scorethird{$\mathbf{38.00}$} & \scoresecond{$\mathbf{58.74}$} & \scorefirst{$\mathbf{32.11}$} & \scorefirst{$\mathbf{43.23}$}\\
\rowcolor{orbiaTableGroupStripe}
\modellogo{kuaishou}Kling3.0 & -- & \scoresecond{$\mathbf{37.62}$} & $35.92$ & \scorethird{$\mathbf{53.22}$} & $27.57$ & \scoresecond{$\mathbf{38.58}$}\\
\modellogo{tencent}HunyuanVideo1.5 & $\times$ & \scorethird{$\mathbf{32.56}$} & \scoresecond{$\mathbf{39.11}$} & \scorefirst{$\mathbf{59.40}$} & $13.29$ & \scorethird{$\mathbf{36.09}$}\\
\rowcolor{orbiaTableGroupStripe}
\modellogo{minimax}MiniMax-H3 & -- & $28.15$ & $24.64$ & $49.15$ & $27.96$ & $32.48$\\
\modellogo{alibaba}Wan2.7 & -- & $26.89$ & $30.00$ & $37.01$ & \scoresecond{$\mathbf{31.36}$} & $31.31$\\
\rowcolor{orbiaTableGroupStripe}
\modellogo{nvidia}Cosmos3-Super & -- & $23.37$ & \scorefirst{$\mathbf{41.87}$} & $28.81$ & $17.61$ & $27.92$\\
\modellogo{lightricks}LTX2.5 & -- & $21.97$ & $27.16$ & $27.63$ & \scorethird{$\mathbf{29.27}$} & $26.51$\\
\rowcolor{orbiaTableGroupStripe}
\modellogo{meituan}LongCat & $\times$ & $22.42$ & $18.95$ & $27.76$ & $25.91$ & $23.76$\\
\bottomrule
\end{tabularx}
\end{tablebody}
\end{table}

\adaptermodel{adapteraction}{amap}{ABot-World}~\citep{jiang2026abotworld} uses the 0-5B-LF streaming checkpoint with 4 steps and $1280\times704$ output. Its input consists of $I_0$, the scene caption, and a sequence of 8-dimensional WASD/IJKL action vectors. Before the benchmark run, we generated 64 controlled clips covering static, translation, and yaw actions, followed by 8 pitch clips. DA3 recovered cameras from these generated clips to estimate turning responses. We adopted symmetric nominal increments of $20^\circ$ for yaw and $17^\circ$ for pitch per action block. A turn is issued only if the accumulated requested angle and the nominal emitted angle differ by at least half a block's increment, and the resulting turn points in the same direction as the requested camera rotation for that action block.

The native decoder produces 9 frames in the initial block and 12 thereafter. We therefore use 30 blocks for a 357-frame short video and 81 blocks for a 969-frame long rollout, retaining the first 961 frames. Each case initializes a fresh stream from $I_0$; actions are updated before generating each block, while the model retains its history within the case. The native 12-FPS frame sequence is published at 16 FPS.

\begin{figure}[t]
\centering
\includegraphics[width=\linewidth]{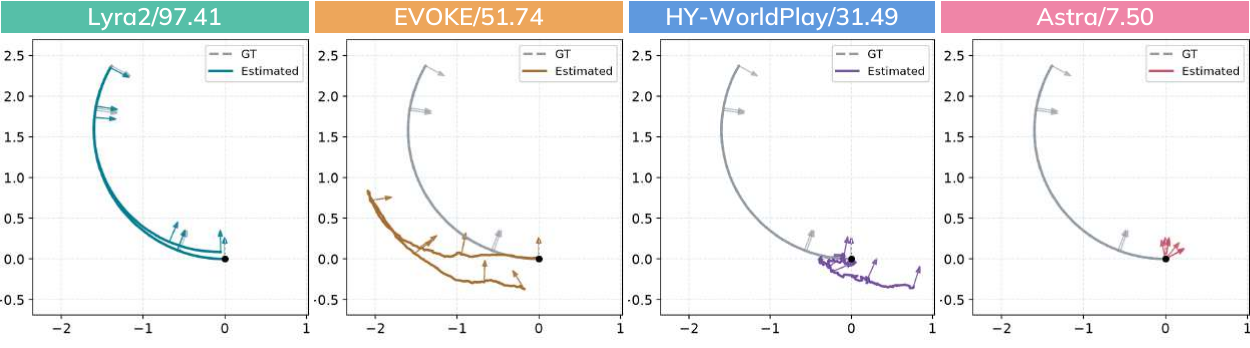}
\caption{\textbf{Control alignment on a UE orbit example.} Requested and recovered camera trajectories are shown on shared axes, with arrows indicating viewing direction. Astra fails to recover the requested trajectory. Displayed scores are for this case.}
\label{fig:tier1-example1}
\end{figure}

\begin{figure}[t]
\centering
\includegraphics[width=\linewidth]{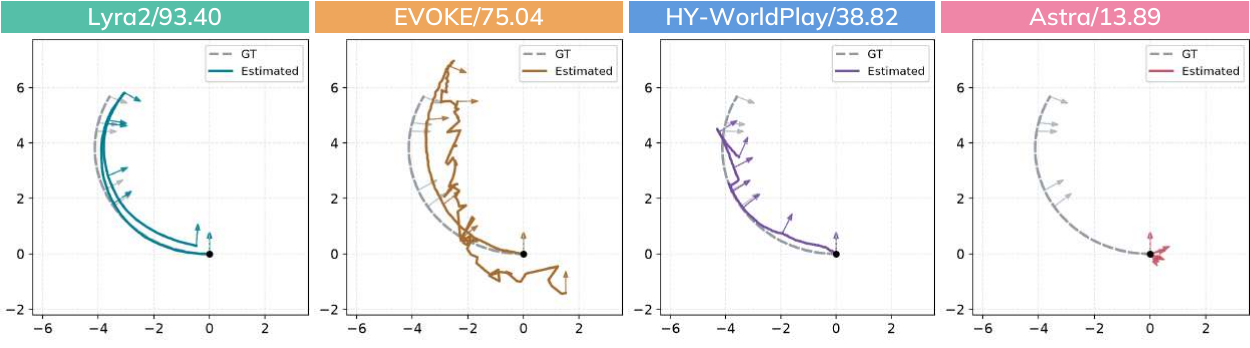}
\caption{\textbf{Control alignment on a UE orbit example.} Requested and recovered camera trajectories are shown on shared axes, with arrows indicating viewing direction. Astra fails to recover the requested trajectory. Displayed scores are for this case.}
\label{fig:tier1-example2}
\end{figure}

\adaptermodel{adapteraction}{deepmind}{Genie3}~\citep{parkerholder2025genie3} is run through the Project Genie webpage in first-person mode. We upload $I_0$ and the scene description, create the world, and begin the action schedule once the interactive view is ready. The trajectory is summarized in 12-frame analysis windows. Movement directions become WASD states, while yaw and pitch become timed look commands. A separate turning pilot checked response directions and approximate angular motion; the final scheduler uses nominal yaw and pitch rates of $70^\circ$/s and $40^\circ$/s. These rates determine hold durations, with translation and look commands dispatched independently.

Browser-side scheduling uses an absolute clock to issue key-down and key-up events while the page remains in the foreground. Short and long cases use 20- and 54-second action schedules, followed by a 2-second key-release interval. The runner exits the world and saves the webpage's downloadable video, together with the planned and dispatched actions. For evaluation, we normally remove the final 2 seconds and uniformly select 357 or 961 frames, then encode them at 16 FPS. 

\adaptermodel{adapteraction}{alibaba}{HappyOyster} is run through its browser SDK in first-person Adventure mode at 720p. We initialize the world with $I_0$ and the scene description, wait for the video stream, and execute a local action plan. A turning pilot was analyzed with DA3 to estimate visible angular response. We use nominal yaw and pitch rates of approximately $18.23^\circ$/s and $20.97^\circ$/s to allocate turn durations. Local motion is grouped into 12-frame analysis windows and mapped to a 15-second short or 40-second long interaction schedule.

The SDK maintains separate held states for translation and rotation. Our scheduler updates a state only when it changes, so releasing a turn does not interrupt ongoing translation. We record the incoming video track with MediaRecorder, retain the action and session timestamps, and release all controls before closing the session. Evaluated outputs retain the first 357 or 961 captured frames and play them at 16 FPS. 

\subsection{TI2V Models}
\label{app:text-models}

TI2V models receive $I_0$ and a prompt combining the scene caption with a motion script derived from the requested trajectory. We first construct the script, then divide it according to each model's generation windows. This keeps the motion instructions consistent with the frames generated by each request. Prompt processing and enhancement follow the model-specific inference configurations.

\begin{table}[t]
\centering
\begin{tablebody}{3pt}{1.2}

\caption{\textbf{Tier 2: Detailed visual quality results.} Scores use a 0--100 scale ($\uparrow$). Models are ordered by Tier Score within each group. \ranklegend{orbiaRankFirst}{1st}, \ranklegend{orbiaRankSecond}{2nd}, \ranklegend{orbiaRankThird}{3rd} mark the top three scores per column and group.}
\label{tab:quality-breakdown}
\fontsize{7.5}{9}\selectfont
\middlecolumns
\begin{tabularx}{\linewidth}{@{}lc*{6}{>{\centering\arraybackslash}X}@{}}
\toprule
\rowcolor{orbiaTableHeader}
\textbf{Model} & {\bfseries\shortstack{3D\\Repr.}} & \tablehead{Aesthetic\\Quality} & \tablehead{Imaging\\Quality} & \tablehead{Temporal\\Flickering} & \tablehead{Motion\\Smoothness} & \textbf{HPS} & {\bfseries\shortstack{Tier\\Score}}\\
\midrule
\rowcolor{orbiaTableHeader}
\multicolumn{8}{l}{\textbf{World Models}}\\
\modellogo{alaya}EVOKE & $\checkmark$ & \scoresecond{$\mathbf{53.23}$} & $72.29$ & $94.48$ & $98.69$ & \scorefirst{$\mathbf{63.69}$} & \scorefirst{$\mathbf{76.48}$}\\
\rowcolor{orbiaTableStripe}
\modellogo{robbyant}LingBot-World v2 & $\times$ & \scorethird{$\mathbf{53.01}$} & $73.54$ & $91.95$ & $97.05$ & \scoresecond{$\mathbf{61.14}$} & \scoresecond{$\mathbf{75.34}$}\\
\modellogo{jd}Echo-WM & $\times$ & \scorefirst{$\mathbf{53.36}$} & \scorefirst{$\mathbf{74.72}$} & $93.73$ & $97.14$ & $56.17$ & \scorethird{$\mathbf{75.02}$}\\
\rowcolor{orbiaTableStripe}
\modellogo{amap}DreamX-World-5B & $\times$ & $52.48$ & $71.27$ & $93.09$ & $98.03$ & $59.94$ & $74.96$\\
\modellogo{cuhksz}SolarWM-H3 & $\times$ & $51.18$ & \scorethird{$\mathbf{74.21}$} & $93.90$ & $97.86$ & $56.62$ & $74.76$\\
\rowcolor{orbiaTableStripe}
\modellogo{deepmind}Genie3 & -- & $49.71$ & $71.58$ & $92.67$ & $97.27$ & \scorethird{$\mathbf{60.95}$} & $74.43$\\
\modellogo{nvidia}Cosmos3-Nano & $\times$ & $52.18$ & $70.36$ & \scorefirst{$\mathbf{97.16}$} & \scorefirst{$\mathbf{99.18}$} & $52.37$ & $74.25$\\
\rowcolor{orbiaTableStripe}
\modellogo{skywork}Matrix game 3.5 & $\checkmark$ & $51.80$ & \scoresecond{$\mathbf{74.29}$} & $92.29$ & $98.25$ & $53.50$ & $74.03$\\
\modellogo{robbyant}LingBot-World v1 & $\times$ & $52.24$ & $69.32$ & $93.55$ & $97.15$ & $56.95$ & $73.84$\\
\rowcolor{orbiaTableStripe}
\modellogo{nvidia}SANA-WM & $\times$ & $52.21$ & $69.37$ & $94.37$ & \scorethird{$\mathbf{98.92}$} & $51.24$ & $73.22$\\
\modellogo{tencent}HY-WorldPlay & $\times$ & $49.17$ & $70.69$ & $95.08$ & $98.90$ & $46.46$ & $72.06$\\
\rowcolor{orbiaTableStripe}
\modellogo{nvidia}Lyra2 & $\checkmark$ & $47.52$ & $71.92$ & $92.74$ & $98.20$ & $46.45$ & $71.37$\\
\modellogo{alaya}AlayaWorld & $\checkmark$ & $49.27$ & $70.82$ & $92.72$ & $98.11$ & $44.30$ & $71.04$\\
\rowcolor{orbiaTableStripe}
\modellogo{amap}FantasyWorld & $\checkmark$ & $50.85$ & $64.34$ & \scorethird{$\mathbf{95.75}$} & \scoresecond{$\mathbf{98.94}$} & $43.64$ & $70.70$\\
\modellogo{alibaba}HappyOyster & -- & $46.80$ & $64.92$ & $94.31$ & $98.47$ & $39.38$ & $68.78$\\
\rowcolor{orbiaTableStripe}
\modellogo{tencent}Voyager & $\checkmark$ & $50.65$ & $62.12$ & $92.36$ & $97.56$ & $37.55$ & $68.05$\\
\modellogo{nvidia}Gen3C & $\checkmark$ & $47.65$ & $63.27$ & $94.02$ & $98.82$ & $33.98$ & $67.55$\\
\rowcolor{orbiaTableStripe}
\modellogo{amap}ABot-World & $\times$ & $48.79$ & $60.24$ & $95.43$ & $97.71$ & $32.66$ & $66.97$\\
\modellogo{tsinghua}Astra & $\times$ & $41.78$ & $51.86$ & \scoresecond{$\mathbf{95.87}$} & $98.45$ & $18.52$ & $61.30$\\
\midrule
\rowcolor{orbiaTableGroupHeader}
\multicolumn{8}{l}{\textbf{TI2V Models}}\\
\modellogo{nvidia}Cosmos3-Super & -- & \scorefirst{$\mathbf{55.69}$} & $74.55$ & \scorefirst{$\mathbf{96.26}$} & $99.16$ & \scorefirst{$\mathbf{69.51}$} & \scorefirst{$\mathbf{79.04}$}\\
\rowcolor{orbiaTableGroupStripe}
\modellogo{minimax}MiniMax-H3 & -- & \scorethird{$\mathbf{54.16}$} & \scoresecond{$\mathbf{76.43}$} & $93.55$ & $98.70$ & \scoresecond{$\mathbf{67.59}$} & \scoresecond{$\mathbf{78.09}$}\\
\modellogo{lightricks}LTX2.5 & -- & $53.94$ & $74.33$ & $95.22$ & \scoresecond{$\mathbf{99.35}$} & \scorethird{$\mathbf{66.95}$} & \scorethird{$\mathbf{77.96}$}\\
\rowcolor{orbiaTableGroupStripe}
\modellogo{kuaishou}Kling3.0 & -- & $53.47$ & $74.34$ & $94.93$ & $98.83$ & $65.35$ & $77.38$\\
\modellogo{meituan}LongCat & $\times$ & \scoresecond{$\mathbf{54.38}$} & \scorefirst{$\mathbf{76.91}$} & $92.10$ & $98.36$ & $64.73$ & $77.30$\\
\rowcolor{orbiaTableGroupStripe}
\modellogo{tencent}HunyuanVideo1.5 & $\times$ & $52.26$ & $74.05$ & \scorethird{$\mathbf{95.33}$} & \scorefirst{$\mathbf{99.40}$} & $64.32$ & $77.07$\\
\modellogo{bytedance}Seedance2.5 & -- & $53.52$ & $73.13$ & \scoresecond{$\mathbf{95.65}$} & \scorethird{$\mathbf{99.27}$} & $63.26$ & $76.97$\\
\rowcolor{orbiaTableGroupStripe}
\modellogo{alibaba}Wan2.7 & -- & $51.93$ & \scorethird{$\mathbf{75.64}$} & $93.11$ & $97.47$ & $64.14$ & $76.46$\\
\bottomrule
\end{tabularx}
\end{tablebody}
\end{table}

\begin{figure}[t]
\centering
\includegraphics[width=\linewidth]{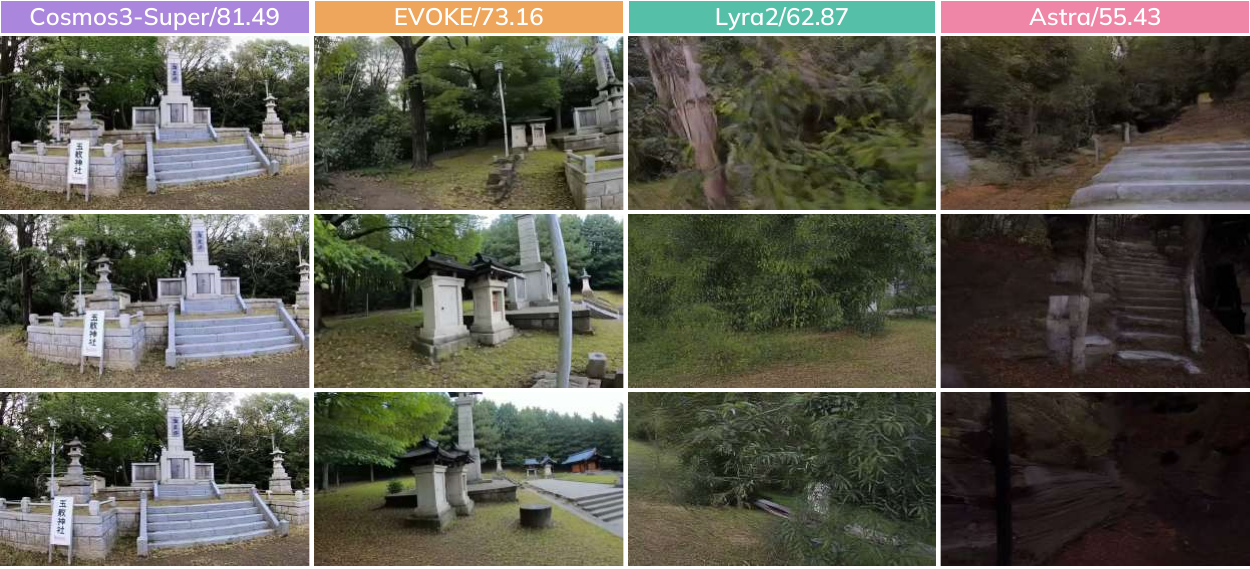}
\caption{\textbf{Visual quality on a MiraData orbit example.} Rows show three matched timestamps, with models arranged in columns. Displayed scores are for this case.}
\label{fig:tier2-example1}
\end{figure}

\begin{figure}[t]
\centering
\includegraphics[width=\linewidth]{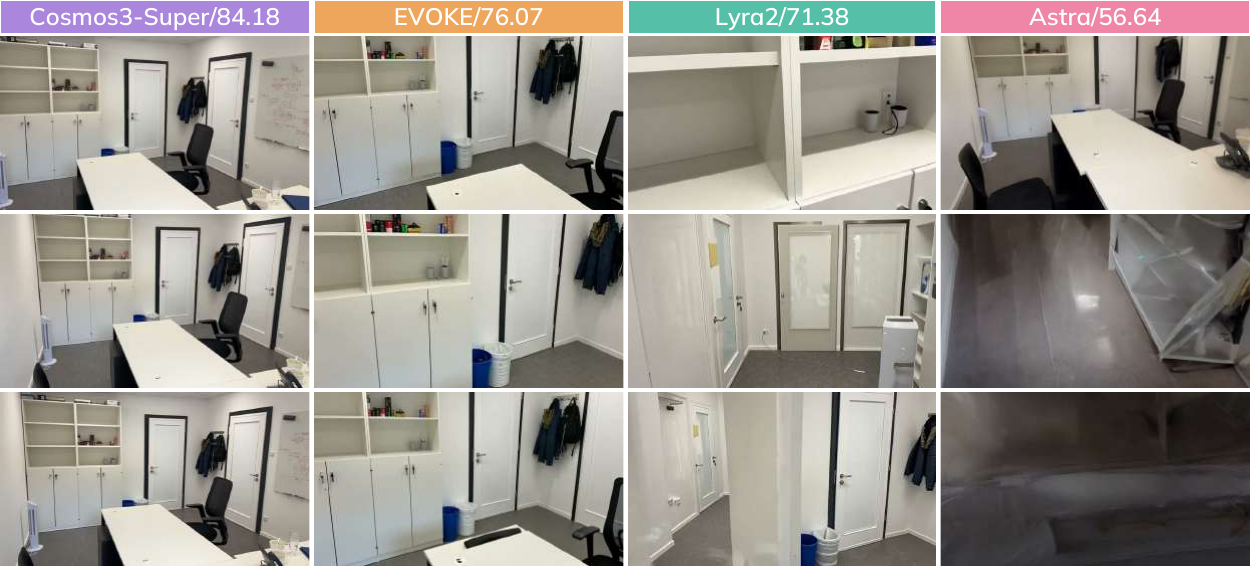}
\caption{\textbf{Visual quality on a ScanNet++ forward--backward example.} Rows show three matched timestamps, with models arranged in columns. Displayed scores are for this case.}
\label{fig:tier2-example2}
\end{figure}

\begin{figure}[t]
\centering
\includegraphics[width=\linewidth]{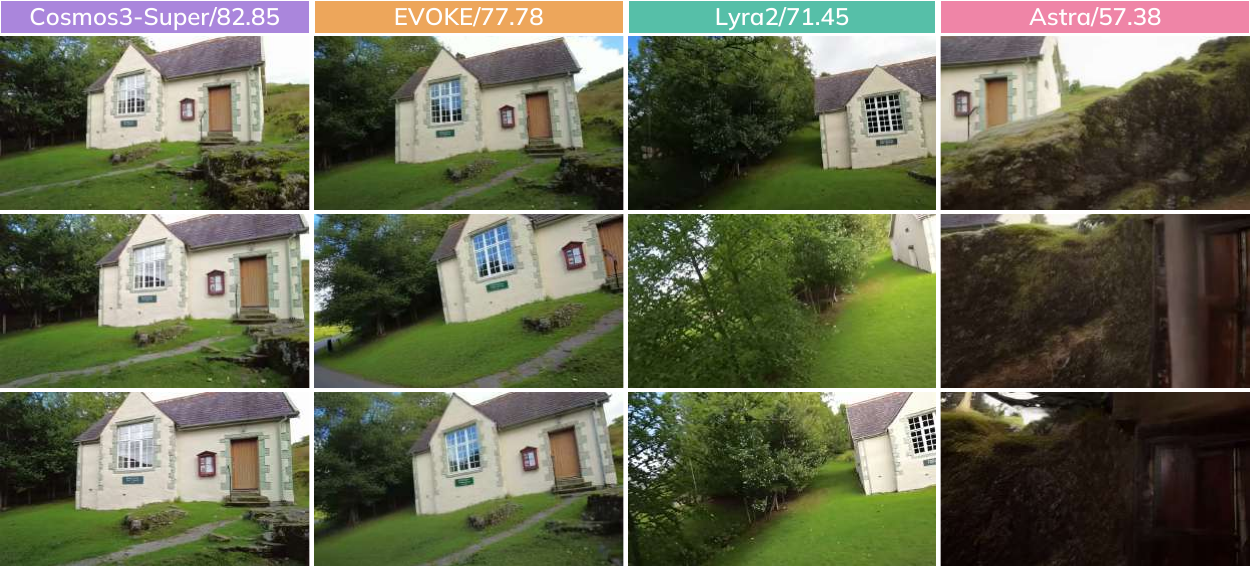}
\caption{\textbf{Visual quality on a SpatialVID lateral-return example.} Rows show three matched timestamps, with models arranged in columns. Displayed scores are for this case.}
\label{fig:tier2-example3}
\end{figure}

\noindent\textbf{From poses to motion descriptions.}{\ }
We analyze translation and rotation separately, smooth their trends over approximately 0.6 seconds, and group them into motion phases: travel, turning, pauses, reversal, and return. The script states the direction and duration of each phase, with turn angles derived from the corresponding poses. For curved trajectories, the script describes both the path and the viewing direction; orbit prompts name the public anchor when available. Brief fluctuations are merged into longer phases, while meaningful pauses and direction changes are retained. This simplification changes the text description, leaving the requested trajectory used for evaluation unchanged.

\noindent\textbf{Window-specific prompts.}{\ }
Each generation window receives only the motion phases that overlap its frame range. We recompute its local times and turning angles at the model's native frame rate, then append the resulting motion text to the scene caption. A continuation prompt starts from its supplied generated image or video context and describes the next part of the trajectory. Return instructions appear when the corresponding return phase falls within that window. Figures~\ref{fig:kling-prompt-short} and~\ref{fig:kling-prompt-long} show the short- and long-horizon scripts for the same lateral-return case in Kling3.0.

\noindent\textbf{Segmented generation.}{\ }
The first request is conditioned on $I_0$. Subsequent requests use the preceding generated last frame, except for LongCat, which retains 13 context frames. We remove repeated conditioning frames at segment boundaries and trim terminal padding to obtain 357 or 961 frames at 16 FPS. API models are run as dependent requests: the current segment must finish before its generated endpoint can initialize the next. We retain each segment's prompt and output alongside the assembled video.

\adaptermodel{adaptertext}{bytedance}{Seedance2.5}~\citep{bytedance2026seedance25} is evaluated through its first-frame image-to-video API at 720p with audio disabled. Short cases use a 15-second request; long cases use 2 consecutive 20-second requests at the native 24-FPS rate. The long prompts cover requested frames 0--480 and 480--960, respectively, each expressed as a local 0--20-second script. The second request uses the first segment's generated last frame. Short output is trimmed to 357 frames; long segments are joined after removing the repeated boundary frame to obtain 961 frames.

\adaptermodel{adaptertext}{kuaishou}{Kling3.0}~\citep{kling2026video3} uses the image-to-video API at 720p and native 24 FPS, with audio and multi-shot generation disabled. Short cases request 15 seconds. Long cases request 15, 15, and 10 seconds, with a separate prompt for each successive camera window. The first request uses $I_0$, and each continuation uses the preceding generated last frame. We assemble the outputs by removing shared boundary frames and trimming the tail.

\begin{figure}[t]
\centering
\includegraphics[width=\linewidth]{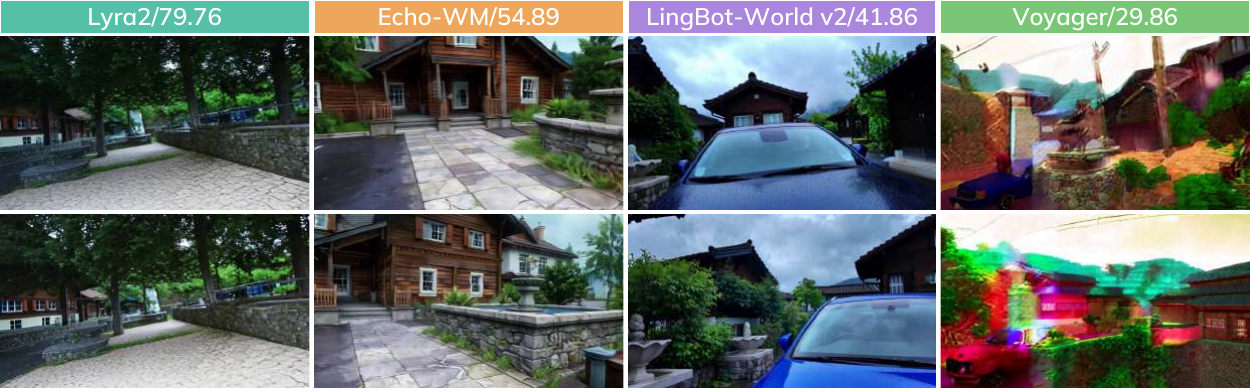}
\caption{\textbf{Generated-content persistence on a Sekai yaw-return example.} Each column compares a model's first visit (top) and revisit (bottom) at registered matching requested poses. Displayed scores are for this case.}
\label{fig:tier4-example2}
\end{figure}

\adaptermodel{adaptertext}{nvidia}{Cosmos3-Super}~\citep{nvidia2026cosmos3} uses the 4-Step Image2Video checkpoint with guidance scale 1 and $832\times480$ output. The scene caption is combined with the motion text for each generation window. Short cases generate 357 frames in one call. Long cases use windows of 357, 357, and 249 frames, covering requested indices 0--356, 356--712, and 712--960. The 2 repeated boundary frames are removed to obtain 961 frames. Prompt times follow the native 24-FPS clock, and the assembled frames are published at 16 FPS.

\adaptermodel{adaptertext}{tencent}{HunyuanVideo1.5}~\citep{tencent2025hy15} uses the 480p image-to-video step-distilled checkpoint, 12 steps, and $848\times480$ output. Each window generates 121 frames at native 24 FPS, with the preceding generated last frame conditioning the next window. Thus, 3 windows produce 361 frames for short cases, cropped to 357; 8 windows produce exactly 961 frames for long cases. Each window receives a scene description based on its actual conditioning image, followed by the window-specific camera script. The camera script preserves the timing, turning angles, and motion reversals specified by the requested trajectory within each generation window.

\adaptermodel{adaptertext}{minimax}{MiniMax-H3}~\citep{minimax2026h3} uses the Base FL2VA variant with 50 sampling time points, corresponding to 49 denoising iterations, at 768p. It receives the conditioning image and the scene caption combined with its window-specific motion script. Short requests produce 362 frames, cropped to 357. Long generation uses 362, 362, and 243 frames, with each later segment initialized from the preceding generated last frame. Removing the 2 repeated boundary frames leaves 965 frames, from which the first 961 are retained. The motion prompts are constructed for these specific windows at native 24 FPS before output conversion to 16 FPS.

\begin{figure}[t]
\centering
\includegraphics[width=\linewidth]{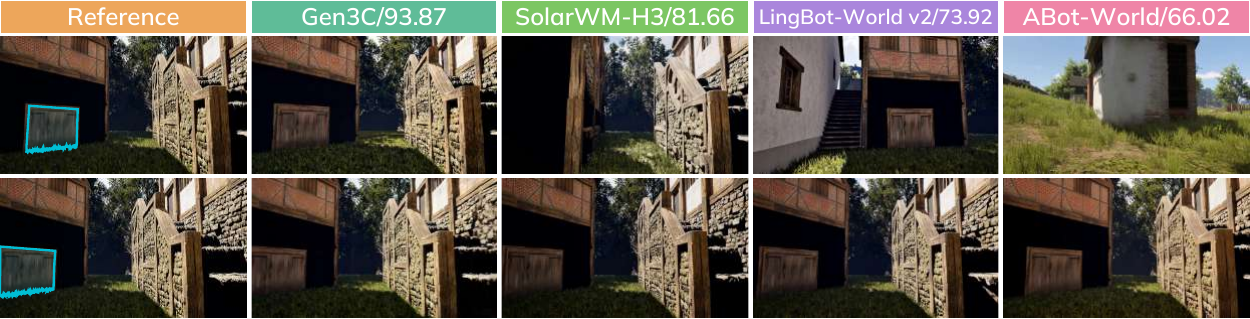}
\caption{\textbf{Input preservation on a UE yaw-return example.} The left column shows the input and registered query references. The other columns show model returns (top) and queries (bottom). Ideally, Return reproduces the input, while Query preserves the reference anchor's position and appearance. Cyan contours mark the wooden-crate anchor in the input and its input-visible query region. Displayed scores are for this case.}
\label{fig:tier3-example1}
\end{figure}

\begin{figure}[t]
\centering
\includegraphics[width=\linewidth]{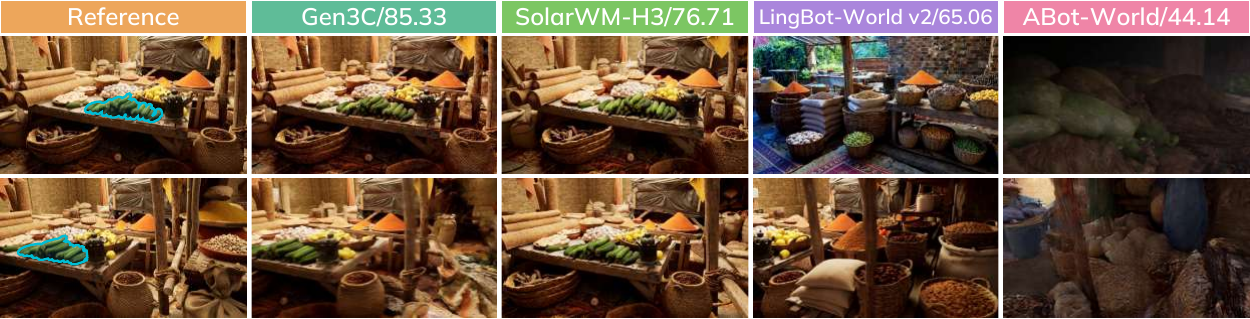}
\caption{\textbf{Input preservation on a UE lateral-return example.} The left column shows the input and registered query references. The other columns show model returns (top) and queries (bottom). Ideally, Return reproduces the input, while Query preserves the reference anchor's position and appearance. Cyan contours mark the input anchor and its input-visible query region. Displayed scores are for this case.}
\label{fig:tier3-example2}
\end{figure}

\adaptermodel{adaptertext}{lightricks}{LTX2.5}~\citep{lightricks2026ltx25} uses the 22B distilled checkpoint and its 2-stage $8+3$-step schedule. The prompt describes the scene and the complete requested trajectory. Both short and long sequences are generated in a single call from $I_0$: the short request is padded to 361 frames and trimmed to 357, while the long request produces 961 frames. Images are padded from $1280\times720$ to $1280\times768$ for inference and cropped back afterward. Prompt timings use the native 24-FPS clock, and output frames are published at 16 FPS.

\adaptermodel{adaptertext}{alibaba}{Wan2.7}~\citep{wan2025wan} uses the image-to-video API at 720p and native 30 FPS. Short cases request 12 seconds; long cases request 15, 15, and 3 seconds. Each request receives the relevant portion of the motion script with local times expressed at 30 FPS. Continuation requests use the generated last frame, and the assembled sequence removes repeated boundary frames before terminal trimming to 357 or 961 frames. The retained frame sequence is played at 16 FPS.

\adaptermodel{adaptertext}{meituan}{LongCat}~\citep{meituan2025longcat} uses the base model with its step-distillation LoRA, 16 steps, guidance scale 1, and $832\times480$ output. Each segment contains 93 frames. The first is initialized from $I_0$; subsequent segments use 13 generated context frames and add 80 new frames through the native video-continuation interface. Consequently, 5 segments yield 413 frames for short cases and 12 yield 973 for long cases, followed by terminal trimming to 357 and 961. The scene caption is preserved and combined with window-specific English motion text. Window starts advance by 80 frames, and local times use the native 15-FPS clock. The motion script continues from the retained context and describes only the new portion of the trajectory. The model carries its native continuation cache from one segment to the next throughout the same evaluation case, for short and long videos.

\flushbottom

\begin{extendedresults}

\section{Extended Results and Analysis}
\label{app:extended}

We examine the capabilities behind the overall scores through complete component tables and qualitative examples. Cross-tier relationships and analyses across trajectories, horizons, and data sources appear in Sections~\ref{sec:analysis-cross-tier}--\ref{sec:analysis-source}. 

\subsection{Detailed Results by Evaluation Tier}
\label{app:component-results}

\subsubsection{Tier 1: Control Alignment}

\begin{table}[t]
\centering
\begin{tablebody}{2pt}{1.2}

\caption{\textbf{Tier 5: Detailed 3D self-consistency results.} Scores use a 0--100 scale ($\uparrow$). Models are ordered by Tier Score within each group. \ranklegend{orbiaRankFirst}{1st}, \ranklegend{orbiaRankSecond}{2nd}, \ranklegend{orbiaRankThird}{3rd} mark the top three scores per column and group.}
\label{tab:tier5-components}
\fontsize{7.5}{9}\selectfont
\begin{tabularx}{\linewidth}{@{}lc*{9}{>{\centering\arraybackslash}X}@{}}
\toprule
\rowcolor{orbiaTableHeader}
 & \textbf{3D} & \multicolumn{2}{c}{\textbf{Depth}} & \multicolumn{2}{c}{\textbf{Reprojection}} & \textbf{Regional} & \multicolumn{3}{c}{\textbf{GS reconstruction}} &\textbf{Tier} \\
\cmidrule(lr){3-4}\cmidrule(lr){5-6}\cmidrule(lr){8-10}
\rowcolor{orbiaTableHeader}
\textbf{Model} & \textbf{Repr.} & \textbf{Short} & \textbf{Medium} & \textbf{Short} & \textbf{Medium} & \textbf{scale} & \textbf{PSNR} & \textbf{SSIM} & \textbf{LPIPS} & \textbf{Score}\\
\midrule
\rowcolor{orbiaTableHeader}
\multicolumn{11}{l}{\textbf{World Models}}\\
\modellogo{tencent}HY-WorldPlay & $\times$ & \scorefirst{$\mathbf{79.83}$} & \scorefirst{$\mathbf{78.33}$} & \scorethird{$\mathbf{78.19}$} & \scorethird{$\mathbf{68.54}$} & \scorefirst{$\mathbf{84.41}$} & $63.44$ & $40.03$ & $30.00$ & \scorefirst{$\mathbf{70.34}$}\\
\rowcolor{orbiaTableStripe}
\modellogo{nvidia}Cosmos3-Nano & $\times$ & $75.79$ & \scorethird{$\mathbf{73.59}$} & \scorefirst{$\mathbf{81.96}$} & \scorefirst{$\mathbf{73.17}$} & $79.05$ & $75.89$ & $41.88$ & $29.52$ & \scoresecond{$\mathbf{70.10}$}\\
\modellogo{cuhksz}SolarWM-H3 & $\times$ & \scoresecond{$\mathbf{77.99}$} & \scoresecond{$\mathbf{76.28}$} & $75.09$ & $62.06$ & \scoresecond{$\mathbf{83.82}$} & $64.79$ & $42.79$ & $29.89$ & \scorethird{$\mathbf{68.84}$}\\
\rowcolor{orbiaTableStripe}
\modellogo{alaya}EVOKE & $\checkmark$ & \scorethird{$\mathbf{75.88}$} & $72.93$ & $75.21$ & $66.40$ & \scorethird{$\mathbf{80.33}$} & $64.55$ & \scorethird{$\mathbf{44.38}$} & $30.70$ & $68.02$\\
\modellogo{jd}Echo-WM & $\times$ & $72.47$ & $70.36$ & $73.15$ & $59.97$ & $78.92$ & $64.80$ & $40.57$ & $27.01$ & $65.26$\\
\rowcolor{orbiaTableStripe}
\modellogo{amap}FantasyWorld & $\checkmark$ & $71.51$ & $67.24$ & \scoresecond{$\mathbf{78.78}$} & \scoresecond{$\mathbf{69.42}$} & $72.42$ & $50.22$ & $39.13$ & $25.25$ & $63.53$\\
\modellogo{nvidia}Lyra2 & $\checkmark$ & $66.65$ & $61.92$ & $71.02$ & $56.12$ & $68.81$ & $70.86$ & \scoresecond{$\mathbf{45.14}$} & \scoresecond{$\mathbf{32.10}$} & $61.51$\\
\rowcolor{orbiaTableStripe}
\modellogo{deepmind}Genie3 & -- & $66.75$ & $63.53$ & $67.53$ & $49.72$ & $70.05$ & $80.60$ & $43.12$ & $29.89$ & $61.25$\\
\modellogo{nvidia}Gen3C & $\checkmark$ & $68.23$ & $60.72$ & $72.33$ & $58.19$ & $65.98$ & $66.47$ & $40.98$ & $27.36$ & $60.16$\\
\rowcolor{orbiaTableStripe}
\modellogo{tsinghua}Astra & $\times$ & $61.70$ & $56.07$ & $71.87$ & $57.95$ & $59.08$ & \scoresecond{$\mathbf{88.38}$} & \scorefirst{$\mathbf{47.54}$} & \scorefirst{$\mathbf{34.36}$} & $59.91$\\
\modellogo{alibaba}HappyOyster & -- & $60.16$ & $55.32$ & $71.49$ & $55.32$ & $63.24$ & \scorethird{$\mathbf{83.19}$} & $42.01$ & \scorethird{$\mathbf{31.73}$} & $59.17$\\
\rowcolor{orbiaTableStripe}
\modellogo{robbyant}LingBot-World v1 & $\times$ & $64.85$ & $60.56$ & $71.63$ & $56.02$ & $69.40$ & $52.06$ & $41.66$ & $25.71$ & $58.93$\\
\modellogo{alaya}AlayaWorld & $\checkmark$ & $65.25$ & $60.96$ & $73.08$ & $59.87$ & $67.55$ & $54.46$ & $36.96$ & $24.25$ & $58.92$\\
\rowcolor{orbiaTableStripe}
\modellogo{robbyant}LingBot-World v2 & $\times$ & $62.13$ & $59.45$ & $71.07$ & $56.73$ & $68.00$ & $53.32$ & $33.28$ & $20.66$ & $57.11$\\
\modellogo{skywork}Matrix game 3.5 & $\checkmark$ & $55.41$ & $53.19$ & $70.35$ & $56.77$ & $60.45$ & $65.77$ & $31.52$ & $23.05$ & $54.61$\\
\rowcolor{orbiaTableStripe}
\modellogo{nvidia}SANA-WM & $\times$ & $52.45$ & $45.98$ & $70.75$ & $57.50$ & $59.96$ & $72.98$ & $34.47$ & $21.87$ & $54.10$\\
\modellogo{amap}DreamX-World-5B & $\times$ & $43.80$ & $41.53$ & $70.84$ & $58.46$ & $51.84$ & $60.42$ & $27.75$ & $19.07$ & $48.73$\\
\rowcolor{orbiaTableStripe}
\modellogo{amap}ABot-World & $\times$ & $38.91$ & $39.34$ & $67.12$ & $53.75$ & $42.44$ & \scorefirst{$\mathbf{88.46}$} & $38.03$ & $27.33$ & $48.32$\\
\modellogo{tencent}Voyager & $\checkmark$ & $47.61$ & $44.22$ & $67.26$ & $53.08$ & $50.26$ & $32.86$ & $37.63$ & $16.27$ & $46.32$\\
\midrule
\rowcolor{orbiaTableGroupHeader}
\multicolumn{11}{l}{\textbf{TI2V Models}}\\
\modellogo{nvidia}Cosmos3-Super & -- & \scorefirst{$\mathbf{84.31}$} & \scorefirst{$\mathbf{83.62}$} & \scorefirst{$\mathbf{80.87}$} & \scorefirst{$\mathbf{72.45}$} & \scorefirst{$\mathbf{90.61}$} & \scorethird{$\mathbf{61.72}$} & \scorefirst{$\mathbf{46.26}$} & \scoresecond{$\mathbf{32.85}$} & \scorefirst{$\mathbf{74.55}$}\\
\rowcolor{orbiaTableGroupStripe}
\modellogo{bytedance}Seedance2.5 & -- & \scorethird{$\mathbf{82.91}$} & \scorethird{$\mathbf{81.78}$} & $77.31$ & \scorethird{$\mathbf{66.06}$} & \scorethird{$\mathbf{88.92}$} & \scoresecond{$\mathbf{63.94}$} & \scorethird{$\mathbf{45.10}$} & \scorethird{$\mathbf{32.60}$} & \scoresecond{$\mathbf{72.54}$}\\
\modellogo{lightricks}LTX2.5 & -- & $80.81$ & $79.33$ & $75.69$ & $63.74$ & $87.27$ & \scorefirst{$\mathbf{69.66}$} & \scoresecond{$\mathbf{46.01}$} & \scorefirst{$\mathbf{32.86}$} & \scorethird{$\mathbf{71.64}$}\\
\rowcolor{orbiaTableGroupStripe}
\modellogo{tencent}HunyuanVideo1.5 & $\times$ & \scoresecond{$\mathbf{83.92}$} & \scoresecond{$\mathbf{82.65}$} & \scoresecond{$\mathbf{78.37}$} & \scoresecond{$\mathbf{70.65}$} & \scoresecond{$\mathbf{89.25}$} & $48.35$ & $38.54$ & $27.68$ & $71.31$\\
\modellogo{kuaishou}Kling3.0 & -- & $81.55$ & $80.15$ & \scorethird{$\mathbf{77.79}$} & $65.80$ & $87.78$ & $61.44$ & $40.97$ & $31.89$ & $71.30$\\
\rowcolor{orbiaTableGroupStripe}
\modellogo{minimax}MiniMax-H3 & -- & $80.29$ & $78.47$ & $74.64$ & $61.82$ & $87.09$ & $53.05$ & $43.59$ & $32.31$ & $69.42$\\
\modellogo{alibaba}Wan2.7 & -- & $72.65$ & $69.57$ & $75.24$ & $62.71$ & $77.70$ & $49.85$ & $39.28$ & $27.80$ & $64.19$\\
\rowcolor{orbiaTableGroupStripe}
\modellogo{meituan}LongCat & $\times$ & $72.77$ & $69.39$ & $71.82$ & $59.83$ & $79.07$ & $56.88$ & $38.62$ & $25.83$ & $64.11$\\
\bottomrule
\end{tabularx}
\end{tablebody}
\end{table}

\label{app:control-results}

\metriccard{metriccontrol}{Direction following and path accuracy distinguish different control skills.}
Lyra2 leads translation and Move Accuracy, while SolarWM-H3 leads rotation and Look Accuracy among world models (Table~\ref{tab:control-breakdown}). LingBot-World v2 illustrates the gap between the two types of measurement: its Move Accuracy is $79.45$, but its translation score is $50.94$. \textbf{Moving in the right direction} does not ensure the correct \emph{distance, timing, or camera trajectory}. The strongest TI2V model, Seedance2.5, also remains well below the leading world models across control components.

\metriccard{metriccontrol}{Turning is a major source of control differences.}
\textbf{Look Accuracy} scores vary the most across camera-conditioned models, and all three action-conditioned models score lower on Look than Move. HunyuanVideo1.5 leads TI2V Move Accuracy but ranks last on Look, showing how differently the two commands can be executed. This distinction matters for yaw sweeps, curved paths, and orbits, which repeatedly change the viewing direction. In several inspected Genie3 cases, lateral requests produce turns instead of translation. Separately, its lower Look Accuracy indicates difficulty executing requested turns; the lateral-motion errors instead reflect confusion between requested translation and turning.

For ABot-World, Genie3, and HappyOyster, Tier Score uses Move and Look Accuracy. Their parenthesized translation and rotation scores provide additional context. Figures~\ref{fig:tier1-example1}--\ref{fig:tier1-example2} compare the requested camera trajectories with those recovered by DA3 from the generated videos for the same cases.

\begin{table}[t]
\centering
\begin{tablebody}{2pt}{1.2}

\caption{\textbf{Tier 6: Detailed temporal stability results.} Short- and long-video means receive weights of 30\% and 70\%, respectively. Scores use a 0--100 scale ($\uparrow$). Models are ordered by Tier Score within each group. \ranklegend{orbiaRankFirst}{1st}, \ranklegend{orbiaRankSecond}{2nd}, \ranklegend{orbiaRankThird}{3rd} mark the top three scores per column and group.}
\label{tab:tier6-components}
\fontsize{7.5}{9}\selectfont
\begin{tabularx}{\linewidth}{@{}lc*{7}{>{\centering\arraybackslash}X}@{}}
\toprule
\rowcolor{orbiaTableHeader}
 & \textbf{3D} & \multicolumn{2}{c}{\textbf{Visual quality}} & \multicolumn{2}{c}{\textbf{Control}} & \multicolumn{2}{c}{\textbf{Geometry}} &\textbf{Tier} \\
\cmidrule(lr){3-4}\cmidrule(lr){5-6}\cmidrule(lr){7-8}
\rowcolor{orbiaTableHeader}
\textbf{Model} & \textbf{Repr.} & \componenthead{Stability\\Score} & \componenthead{Mean\\Performance} & \componenthead{Stability\\Score} & \componenthead{Mean\\Performance} & \componenthead{Stability\\Score} & \componenthead{Mean\\Performance} & \textbf{Score}\\
\midrule
\rowcolor{orbiaTableHeader}
\multicolumn{9}{l}{\textbf{World Models}}\\
\modellogo{cuhksz}SolarWM-H3 & $\times$ & $83.16$ & $60.29$ & \scorethird{$\mathbf{80.01}$} & \scoresecond{$\mathbf{76.78}$} & \scorefirst{$\mathbf{88.96}$} & \scorethird{$\mathbf{75.85}$} & \scorefirst{$\mathbf{84.05}$}\\
\rowcolor{orbiaTableStripe}
\modellogo{alaya}EVOKE & $\checkmark$ & \scorefirst{$\mathbf{87.61}$} & \scorefirst{$\mathbf{64.01}$} & $74.09$ & $66.73$ & $87.83$ & $73.33$ & \scoresecond{$\mathbf{83.18}$}\\
\modellogo{deepmind}Genie3 & -- & $84.70$ & $61.43$ & \scorefirst{$\mathbf{82.72}$} & $59.83$ & $82.04$ & $64.91$ & \scorethird{$\mathbf{83.15}$}\\
\rowcolor{orbiaTableStripe}
\modellogo{jd}Echo-WM & $\times$ & \scoresecond{$\mathbf{85.64}$} & \scorethird{$\mathbf{61.61}$} & $74.40$ & $68.87$ & \scorethird{$\mathbf{88.02}$} & $72.64$ & $82.69$\\
\modellogo{nvidia}Lyra2 & $\checkmark$ & $81.12$ & $54.94$ & \scoresecond{$\mathbf{81.97}$} & \scorefirst{$\mathbf{78.47}$} & $83.83$ & $64.49$ & $82.31$\\
\rowcolor{orbiaTableStripe}
\modellogo{nvidia}Gen3C & $\checkmark$ & $76.12$ & $46.67$ & $75.17$ & \scorethird{$\mathbf{70.61}$} & $85.58$ & $66.40$ & $78.96$\\
\modellogo{alibaba}HappyOyster & -- & $76.13$ & $49.28$ & $79.66$ & $59.98$ & $78.97$ & $63.57$ & $78.25$\\
\rowcolor{orbiaTableStripe}
\modellogo{tencent}Voyager & $\checkmark$ & $77.53$ & $49.49$ & $74.48$ & $65.08$ & $82.12$ & $56.86$ & $78.04$\\
\modellogo{tencent}HY-WorldPlay & $\times$ & $77.23$ & $52.96$ & $66.05$ & $46.37$ & \scoresecond{$\mathbf{88.70}$} & \scorefirst{$\mathbf{77.46}$} & $77.32$\\
\rowcolor{orbiaTableStripe}
\modellogo{robbyant}LingBot-World v2 & $\times$ & \scorethird{$\mathbf{84.74}$} & \scoresecond{$\mathbf{62.71}$} & $65.08$ & $58.41$ & $82.02$ & $66.08$ & $77.28$\\
\modellogo{skywork}Matrix game 3.5 & $\checkmark$ & $82.35$ & $59.11$ & $69.21$ & $62.75$ & $77.68$ & $60.96$ & $76.41$\\
\rowcolor{orbiaTableStripe}
\modellogo{amap}ABot-World & $\times$ & $73.87$ & $46.83$ & $78.08$ & $53.59$ & $77.01$ & $52.80$ & $76.32$\\
\modellogo{alaya}AlayaWorld & $\checkmark$ & $81.48$ & $54.81$ & $62.86$ & $54.32$ & $84.59$ & $67.16$ & $76.31$\\
\rowcolor{orbiaTableStripe}
\modellogo{robbyant}LingBot-World v1 & $\times$ & $81.74$ & $58.20$ & $61.36$ & $50.08$ & $83.01$ & $64.77$ & $75.37$\\
\modellogo{nvidia}SANA-WM & $\times$ & $83.19$ & $57.21$ & $64.72$ & $58.19$ & $77.19$ & $56.98$ & $75.03$\\
\rowcolor{orbiaTableStripe}
\modellogo{amap}DreamX-World-5B & $\times$ & $83.21$ & $60.44$ & $67.91$ & $51.56$ & $73.34$ & $54.14$ & $74.82$\\
\modellogo{nvidia}Cosmos3-Nano & $\times$ & $78.45$ & $55.77$ & $59.33$ & $34.93$ & $86.36$ & \scoresecond{$\mathbf{76.22}$} & $74.71$\\
\rowcolor{orbiaTableStripe}
\modellogo{amap}FantasyWorld & $\checkmark$ & $74.85$ & $50.26$ & $56.42$ & $40.73$ & $86.26$ & $72.02$ & $72.51$\\
\modellogo{tsinghua}Astra & $\times$ & $64.70$ & $35.58$ & $53.97$ & $33.04$ & $83.26$ & $64.17$ & $67.31$\\
\midrule
\rowcolor{orbiaTableGroupHeader}
\multicolumn{9}{l}{\textbf{TI2V Models}}\\
\modellogo{bytedance}Seedance2.5 & -- & $86.92$ & $63.81$ & \scorefirst{$\mathbf{62.88}$} & \scorefirst{$\mathbf{43.55}$} & \scorethird{$\mathbf{91.38}$} & \scorethird{$\mathbf{80.15}$} & \scorefirst{$\mathbf{80.39}$}\\
\rowcolor{orbiaTableGroupStripe}
\modellogo{nvidia}Cosmos3-Super & -- & \scorefirst{$\mathbf{89.24}$} & \scorefirst{$\mathbf{67.43}$} & $57.27$ & $33.42$ & \scorefirst{$\mathbf{92.69}$} & \scorefirst{$\mathbf{82.80}$} & \scoresecond{$\mathbf{79.73}$}\\
\modellogo{kuaishou}Kling3.0 & -- & $87.31$ & $65.18$ & \scoresecond{$\mathbf{61.29}$} & \scoresecond{$\mathbf{38.15}$} & $89.31$ & $79.52$ & \scorethird{$\mathbf{79.30}$}\\
\rowcolor{orbiaTableGroupStripe}
\modellogo{tencent}HunyuanVideo1.5 & $\times$ & $85.37$ & $63.09$ & \scorethird{$\mathbf{59.88}$} & \scorethird{$\mathbf{37.05}$} & \scoresecond{$\mathbf{91.62}$} & \scoresecond{$\mathbf{81.08}$} & $78.96$\\
\modellogo{minimax}MiniMax-H3 & -- & \scoresecond{$\mathbf{88.35}$} & \scoresecond{$\mathbf{66.75}$} & $55.62$ & $34.32$ & $88.56$ & $76.51$ & $77.51$\\
\rowcolor{orbiaTableGroupStripe}
\modellogo{lightricks}LTX2.5 & -- & \scorethird{$\mathbf{88.03}$} & \scorethird{$\mathbf{66.00}$} & $54.52$ & $31.90$ & $89.35$ & $78.63$ & $77.30$\\
\modellogo{alibaba}Wan2.7 & -- & $86.62$ & $64.66$ & $56.90$ & $35.02$ & $83.85$ & $72.37$ & $75.79$\\
\rowcolor{orbiaTableGroupStripe}
\modellogo{meituan}LongCat & $\times$ & $86.03$ & $64.93$ & $50.29$ & $27.88$ & $86.33$ & $70.89$ & $74.22$\\
\bottomrule
\end{tabularx}
\end{tablebody}
\end{table}

\subsubsection{Tier 2: Visual Quality}
\label{app:quality-results}

\metriccard{metricquality}{TI2V models achieve more uniformly high visual quality.}
\textbf{Seven of the eight TI2V models} exceed the best world-model quality aggregate (Table~\ref{tab:quality-breakdown}). Their advantage spans aesthetic, imaging, and human-preference scores, with a smaller spread than among world models. Cosmos3-Super leads the TI2V group, while EVOKE leads world models. Figures~\ref{fig:tier2-example1}--\ref{fig:tier2-example3} show the corresponding differences in detail, clarity, and scene appearance.

\metriccard{metricquality}{Perceptual quality shows larger differences between models than temporal smoothness.}
\textbf{HPS} scores vary the most across models in both groups, and world models also differ in imaging quality. Motion-smoothness scores range only from $97.05$ to $99.40$, while flicker scores are also high and similar across models. These temporal measures reveal broadly smooth outputs; perceptual quality and preference account for more of the remaining differences between models. Cosmos3-Nano leads world models in flicker and smoothness, whereas EVOKE's overall quality advantage comes from its perceptual scores.

\subsubsection{Tier 3: Input Preservation}
\label{app:tier3-results}

\metriccard{metricinput}{Recovering the initial scene remains harder than retaining recognizable anchors.}
\textbf{Return} scores vary the most across models in both groups, while Identity scores are generally high and more similar across models (Table~\ref{tab:tier3-components}). Identity is evaluated during the input-to-query interval, when the initial observation is more recent. Return requires recovery of the starting scene after the full exploration. The contrast points to a longer-term memory challenge: recognizable objects can survive early exploration even when the original view is not faithfully recovered after completing the full exploration trajectory and returning to the initial viewpoint.

\metriccard{metricinput}{Stored scene evidence can support recovery after exploration.}
Gen3C leads world-model Return and overall input preservation despite ranking 17th in visual quality. Its \emph{persistent 3D cache} supplies target-view renderings during generation~\citep{ren2025gen3c}, keeping earlier scene evidence accessible when the camera returns. This cache may help it recover the original view after exploration. SolarWM-H3 instead leads Query, and Cosmos3-Nano leads Identity, showing that different models lead recovery of the starting view, fidelity at the requested query viewpoint, and recognition of the original anchors.

TI2V models have higher mean Identity and Return scores but lower mean Query fidelity than world models. Seedance2.5 leads both checkpoints within its group; other models retain recognizable content with weaker viewpoint-specific fidelity. Figures~\ref{fig:tier3-example1} and~\ref{fig:tier3-example2} compare these checkpoints with their registered references.

\subsubsection{Tier 4: Generated-Content Persistence}
\label{app:tier4-results}

\metriccard{metricrevisit}{Recognizable revisits can still lose detailed appearance.}
Different-view global DINO similarity remains relatively high for many models, while same-view perceptual fidelity exposes larger failures (Table~\ref{tab:tier4-components}). MiniMax-H3, for example, scores $77.27$ on global DINO but $17.32$ on normalized same-view LPIPS. A scene can retain its \emph{semantic character} while its \textbf{textures, objects, or layout change}. The matched views in Figures~\ref{fig:main-preservation} and~\ref{fig:tier4-example2} make these failures visible.

\begin{figure}[t]
\centering
\includegraphics[width=\linewidth]{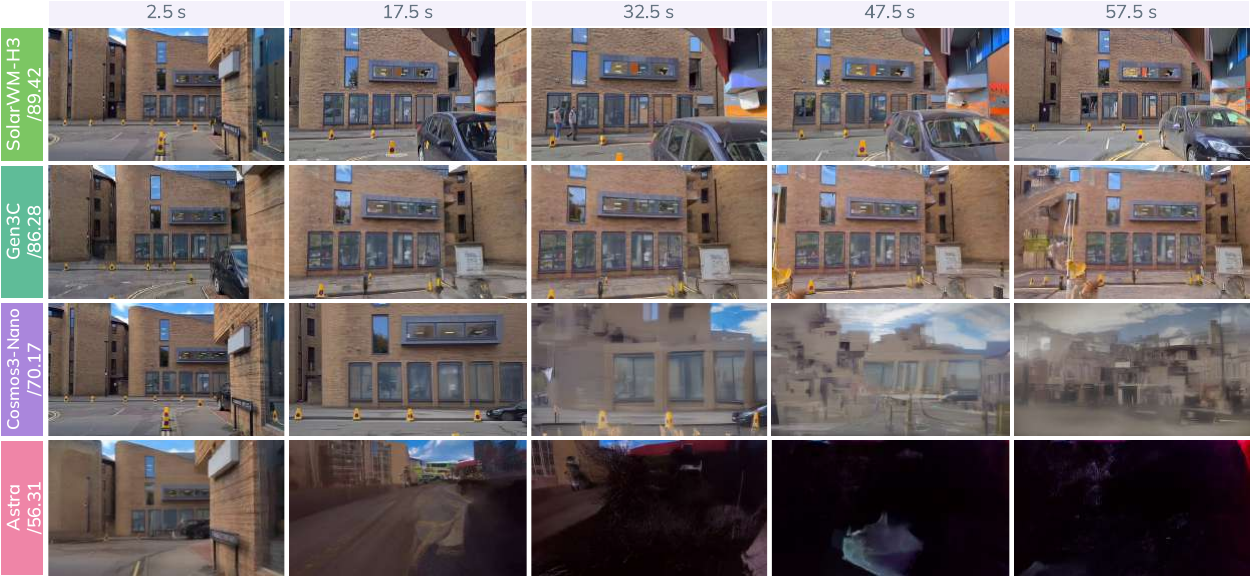}
\caption{\textbf{Temporal stability on a Sekai forward--backward example.} Rows show models and columns show five timestamps from the minute-long sequence. Displayed scores are for this case.}
\label{fig:tier6-example1}
\end{figure}

\begin{figure}[t]
\centering
\includegraphics[width=\linewidth]{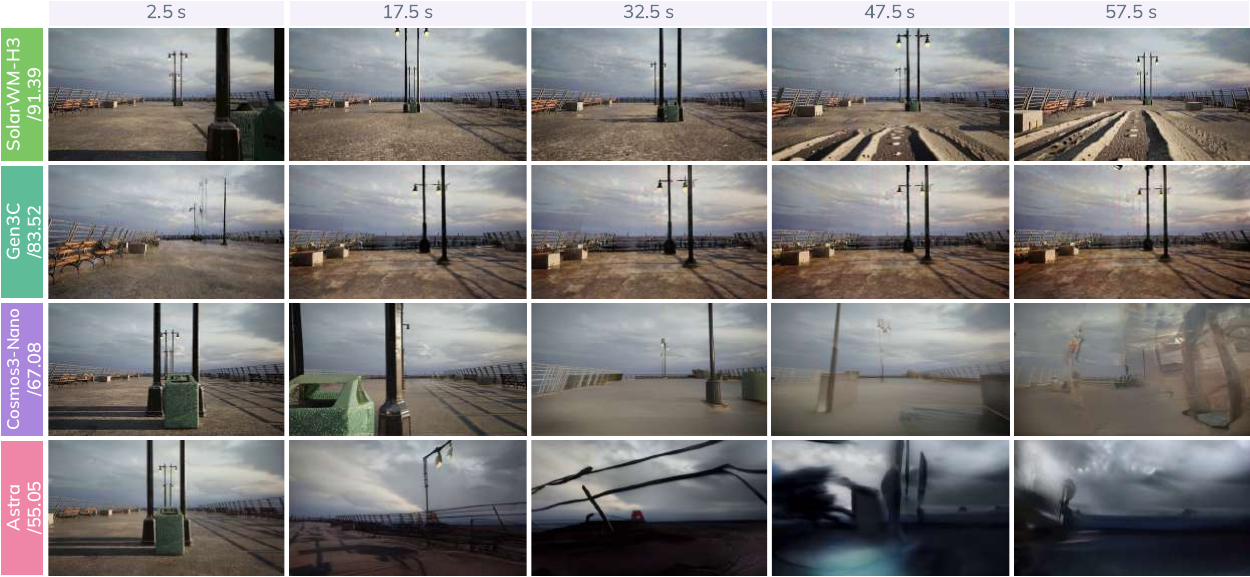}
\caption{\textbf{Temporal stability on a UE arc-return example.} Rows show models and columns show five timestamps from the minute-long sequence. Displayed scores are for this case.}
\label{fig:tier6-example2}
\end{figure}

\begin{figure}[t]
\centering
\includegraphics[width=\linewidth]{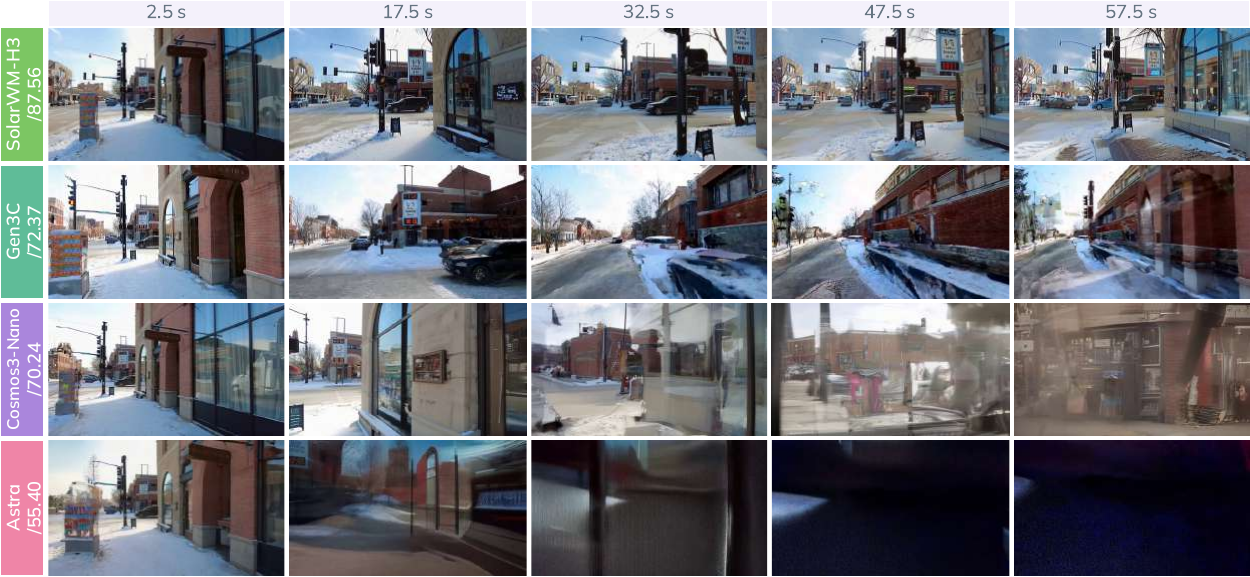}
\caption{\textbf{Temporal stability on a SpatialVID arc-return example.} Rows show models and columns show five timestamps from the minute-long sequence. Displayed scores are for this case.}
\label{fig:tier6-example3}
\end{figure}

\metriccard{metricrevisit}{Appearance preservation and spatial agreement favor different mechanisms.}
Lyra2 leads world-model same-view LPIPS, patch DINO, and reprojected DINO. Its stored geometry and dense correspondences retrieve historical frames~\citep{shen2026lyra2}, which may help preserve appearance across revisits. HY-WorldPlay leads reprojected depth consistency, while SolarWM-H3 leads surface F-score. Depth compares distances on overlapping surfaces; F-score measures agreement between recovered point sets. These distinct strengths bring SolarWM-H3 and Lyra2 to \textbf{nearly identical tier scores}.

Cosmos3-Super leads TI2V persistence through strong appearance and surface agreement, while Seedance2.5 leads reprojected depth. The best models in the two groups are close, but their within-group orderings differ from visual quality: MiniMax-H3 is second in TI2V quality and sixth in persistence.

\subsubsection{Tier 5: 3D Self-Consistency}
\label{app:tier5-results}

\metriccard{metricgeometry}{Wider temporal separation exposes inconsistencies hidden between nearby views.}
\textbf{Every model} has lower feature consistency at \emph{medium} than at \emph{short} temporal offsets, with greater between-model variation at the medium interval (Table~\ref{tab:tier5-components}). Depth and regional scale also separate world models more strongly than short-range features. Together, these results locate the harder part of 3D consistency: maintaining corresponding content and spatial relationships as the camera moves farther through the scene.

\metriccard{metricgeometry}{High reconstruction scores can coexist with weak depth and scale agreement.}
HY-WorldPlay leads depth and regional scale. Its Reconstituted Context Memory retrieves geometrically relevant historical frames, and Context Forcing preserves memory use during distillation~\citep{sun2025worldplay}, consistent with its local geometric strength. Astra achieves strong reconstruction scores but weaker depth and regional-scale agreement. Reconstruction measures how well a shared 3D representation reproduces the generated views; low-detail outputs may make this task easier without yielding equally strong geometric agreement. ABot-World also combines high reconstruction PSNR with low depth and scale consistency.

Five TI2V models exceed the best world-model self-consistency aggregate, with their largest group advantages in depth and regional scale. Their weaker control results indicate that internally compatible generated views can still follow a different path from the requested cameras.

\subsubsection{Tier 6: Temporal Stability}
\label{app:tier6-results}

\metriccard{metriclong}{Models differ most in temporal control.}
\textbf{Temporal control} scores vary more across models than quality or geometry scores, in both average performance and stability (Table~\ref{tab:tier6-components}). TI2V models have higher group means in quality and local geometry but weaker control, bringing their aggregate Tier~6 performance close to that of world models. Seedance2.5 leads its group overall through stronger control, while Cosmos3-Super leads quality and geometry.

\metriccard{metriclong}{Different models lead in average performance and temporal stability.}
Genie3 leads the \emph{temporal control score}, while Lyra2 has the highest \emph{mean control level}. EVOKE leads world-model temporal quality; SolarWM-H3 leads geometric stability and the overall tier, while HY-WorldPlay has the highest mean geometry level. SolarWM's camera-conditioned autoregressive training, distribution-matching distillation~\citep{huang2026solarwm}, and retained rollout history offer a possible explanation for its sustained performance. Figures~\ref{fig:tier6-example1}--\ref{fig:tier6-example3} show how the generated views evolve; Section~\ref{sec:analysis-temporal} examines the corresponding temporal performance.

\subsection{Additional Annotated Examples and Review Interface}
\label{app:additional-annotations}

These examples extend the four source-group figures in Appendix~\ref{app:annotated-examples} with additional exploration templates. The final figure shows the manual review interface for a Sekai orbit case.

\begin{figure}[tp]
\centering
\includegraphics[width=0.92\linewidth]{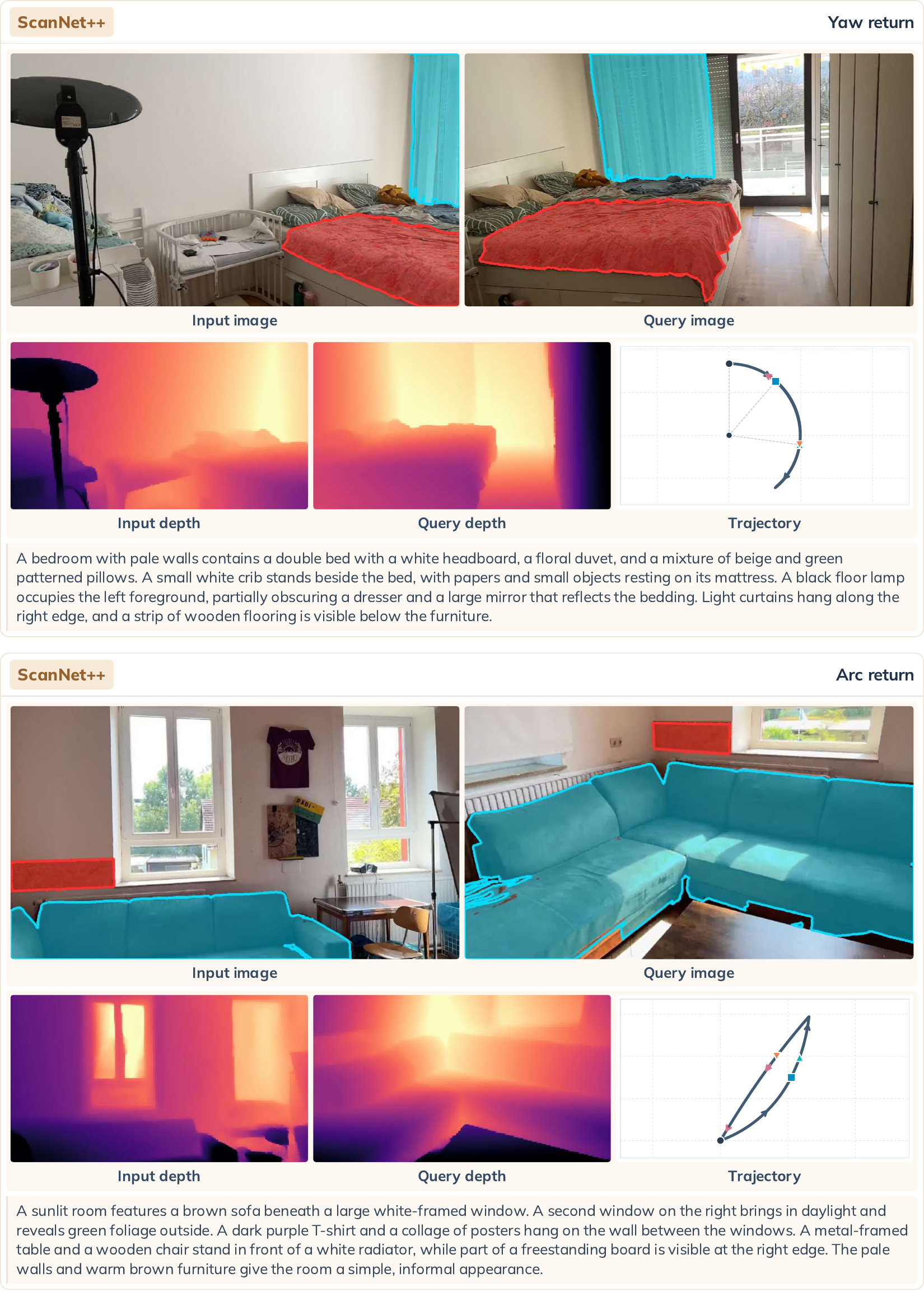}
\caption{Examples of the yaw return (top) and arc return (bottom) templates from ScanNet++.}
\label{fig:annotated-02}
\end{figure}

\begin{figure}[tp]
\centering
\includegraphics[width=0.92\linewidth]{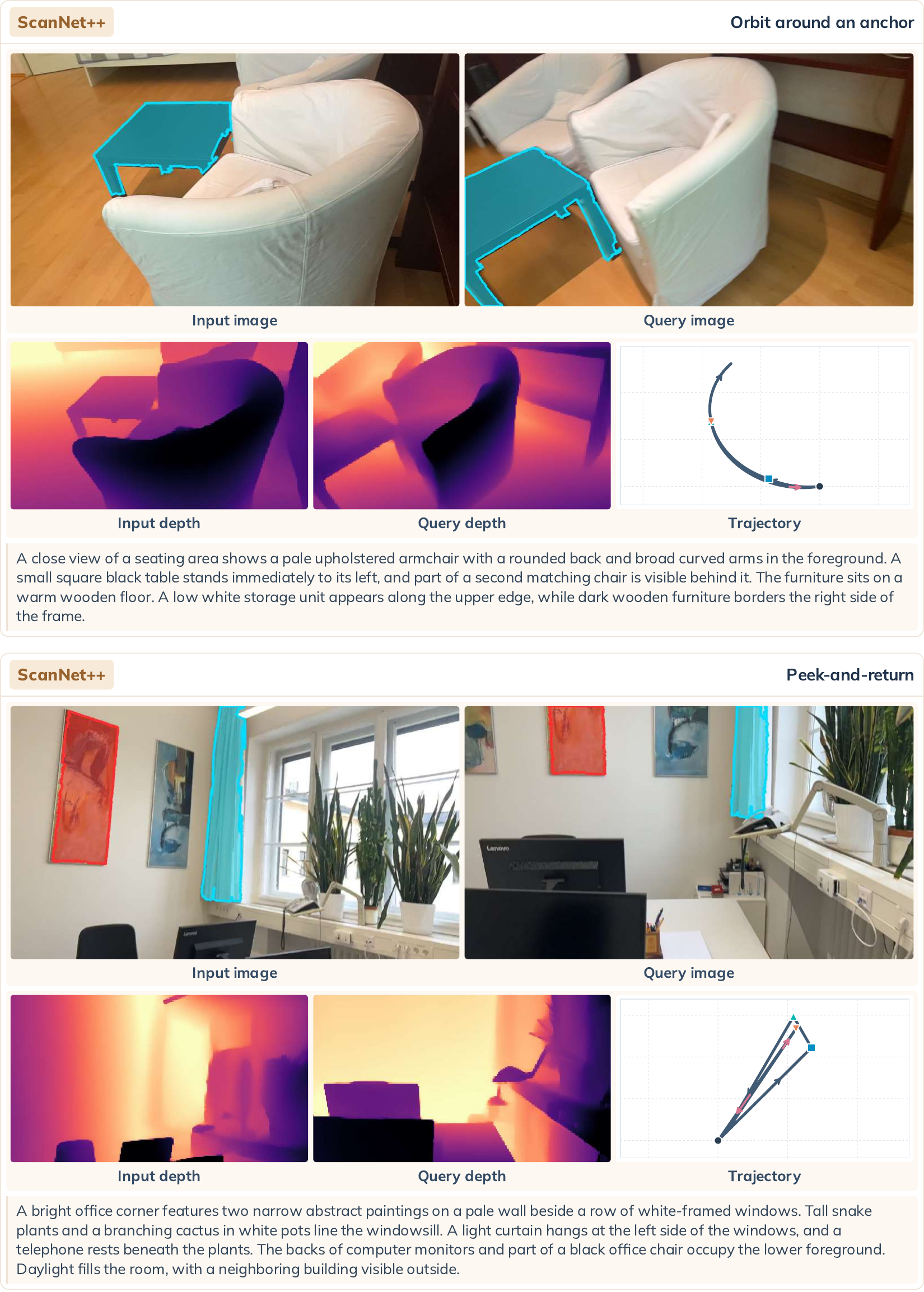}
\caption{Examples of the orbit around an anchor (top) and peek-and-return (bottom) templates from ScanNet++.}
\label{fig:annotated-03}
\end{figure}

\begin{figure}[tp]
\centering
\includegraphics[width=0.92\linewidth]{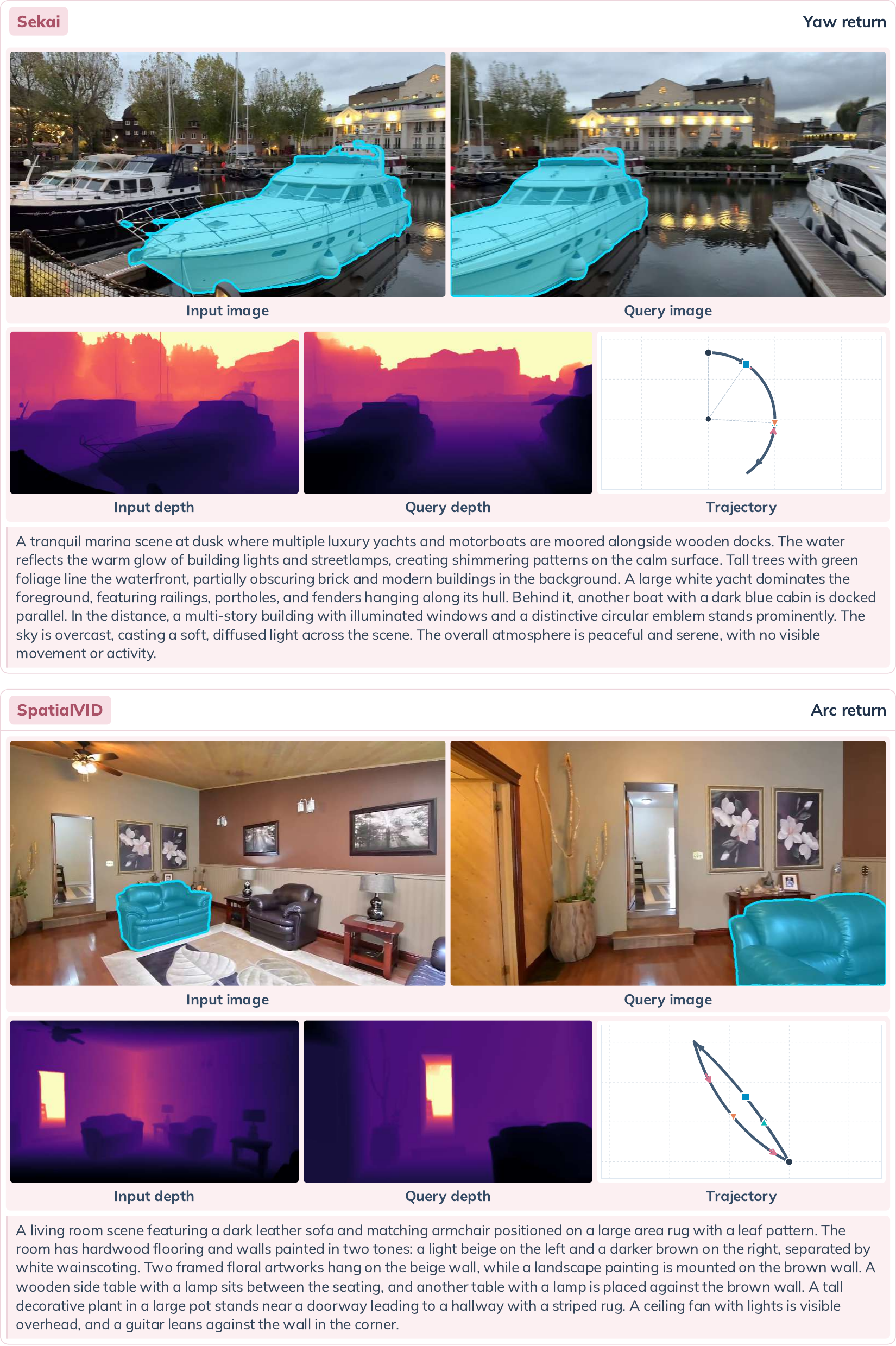}
\caption{Examples of the yaw return template from Sekai (top) and the arc return template from SpatialVID (bottom).}
\label{fig:annotated-05}
\end{figure}

\begin{figure}[tp]
\centering
\includegraphics[width=0.92\linewidth]{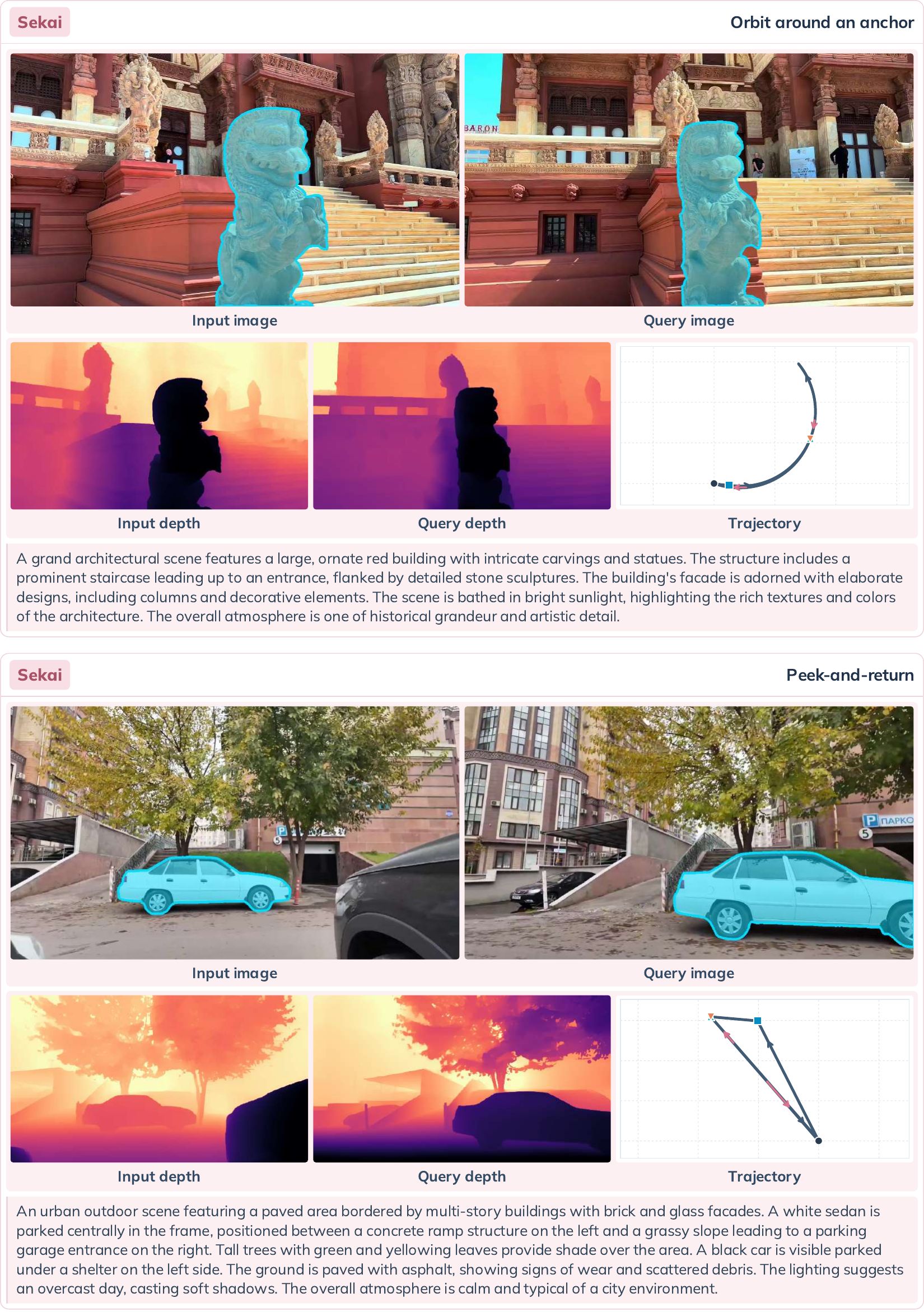}
\caption{Examples of the orbit around an anchor (top) and peek-and-return (bottom) templates from Sekai.}
\label{fig:annotated-06}
\end{figure}

\begin{figure}[tp]
\centering
\includegraphics[width=0.92\linewidth]{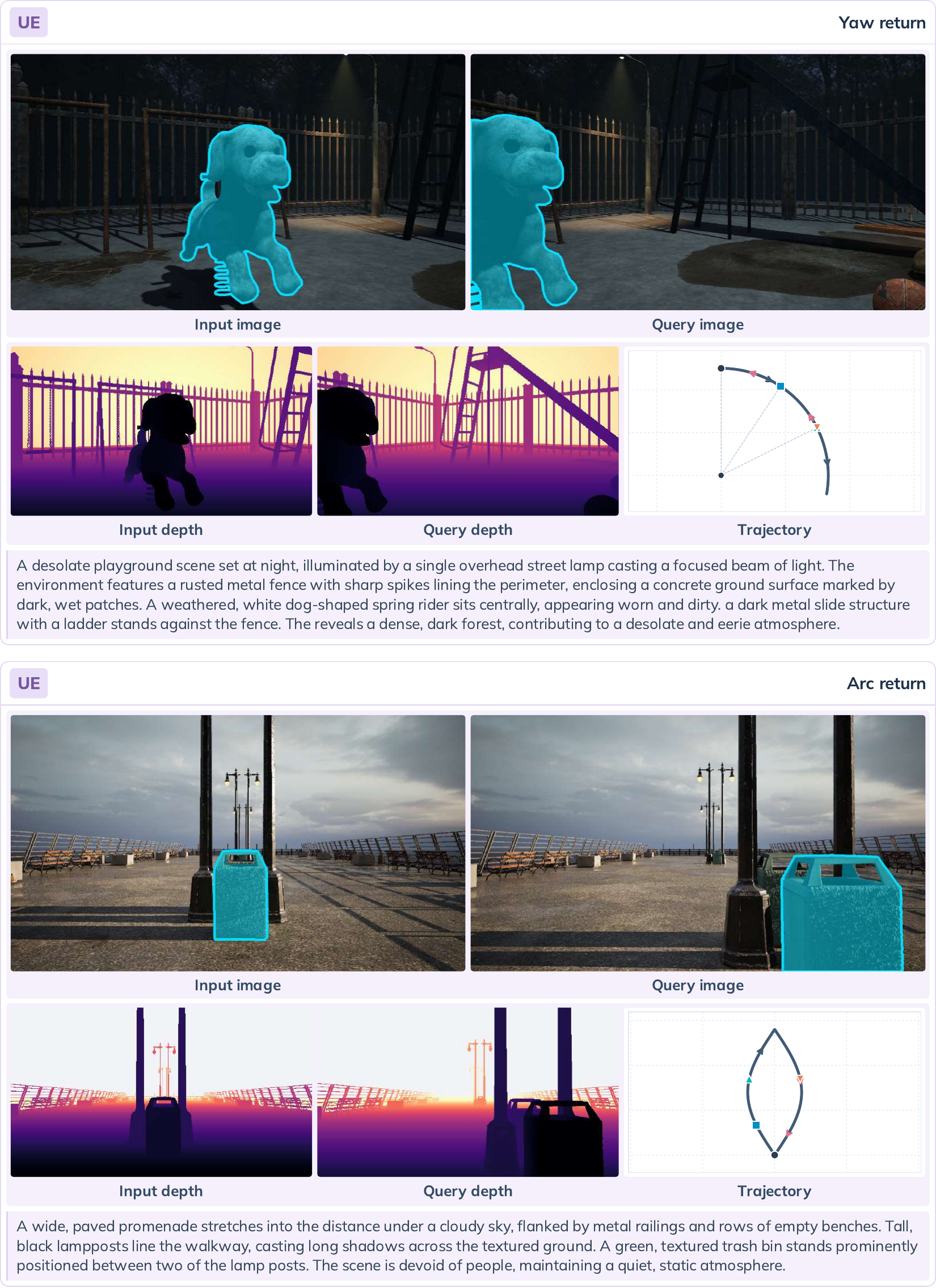}
\caption{Examples of the yaw return (top) and arc return (bottom) templates from UE.}
\label{fig:annotated-08}
\end{figure}

\begin{figure}[tp]
\centering
\includegraphics[width=0.92\linewidth]{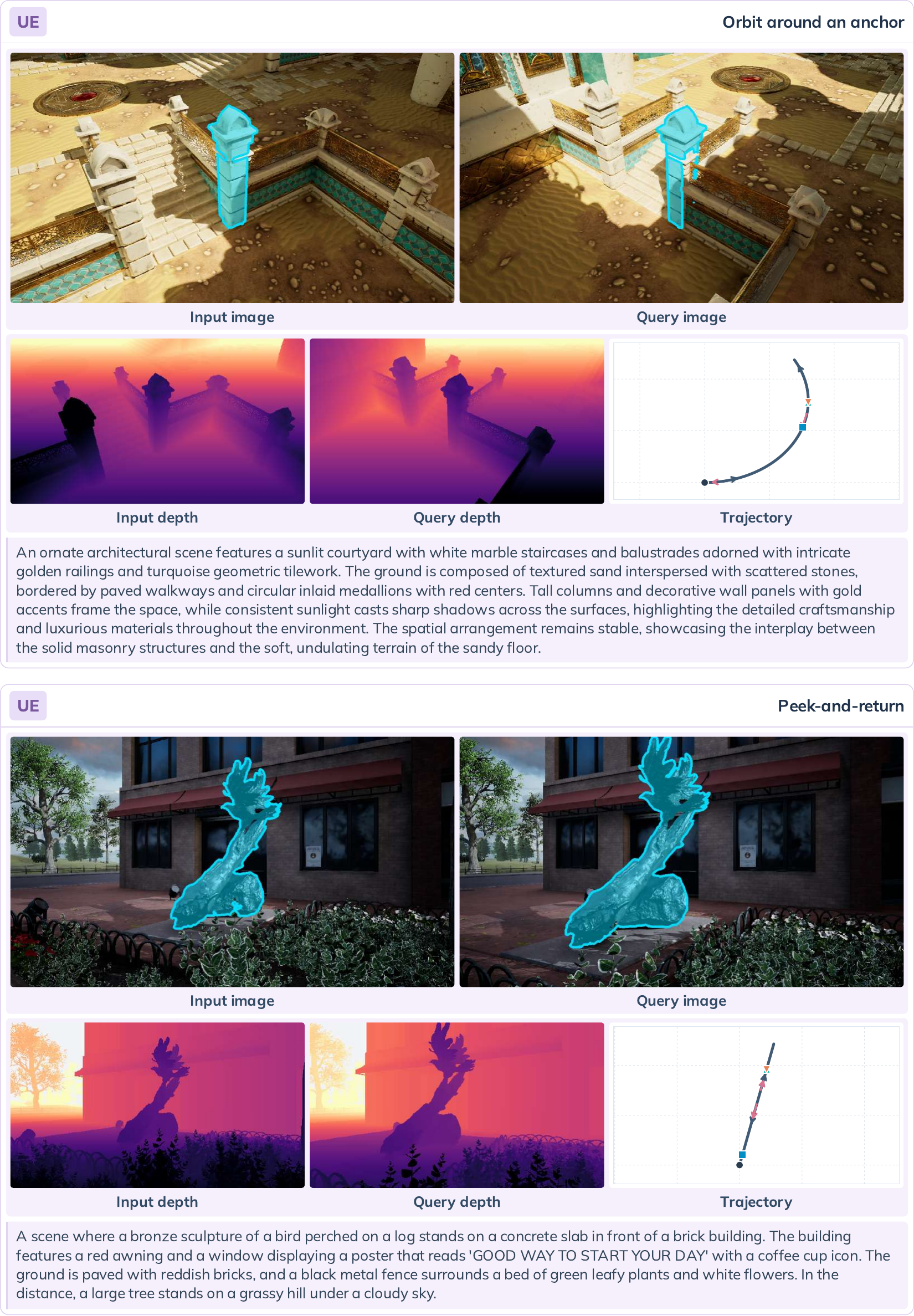}
\caption{Examples of the orbit around an anchor (top) and peek-and-return (bottom) templates from UE.}
\label{fig:annotated-09}
\end{figure}

\begin{figure}[tp]
\centering
\includegraphics[width=0.92\linewidth]{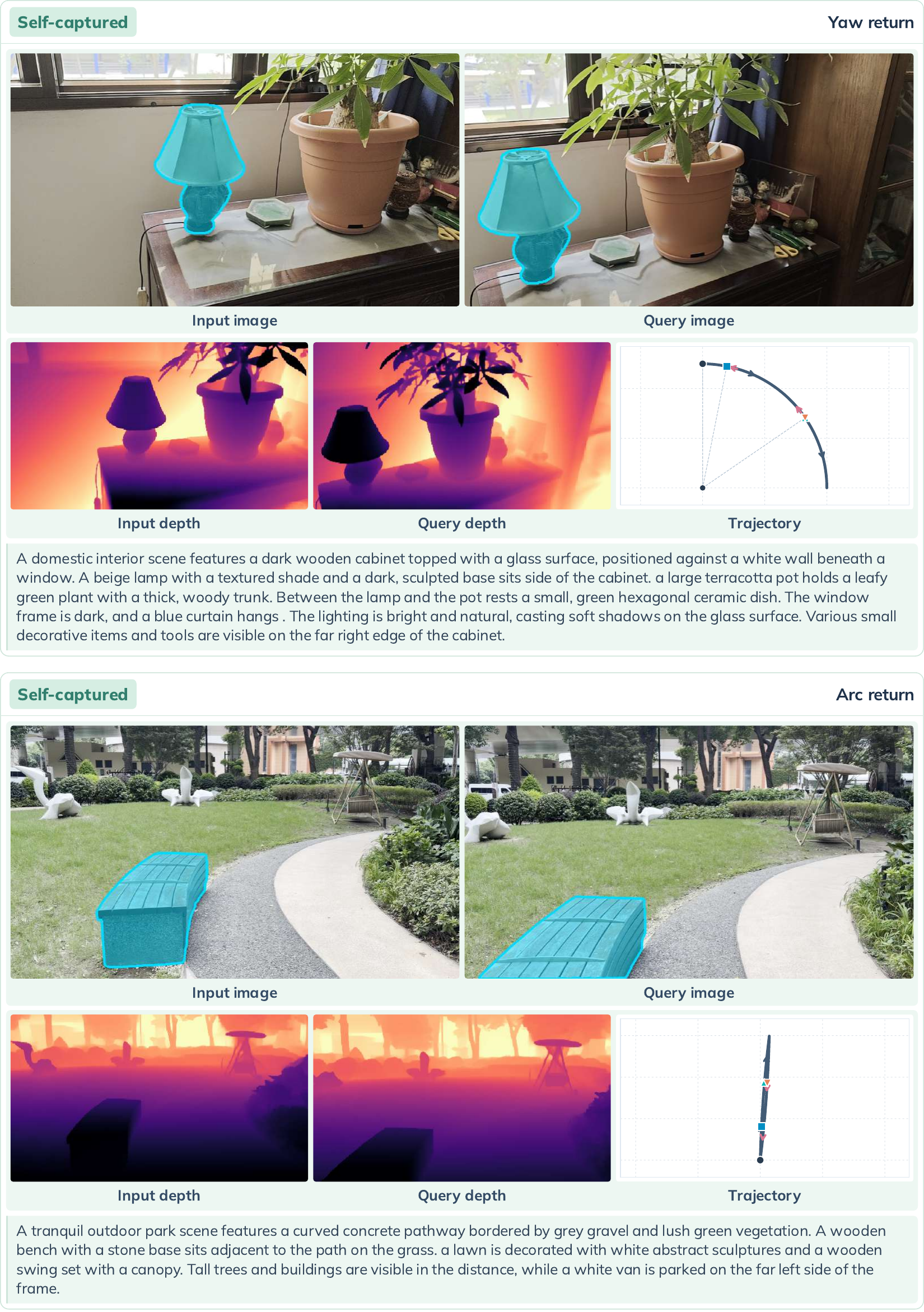}
\caption{Examples of the yaw return (top) and arc return (bottom) templates from self-captured videos.}
\label{fig:annotated-11}
\end{figure}

\begin{figure}[tp]
\centering
\includegraphics[width=0.92\linewidth]{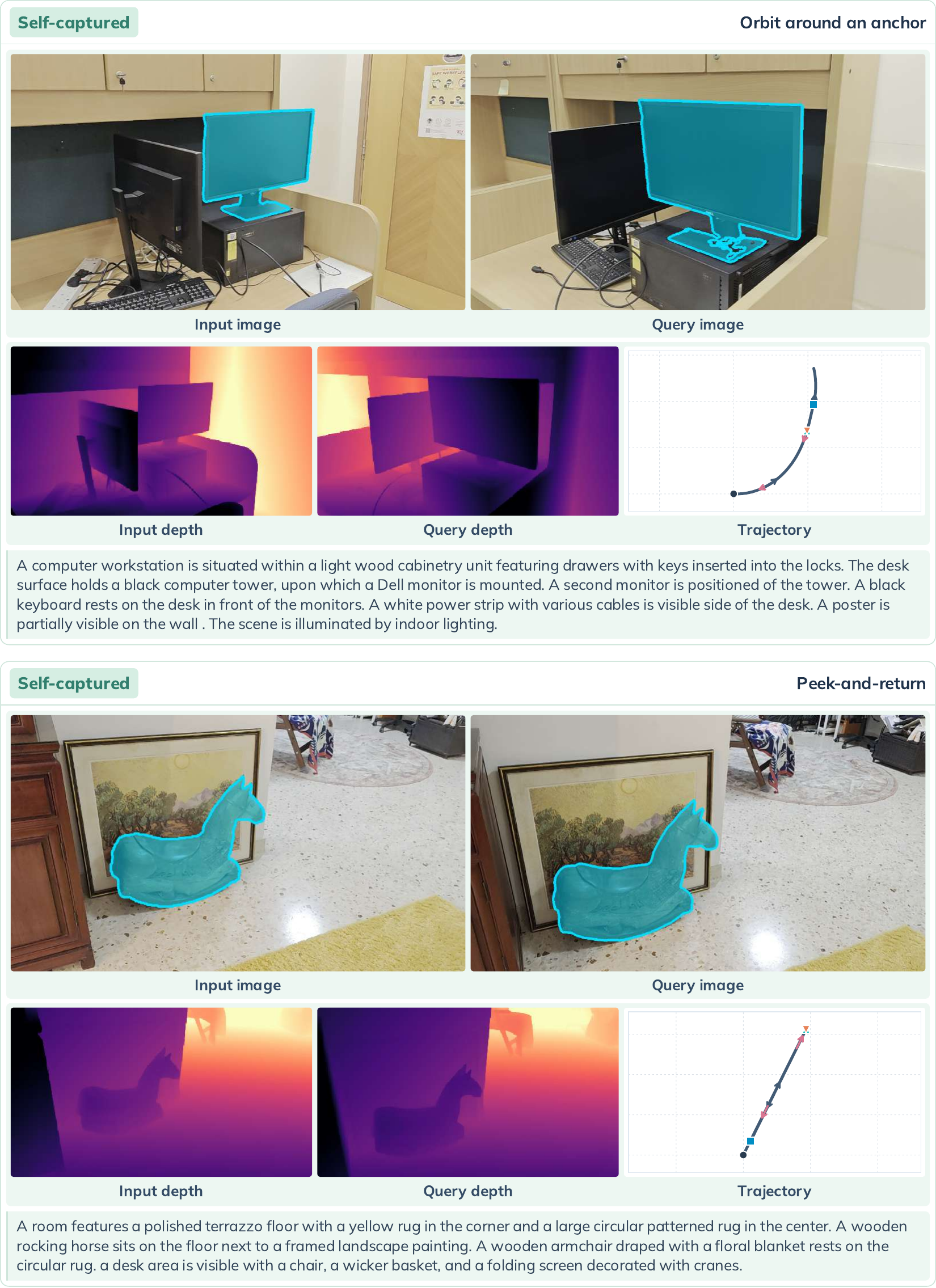}
\caption{Examples of the orbit around an anchor (top) and peek-and-return (bottom) templates from self-captured videos.}
\label{fig:annotated-12}
\end{figure}

\begin{figure}[tp]
\centering
\includegraphics[width=0.88\linewidth]{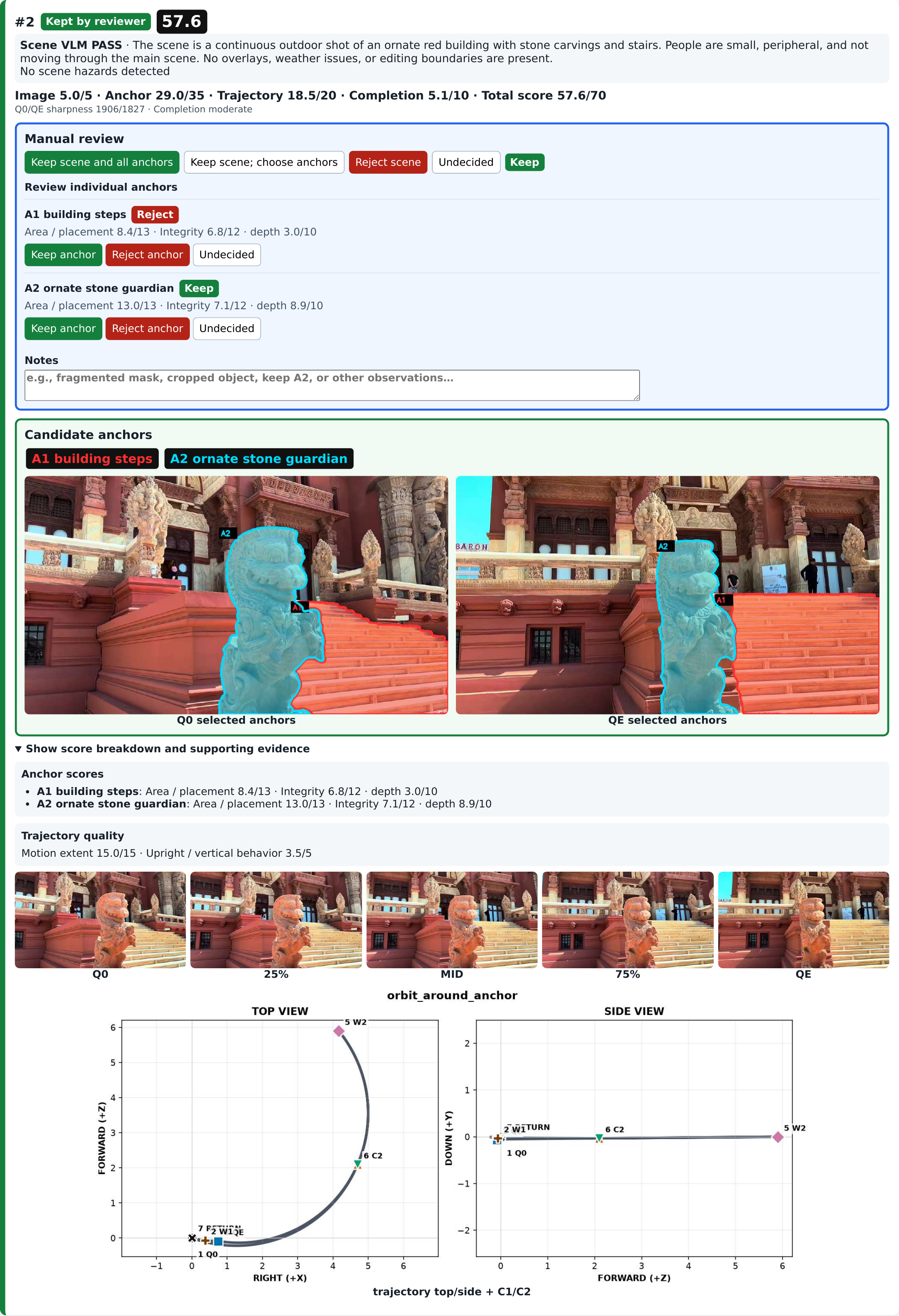}
\caption{\textbf{Manual review of a Sekai orbit case.} The interface combines scene review, annotation scores, per-anchor decisions, reference overlays, sampled frames, and camera trajectories.}
\label{fig:qa-sekai}
\end{figure}

\end{extendedresults}

\orbiareferences{references}
\end{document}